\documentclass[lettersize,onecolumn]{IEEEtran}

\usepackage{amsmath,amsfonts}
\usepackage{algorithmic}
\usepackage{algorithm}
\usepackage{array}
\usepackage{textcomp}
\usepackage{stfloats}
\usepackage{url}
\usepackage{verbatim}
\usepackage{graphicx}
\usepackage{cite}
\usepackage{amssymb,amsthm,subcaption,bm,hyperref,xcolor}

\newtheorem{theorem}{Theorem}
\newtheorem{lemma}{Lemma}
\newtheorem{proposition}{Proposition}
\newtheorem{remark}{Remark}

\newtheorem{subclass}{Subclass}
\DeclareMathOperator*{\argsup}{arg\,sup}

\usepackage[capitalize,noabbrev]{cleveref}
\Crefname{section}{Sec.}{Secs.}
\Crefname{equation}{Eq.}{Eqs.}
\Crefname{figure}{Fig.}{Figs.}
\Crefname{theorem}{Theorem}{Theorems}
\Crefname{lemma}{Lemma}{Lemmas}
\Crefname{proposition}{Proposition}{Propositions}
\Crefname{assumption}{Assumption}{Assumptions}
\Crefname{remark}{Remark}{Remarks}
\Crefname{example}{Example}{Examples}
\Crefname{subclass}{Subclass}{Subclasses}

\newcommand{\td}{\tilde{d}}
\newcommand{\hi}{I}
\newcommand{\tto}{\tilde{o}}

\newcommand{\mbo}{\mathbb{O}}
\newcommand{\tio}{\tilde{\mbo}}
\newcommand{\mbr}{\mathbb{R}}
\newcommand{\mso}{\mathcal{O}}
\newcommand{\calooo}{\vert \mso \vert}

\newcommand{\st}{S}
\newcommand{\ts}{\tilde{\st}}
\newcommand{\pomle}{OMLE-POSI}
\newcommand{\expect}{\mathbb{E}}
\newcommand{\ac}{A}
\newcommand{\hh}{H}
\newcommand{\hth}{\hat{\theta}}

\newcommand{\pr}{\text{Pr}}
\newcommand{\reg}{\text{Reg}}
\newcommand{\gamallt}{\Gamma_{H+1}^{\tau}}
\newcommand{\gamallk}{\Gamma_{H+1}^{k}}
\newcommand{\gamall}{\Gamma_{H+1}}
\newcommand{\gampark}{\Gamma_{h}^{k}}
\newcommand{\gampar}{\Gamma_{h}}
\newcommand{\bio}{\bar{\mbo}}
\newcommand{\mbp}{\mathbb{P}}
\newcommand{\ggam}{\Gamma}
\newcommand{\swi}{q_c}
\newcommand{\mbs}{\tilde{\mathbb{S}}}
\newcommand{\calsss}{\vert \mbs \vert}
\newcommand{\kk}{K}
\newcommand{\mcm}{\mathcal{M}}
\newcommand{\mcs}{\mathcal{S}}
\newcommand{\mca}{\mathcal{A}}
\newcommand{\mco}{\mathcal{O}}
\newcommand{\oo}{O}

\newcommand{\vph}{\vec{\phi}}

\newcommand{\ta}{\eta_1}
\newcommand{\tta}{\eta_2}
\newcommand{\tw}{\tilde{w}}
\newcommand{\tp}{\tilde{p}}
\newcommand{\algo}{OIPOSL}
\newcommand{\algoo}{PDOL}

\newcommand{\bb}{\bm{b}}
\newcommand{\bbb}{\bm{B}}
\newcommand{\hph}{\hat{\Phi}}

\newcommand{\rv}{\color{black}}
\newcommand{\diag}{\mathrm{diag}}
\newcommand{\rank}{\mathrm{rank}}
\newcommand{\Var}{\mathrm{Var}}

\begin{document}

\title{Near-Optimal Partially Observable Reinforcement Learning with Partial Online State Information}

\author{Ming~Shi,~\IEEEmembership{Member,~IEEE,}
Yingbin~Liang,~\IEEEmembership{Fellow,~IEEE,}
and Ness~B.~Shroff,~\IEEEmembership{Fellow,~IEEE}
\thanks{A preliminary version of this paper was presented at IEEE Military Communications Conference (MILCOM) in October 2024~\cite{shi2024designing}.}

\thanks{Ming Shi is with the Department of Electrical Engineering, The State University of New York at Buffalo, Buffalo, NY 14260, USA (e-mail: mshi24@buffalo.edu).}
\thanks{Yingbin Liang is with the Department of Electrical and Computer Engineering, The Ohio State University, Columbus, OH 43210, USA (e-mail: liang.889@osu.edu).}
\thanks{Ness B. Shroff is with the Department of Electrical and Computer Engineering and the Department of Computer Science and Engineering, The Ohio State University, Columbus, OH 43210, USA (e-mail: shroff.11@osu.edu).}
}



\maketitle

\begin{abstract}
Partially observable Markov decision processes (POMDPs) are a general framework for sequential decision-making under latent state uncertainty, yet learning in POMDPs is intractable in the worst case. Motivated by sensing and probing constraints in practice, we study how much online state information (OSI) is sufficient to enable efficient learning guarantees. We formalize a model in which the learner can query only partial OSI (POSI) during interaction. We first prove an information-theoretic hardness result showing that, for general POMDPs, achieving an $\epsilon$-optimal policy can require sample complexity that is exponential unless full OSI is available. We then identify two structured subclasses that remain learnable under POSI and propose corresponding algorithms with provably efficient performance guarantees. In particular, we establish regret upper bounds {\rv with $\tilde{O}(\sqrt{K})$ dependence on the number of episodes $K$}, together with complementary lower bounds, thereby delineating when POSI suffices for efficient reinforcement learning. Our results highlight a principled separation between intractable and tractable regimes under incomplete online state access and provide new tools for jointly optimizing POSI queries and learning control actions.
\end{abstract}

\begin{IEEEkeywords}
Partially observable Markov decision processes (POMDPs), partial online state information (POSI), reinforcement learning (RL), regret bounds, information-theoretic lower bounds.
\end{IEEEkeywords}

\section{Introduction}\label{sec:introduction}

\IEEEPARstart{W}{e} investigate partially observable Markov decision processes (POMDPs) in reinforcement learning (RL) systems, where an agent interacts sequentially with an environment without observing the underlying latent states. Instead, the agent must rely on imperfect observations generated by an emission model based on these latent states. The goal is to maximize cumulative reward over time. POMDPs generalize the classic (i.e., fully observable) Markov decision processes (MDPs) and serve as powerful frameworks for modeling and analyzing various real-world applications. For example, in wireless channel scheduling~\cite{zhao2007decentralized,chen2008joint,ouyang2015exploiting}, a user selects one channel for data transmission at each time while the channel conditions evolve dynamically. Due to sensing and probing constraints, the user often cannot access complete information on all channel conditions. Similarly, autonomous vehicles face limited visibility of global traffic conditions due to constrained reception~\cite{levinson2011towards}, and AI-trained robots typically receive noisy observations of the environment from their sensors due to sensory noise~\cite{akkaya2019solving}. These challenges are also prevalent in diverse domains, including healthcare~\cite{hauskrecht2000planning}, recommendation systems~\cite{li2010contextual}, games~\cite{berner2019dota}, and economic modeling~\cite{zheng2020ai}.

Existing information-theoretic results show that learning in general POMDPs is intractable in the worst case~\cite{krishnamurthy2016pac}, and planning in POMDPs is PSPACE-complete~\cite{papadimitriou1987complexity,mundhenk2000complexity}. This stands in contrast to fully observable MDPs, for which numerous efficient algorithms have been developed~\cite{azar2017minimax,jin2018q,agarwal2019reinforcement,bai2019provably,jin2020provably,ayoub2020model,xie2020learning,foster2021statistical,wang2021online,jin2022v,shi2023nearswitch,zhu2023provably,zhong2023can,shi2023near}. The main challenge in POMDPs lies in the lack of latent state information: the Markov property that simplifies fully observable MDPs does not hold for the agent's observations. Without this property, the agent can no longer rely on the current observation alone to predict future observations and rewards, and instead must account for the entire history of past observations and actions, \emph{exponentially} increasing the complexity of planning and learning.

Despite this worst-case intractability, recent work has identified structured POMDP variants where efficient algorithms with polynomial dependence on problem parameters can be developed, e.g., $m$-step decodable POMDPs~\cite{efroni2022provable}, block MDPs~\cite{zhang2022efficient}, latent MDPs~\cite{kwon2021rl}, and POMDPs with reachability~\cite{xiong2022sublinear}. A related line of work studies tractable subclasses under weakly revealing conditions~\cite{liu2023optimistic}, predictive state representations~\cite{zhong2022posterior}, and $B$-stability~\cite{chen2022partially}. However, these conditions may not hold in practical scenarios such as robotics~\cite{pinto2018asymmetric,lee2023learning} and networking~\cite{sinclair2023hindsight,lee2023learning}. Moreover, the resulting regret guarantees can scale unfavorably when the emission measure differences between disjoint latent-state supports are small.

To avoid strong assumptions on the emission model or regret dependencies that deteriorate in poorly separated regimes, some recent work has studied \emph{hindsight state information}~\cite{sinclair2023hindsight,lee2023learning,guo2023sample}, where (full) state information is revealed at the end of each episode of interaction. This line of work is motivated by the observation that, although the true latent state may not be accessible before the agent takes an action, it may become accessible in hindsight. However, these studies rely on the assumption of \emph{full} hindsight state information, which may be unrealistic. For example, in wireless communication systems~\cite{zhao2007decentralized,chen2008joint,ouyang2015exploiting}, the conditions of unprobed channels remain unknown to the user, even in hindsight. Similarly, in autonomous driving~\cite{levinson2011towards,pinto2018asymmetric,jennings2019study}, only the conditions of roads that are locally observed or actively probed can be known to vehicles or robots. This raises a natural and practical question: can tractability still be achieved if state information is not fully revealed, even in hindsight? Existing lower bounds~\cite{krishnamurthy2016pac,liu2022partially} imply that such settings remain fundamentally intractable in general.

Motivated by this gap, we investigate the fundamental limits of \emph{partial} (i.e., not full) state information revealed \emph{online during an episode} (i.e., not only at the end of an episode). We refer to this as \emph{``Partial Online State Information'' (POSI)}, e.g., the conditions of the channels that are actively probed at each time. To model POSI concretely, we introduce a formulation based on vector-structured states. Specifically, the latent state is represented as a $d$-dimensional vector, with each element corresponding to an abstract feature (e.g., the condition of one wireless channel or traffic road). At each step of an episode, a subset of $\td$ $(1\leq \td < d)$ state elements can be revealed to the agent after a query, e.g., the $\td$ probed channels.

\subsection{Contributions}\label{subsec:contribution}

This paper addresses a central question in partially observable reinforcement learning: \emph{Can POMDPs become tractable when the learner has access only to Partial Online State Information (POSI)?} Our main contributions are summarized as follows.

\emph{(1) Fundamental hardness under POSI.} In~\cref{theorem:lowerboundproblem2}, we establish a new lower bound showing that, in the worst case, unless \emph{full} OSI is available, achieving an $\epsilon$-optimal policy can require sample complexity that scales exponentially with the episode length, namely $\tilde{\Omega} \big(\ac^{\hh}/\epsilon^2\big)$, where $\ac$ is the number of actions and $\hh$ is the horizon. This result draws a sharp separation between POMDPs with partial OSI and those with full OSI (or with full hindsight state information). Notably, the bound rules out the intuitive strategy of ``stitching together'' POSI across multiple steps (e.g., querying different state elements over time) to effectively recover full state information with only polynomial samples: our construction ensures that, under polynomial sample size, POSI observed at each step (even aggregated across steps) remains insufficient to identify an $\epsilon$-optimal policy.

This hardness result naturally raises the question: \emph{Which structured POMDPs remain learnable under POSI?} To answer this, we identify two tractable subclasses and provide algorithms with provable regret guarantees.

\emph{(2) Tractable~\cref{subclass1}: POSI with additional partial-revealing noisy information.} In~\cref{sec:problem3}, we identify a tractable subclass in which, in addition to queried POSI, the learner receives partial noisy information about the \emph{unqueried} state components via a partial emission mechanism. This models settings where unobserved components can be weakly inferred from correlations and side measurements (e.g., spatial/frequency correlations in wireless systems). {\rv This subclass can be viewed (via a state augmentation) as a two-step weakly revealing POMDP. We provide a detailed discussion of this connection and its implications in~\cref{sec:problem3}.} For this subclass, we develop a provably efficient algorithm and establish regret upper and lower bounds. To achieve a regret that improves with the query capability $\td$, our algorithm design and analysis require a nontrivial extension of operator-based learning methods (e.g., observable-operator/spectral-style approaches~\cite{liu2022partially,jaeger2000observable}) to handle query-dependent observations and the coupling between POSI querying and action selection under adaptively chosen queries.

\emph{(3) Tractable~\cref{subclass2}: independent state-element transitions with restricted query adaptation.} In~\cref{sec:problem4}, we identify a second tractable subclass characterized by independent transitions across state elements and a restricted class of query policies. In this setting, we do not assume additional noisy information for unqueried elements. Tractability is enabled instead by the transition independence together with controlled query adaptation. Designing a provably efficient algorithm in this regime requires addressing correlations induced by the interaction between the query policy and the action policy, as well as carefully controlling the interplay between within-episode and across-episode estimation biases. We further show that the resulting regret bound improves as the querying capability $\td$ increases, quantifying the value of enhanced probing under partial observability.

\subsection{Related Work on POMDPs}\label{app:relatedwork}

Theoretical studies on partially observable Markov decision processes (POMDPs) have a long history in control, planning, and systems theory~\cite{jaeger2000observable,aastrom1965optimal,smallwood1973optimal,sondik1978optimal,kaelbling1998planning,hauskrecht2000value}. Early works analyzed structural properties and solution methods for POMDP planning, including limitations of dynamic programming~\cite{aastrom1965optimal}, properties of optimal policies and value functions in finite-state settings~\cite{smallwood1973optimal}, approximation methods for stationary policies~\cite{sondik1978optimal}, and finite-memory controllers motivated by robotics~\cite{kaelbling1998planning}. These works primarily focused on planning and approximation, and did not provide finite-sample regret guarantees for online learning.

\emph{Performance-guaranteed RL for structured POMDPs.} There has been significant recent progress on RL with provable guarantees in structured POMDP models. A prominent line of work assumes that latent states can be decoded (exactly or approximately) from short histories or predictive features, including $m$-step decodable POMDPs~\cite{efroni2022provable} and reactive POMDPs~\cite{jiang2017contextual}. Other works study structural observation models such as block MDPs~\cite{zhang2022efficient} and latent MDPs~\cite{kwon2021rl}, as well as settings that emphasize reachability-type objectives~\cite{xiong2022sublinear}. Related approaches characterize learnability through observability and separation conditions, e.g., $\gamma$-observability~\cite{golowich2022planning}. Another influential direction uses predictive state representations and posterior/predictive embeddings to obtain tractable learning and planning procedures~\cite{chen2022partially,zhong2022posterior}, and more generally studies decision making with structured observations that includes POMDPs as a special case~\cite{foster2021statistical,chen2022unified}.

A particularly relevant line of work establishes tractability under \emph{weakly revealing} conditions, where the current latent state can be statistically identified using a short window of near-term future observations; see, e.g., $m$-step weakly revealing POMDPs and optimistic maximum-likelihood style methods in~\cite{liu2022partially,liu2023optimistic}. These results provide powerful guarantees but typically require assumptions controlling state distinguishability through the emission model.

\emph{Hindsight state information.} To mitigate sensitivity to emission-based separation conditions, recent works have studied settings where additional state information becomes available in hindsight~\cite{sinclair2023hindsight,lee2023learning}. Specifically,~\cite{sinclair2023hindsight} considers POMDPs with exogenous inputs that become observable after actions are taken, and~\cite{lee2023learning} studies settings where the latent state is revealed at the end of each episode (full hindsight state information). While hindsight information can substantially simplify learning, these frameworks typically rely on \emph{full} hindsight state access or full observability, which may be unrealistic in applications where unobserved components remain inaccessible even retrospectively.

In contrast to the above models, we study \emph{partial online state information} (POSI), where the learner adaptively queries only a subset of state components during the episode. This differs fundamentally from both passive observation models (where the observation process is exogenously specified) and full hindsight state revelation. Moreover, while our~\cref{subclass1} can be viewed (via state augmentation) as a two-step weakly revealing POMDP, the POSI setting introduces an additional \emph{query-driven} information acquisition layer and query-dependent observation operators coupled with action selection. These features necessitate new algorithmic and analytical treatments beyond directly applying existing weakly revealing frameworks.

\begin{figure}[t]
\centering
\begin{subfigure}[b]{0.6\textwidth}
    \centering
    \includegraphics[width=\linewidth]{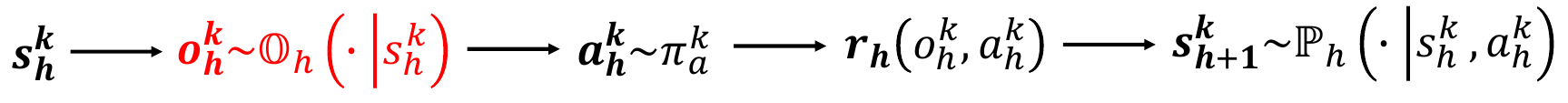}
    \caption{Traditional POMDP}
    \label{fig:sketchgeneralpomdp}
\end{subfigure}
\hfill
\begin{subfigure}[b]{0.99\textwidth}
    \centering
    \includegraphics[width=\linewidth]{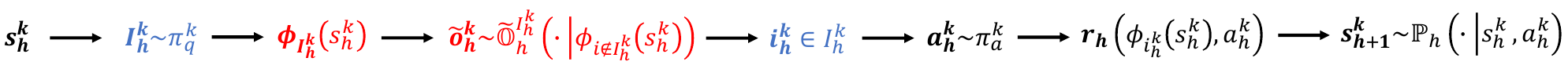}
    \caption{\cref{subclass1}}
    \label{fig:sketchpomdplimitedhopwrc}
\end{subfigure}
\hfill
\begin{subfigure}[b]{0.99\textwidth}
    \centering
    \includegraphics[width=\linewidth]{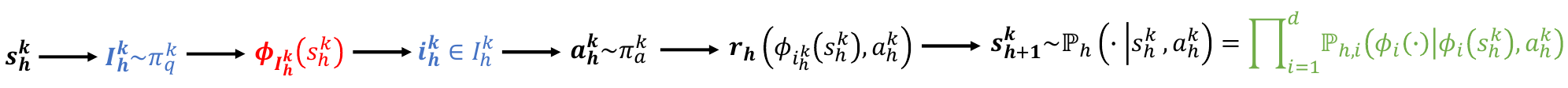}
    \caption{\cref{subclass2}}
    \label{fig:sketchpomdplimitedho}
\end{subfigure}
\caption{One-step interaction sketch for the traditional POMDP and the two POSI subclasses studied in this paper}
\label{fig:sketchpomdp}
\end{figure}

\section{Problem Formulation}\label{sec:problemformulation}

In this section, we first review the traditional episodic partially observable Markov decision process (POMDP) formulation. We then introduce partial online state information (OSI), which is the central focus of this paper.

\subsection{Traditional Episodic POMDP}\label{subsec:genericpomdp}

A traditional episodic POMDP is modeled by a tuple $\mcm = (\mcs,\mca,\mco,\hh,\Delta_1,\mbp,\mbo,r)$, where $\mcs$, $\mca$, and $\mco$ denote the state space with $\st$ states, the action space with $\ac$ actions, and the observation space with $\oo$ observations, respectively; $\hh$ is the number of steps in an episode; $\Delta_1: \mcs \rightarrow [0,1]$ is the initial state distribution at the beginning of each episode; $\mbp = \{ \mbp_h: \mcs \times \mca \times \mcs \rightarrow [0,1] \}_{h=1}^{\hh-1}$ and $\mbo = \{ \mbo_h: \mcs \times \mco \rightarrow [0,1] \}_{h=1}^{\hh}$ denote the \emph{unknown} transition and emission probability kernels, respectively; and $r = \{ r_h: \mco \times \mca \rightarrow [0,1] \}_{h=1}^{\hh}$
denotes the reward functions.

In the online setting, an agent interacts with the environment over $\kk$ episodes. At each step $h$ of episode $k$ (see~\cref{fig:sketchgeneralpomdp}), the agent receives a noisy observation $o_h^k \sim \mbo_h(\cdot \mid s_h^k)$, where $s_h^k$ is the \emph{unknown} latent state. The agent then selects an action $a_h^k \in \mca$ according to a policy $\pi_a^k$ and receives reward $r_h(o_h^k,a_h^k)$. The environment subsequently transitions to the next state $s_{h+1}^k \sim \mbp_h(\cdot \mid s_h^k,a_h^k)$. The goal is to maximize the expected cumulative reward $\expect_{\{\mbp,\mbo,\Delta_1\}} \left[\sum_{k=1}^{\kk}\sum_{h=1}^{\hh} r_h(o_h^k,a_h^k)\right]$.

\subsection{Partial Online State Information (POSI)}\label{subsec:formulatestructuredstate}

Based on the traditional episodic POMDP described above, we now introduce the episodic POMDP with partial online state information (POSI) studied in this paper. Specifically, we consider vector-structured latent states. Each state $s \in \mcs$ is represented by a $d$-dimensional feature vector
\begin{align}
\vph(s) = [\phi_1(s),\ldots,\phi_d(s)]^{\top} \in \mbs^d, \nonumber
\end{align}
where $\mbs$ is the finite alphabet of each \emph{sub-state} (state element), and $[\cdot]^{\text{T}}$ denotes vector transpose. We denote the cardinality of $\mbs$ by $\calsss$ and denote the index set $\{1,2,\ldots,d\}$ by $[d]$.

{\rv
\emph{Base POSI Protocol:} At each step $h$ of episode $k$, the agent interacts with the environment as follows:
\begin{itemize}
    \item \textbf{Step~1 (Query):} According to a \emph{query policy} $\pi_q^k$, the agent selects a subset $\hi_h^k \subseteq [d]$ of $\td$ sub-states to query, where $1 \le \td < d$. The query is chosen before the action and reward at step $(k,h)$.
    \item \textbf{Step~2 (POSI observation):} The queried sub-states $\{\phi_i(s_h^k)\}_{i \in \hi_h^k}$ are revealed to the agent.
    \item \textbf{Step~3 (Action selection):} According to an \emph{action policy} $\pi_a^k$, the agent selects one queried sub-state $i_h^k \in \hi_h^k$ and takes an action $a_h^k \in \mca$.
    \item \textbf{Step~4 (Reward):} The agent receives a reward $r_h(\phi_{i_h^k}(s_h^k),a_h^k)$, where $r_h: \mbs \times \mca \rightarrow [0,1]$.
    \item \textbf{Step~5 (Transition):} The environment transitions to the next state $s_{h+1}^k \sim \mbp_h(\cdot \mid s_h^k,a_h^k)$.
\end{itemize}

Unless otherwise specified, both $\pi_q^k$ and $\pi_a^k$ may be history-dependent: $\pi_q^k$ selects $\hi_h^k$ based on the past interaction history, and $\pi_a^k$ selects $(i_h^k,a_h^k)$ based on the history as well as the revealed POSI $\{\phi_i(s_h^k)\}_{i\in \hi_h^k}$.

\begin{remark}
Under the base POSI protocol, the agent observes only the queried sub-states $\{\phi_i(s_h^k)\}_{i \in \hi_h^k}$ and does \emph{not} receive a separate global observation $o_h^k$. In~\cref{sec:problem2}, we establish the fundamental hardness of learning under this base protocol. In~\cref{sec:problem3} and~\cref{sec:problem4}, we then identify two tractable subclasses (\cref{subclass1} and~\cref{subclass2}). Specifically, in~\cref{subclass1}, the agent additionally receives a partial noisy observation $\tilde o_h^k$ about the unqueried sub-states (generated by a partial emission model), whereas in~\cref{subclass2}, no additional noisy observation is present.
\end{remark}
}

This model is motivated by a range of practical applications, including wireless channel scheduling~\cite{zhao2007decentralized,chen2008joint,ouyang2015exploiting}, autonomous driving~\cite{levinson2011towards,pinto2018asymmetric,jennings2019study}, robotics~\cite{akkaya2019solving,lee2023learning,silver2010monte}, and healthcare~\cite{qu2022scalable,hauskrecht2000planning}. We briefly describe two motivating examples below.

\paragraph{Motivating example 1 (Wireless channel scheduling)} Each sub-state $\phi_i(s)$ represents the condition of a wireless channel (e.g., busy or idle). At each step, the user probes a subset of channels indexed by $\hi_h^k$ and observes their conditions via POSI. Due to probing and sensing constraints, not all channels can be probed. The user then selects one probed channel $i_h^k \in \hi_h^k$ for transmission and receives a reward associated with the chosen channel and action.

\paragraph{Motivating example 2 (Autonomous driving)} Each sub-state $\phi_i(s)$ represents the condition of a traffic road (e.g., congestion level). At each step, the vehicle queries traffic conditions on a subset of roads indexed by $\hi_h^k$, observes their conditions, and selects one queried road $i_h^k \in \hi_h^k$ to follow. Due to sensing latency and energy constraints, only partial road information can be queried at each step. Finally, it receives a reward associated with the chosen road and the driving action.

\subsection{Performance Metric}\label{subsec:formulateperformancemetric}

We use the $V$-value below to denote the expected reward in episode $k$ by following query policy $\pi_{q}^k$ and action policy $\pi_{a}^k$:
\begin{align}
V^{\pi^k} \triangleq \expect_{\pi^k,\mbp,\Delta_1} \left[ \sum\nolimits_{h=1}^{\hh} r_h \left(\phi_{i_h^{k}}(s_h^k),a_h^{k}\right) \right], \label{eq:definevvaluemetric}
\end{align}
where $\pi^k = (\pi_q^k,\pi_a^k)$ denotes the joint policy and the expectation is over the initial state, state transitions, and any randomness in the policy. We take regret as the performance metric:
\begin{align}\label{eq:defregret}
\reg^{\pi^{1:\kk}}(\kk) \triangleq \sum\nolimits_{k=1}^{\kk} \left[ V^{\pi^*} - V^{\pi^k} \right],
\end{align}
where $\pi^* \triangleq \argsup_{\{\pi_q,\pi_a\}} V^{\pi}$ denotes an optimal (possibly history-dependent) joint policy. \emph{This work studies the fundamental limits of online state information (OSI) and provides provable regret guarantees for partially observable RL with POSI.} For the reader's convenience, we summarize all notations in~\cref{tab:notations} at the beginning of the Appendix.

\section{Fundamental Hardness Without Full Online State Information}\label{sec:problem2}

In this section, we address the following question: \emph{can POMDPs with online state information (OSI) be made tractable without \emph{full} OSI?} Recall that \emph{full} OSI corresponds to \(\td=d\), i.e., the learner can query all \(d\) state components at each step. The theorem below establishes a sharp worst-case obstruction: if \(\td<d\), then there exist POSI-POMDP instances for which any algorithm requires exponentially many episodes to obtain an \(\epsilon\)-optimal policy.

\begin{theorem}\label{theorem:lowerboundproblem2}
\textbf{(Intractability without full OSI).}
Fix any \(\hh\ge 2\), \(\ac\ge 2\), and any \((d,\td)\) with \(1\le \td<d\). For the POSI model in~\cref{subsec:formulatestructuredstate}, there exist instances such that the following holds: for any (possibly randomized) learning algorithm that, after \(\kk\) episodes, outputs a policy \(\hat\pi\), if $\kk = o\left(\frac{\ac^{\hh-1}}{\epsilon^2}\right)$, then with probability at least \(1/3\), \(\hat\pi\) is \emph{not} \(\epsilon\)-optimal, i.e., \(V^{\hat\pi} < V^{\pi^*}-\epsilon\). Equivalently, achieving \(\Pr\{V^{\hat\pi}\ge V^{\pi^*}-\epsilon\}\ge 2/3\) requires \(\kk=\Omega(\ac^{\hh-1}/\epsilon^2)\).
\end{theorem}

\cref{theorem:lowerboundproblem2} shows that, in the worst case, partial OSI cannot yield a polynomial sample complexity \(\mathrm{poly}(\ac,\hh,\st,\kk)\) guarantee for learning \(\epsilon\)-optimal policies. At first glance, this may seem counterintuitive: one might hope to reconstruct missing state information by combining POSI across steps (e.g., querying different components over time). The hard instance behind~\cref{theorem:lowerboundproblem2} rules this out by ensuring that (i) the POSI transcript before the terminal reward is statistically \emph{uninformative} about whether the learner is on the ``correct'' path, and (ii) only \emph{one} action sequence yields a slightly better terminal payoff, so the learner must effectively search over \(\ac^{\hh-1}\) possibilities. A proof sketch is given below, and the complete proof is provided in Appendix~\ref{app:pftheoremlowerboundproblem2}.

\begin{remark}\label{rem:lb_still_holds_subclass1}
The intractability in~\cref{theorem:lowerboundproblem2} still holds even if, in addition to POSI, the learner receives extra partial noisy observations (as in~\cref{sec:problem3}): we can choose the additional observation channel to be \emph{uninformative} (identical across latent states), so it provides no extra information.
\end{remark}

\begin{proof}[Proof sketch]
We construct a family of POSI-POMDP instances indexed by a hidden optimal action sequence. In each instance, there is a unique length-$(\hh-1)$ action sequence that keeps the latent process in a ``good'' mode; any deviation switches the process to a ``bad'' mode that persists. Crucially, for any query budget $\td<d$, the queried POSI observations at steps $1,\ldots,\hh-1$ have the same distribution in both modes (even conditioned on the learner's past), so they do not statistically reveal whether a deviation has occurred. We then assign rewards so that only when the agent follows the unique optimal action sequence does it receive a terminal reward with mean $\frac{1}{2}+\epsilon$; otherwise the terminal reward has mean $\frac{1}{2}$. Therefore, each episode yields one Bernoulli sample whose mean depends only on whether the chosen action sequence matches the hidden optimal sequence. This induces a reduction to identifying the best arm among $\ac^{\hh-1}$ Bernoulli arms with gap $\epsilon$, which requires $\Omega(\ac^{\hh-1}/\epsilon^2)$ episodes to succeed with constant probability.
\end{proof}

\section{Fundamental Tractability under POSI and Partial Noisy Observations}\label{sec:problem3}

In~\cref{sec:problem2}, we established that learning can be information-theoretically intractable under the \emph{base} POSI protocol when $\td<d$, even if the learner can adaptively query different state components over time. This naturally motivates the following question: \emph{Are there practically meaningful subclasses in which POSI is still sufficient for provably efficient learning?}

In this section, we answer this question affirmatively by identifying a tractable subclass in which, in addition to queried POSI, the learner receives a \emph{partial noisy observation} about the \emph{unqueried} state components. {\rv Intuitively, POSI provides exact information on a small subset of components, while the auxiliary channel provides \emph{weak but non-degenerate} information about the remaining components, enabling efficient latent-state inference.}

\begin{subclass}\label{subclass1}
\textbf{(POMDPs with POSI and partial noisy observations).} At each step $h$ of episode $k$ (see~\cref{fig:sketchpomdplimitedhopwrc} for a sketch), the interaction is as follows:
\begin{itemize}
    \item \textbf{Step-i (Query):} The agent selects a query set $\hi_h^k\subseteq[d]$ with $|\hi_h^k|\le \td$.
    \item \textbf{Step-ii (POSI):} The queried POSI $\phi_{\hi_h^k}(s_h^k)=\{\phi_i(s_h^k)\}_{i\in \hi_h^k}$ is revealed to the agent.
    \item \textbf{Step-iii (Auxiliary partial noisy observation):} The agent additionally observes $\tto_h^k$, generated according to a \emph{query-indexed} partial emission kernel $\tio_h^{\hi_h^k} \left(\cdot \middle| \{\phi_i(s_h^k)\}_{i\notin \hi_h^k}\right)$. {\rv Define the \emph{unqueried sub-state vector} $z_h^k(\hi)\triangleq \{\phi_i(s_h^k)\}_{i\notin \hi}\in\mbs^{d-|\hi|}$. For each fixed $(h,\hi)$, $\tio_h^{\hi}$ can be viewed as a matrix in $\mbr^{\calooo\times \ts(\hi)}$ with $\ts(\hi)\triangleq \calsss^{d-|\hi|}$, whose columns are indexed by assignments of $z_h^k(\hi)$.} We assume a \emph{partially revealing} condition: there exists $\alpha>0$ such that
    \begin{align}
        \min\nolimits_{\{h,\hi:|\hi|\le \td\}} \sigma_{\ts(\hi)}(\tio_h^{\hi}) \ge \alpha, \label{eq:partial_revealing_condition}
    \end{align}
    where $\sigma_{\ts(\hi)}(\cdot)$ denotes the $\ts(\hi)$-th largest singular value. {\rv (Equivalently, for every feasible query, $\tio_h^{\hi}$ has full column rank and is uniformly well-conditioned.)} Let $\Theta$ denote the set of all valid parameters $\theta \triangleq (\mbp,\tio,\Delta_1)$ satisfying~\eqref{eq:partial_revealing_condition}.
    \item \textbf{Step-iv (Action and reward):} The agent selects a sub-state $i_h^k\in \hi_h^k$, takes an action $a_h^k\in\mca$, and receives reward $r_h(\phi_{i_h^k}(s_h^k),a_h^k)$.
    \item \textbf{Step-v (Transition):} The next state $s_{h+1}^k$ is drawn according to $\mbp_h(\cdot \mid s_h^k,a_h^k)$.
\end{itemize}
\end{subclass}

Although we allow $|\hi_h^k|\le \td$, one may w.l.o.g.\ restrict to $|\hi_h^k|=\td$ by padding the query set with arbitrary indices, since querying additional components can only increase the available information and action choices (the agent may still ignore the padded components when selecting $i_h^k$). Under this convention, $\ts(\hi)=\calsss^{d-\td}$ is constant.

In wireless systems, probing every channel with high-fidelity measurements at every decision epoch is typically costly (time/energy/overhead), so only a subset of channels can be directly sensed. In this context, POSI corresponds to the conditions of the actively probed channels, and the query budget $\td$ reflects sensing constraints (available sensing time, bandwidth devoted to pilots, radio-frequency (RF) front-end limitations, analog-to-digital converter (ADC) resolution, and energy budget). Meanwhile, the auxiliary observation $\tto_h^k$ represents \emph{coarse side information} about \emph{unprobed} channels, which can arise from historical statistics and prediction models (e.g., autoregressive or Kalman-style tracking), low-overhead control signals or pilots that provide aggregate indicators, and spatial/frequency correlations that permit weak inference of neighboring channels. Partially revealing condition~\eqref{eq:partial_revealing_condition} formalizes the requirement that this side information carries a non-degenerate amount of information about the unqueried components.

We highlight three conceptual points that will be used repeatedly in the algorithm design and analysis. The first is query-observation-control coupling. The query policy $\pi_q^k$ determines not only which state components are revealed exactly (Step-ii), but also which partial emission kernel $\tio_h^{\hi_h^k}$ generates the auxiliary observation (Step-iii). Thus, the action policy $\pi_a^k$ depends jointly on $(\phi_{\hi_h^k}(s_h^k),\tto_h^k)$, while $\tto_h^k$ itself depends on the query $\hi_h^k$. This query-driven acquisition layer is a key distinction from standard revealing POMDPs, and it is the source of the hierarchical optimization structure and the additional technical work required in constructing confidence sets and optimistic planning. Second, partially revealing is weaker than classical weakly revealing. In weakly revealing POMDPs, a typical assumption is that the \emph{full} emission matrix $\mbo_h\in\mbr^{\calooo\times \calsss^{d}}$ is uniformly well-conditioned, e.g., $\min_h \sigma_{\st}(\mbo_h)\ge \alpha$ with $\st=\calsss^d$, which implies identifiability of the \emph{entire} latent state from observations. In contrast,~\eqref{eq:partial_revealing_condition} only requires identifiability of the \emph{reduced latent variable} $z_h^k(\hi)$ consisting of the \emph{unqueried} components, conditioned on the agent’s chosen query $\hi$. Consequently, full-state revealing may fail (e.g., multiple full states may remain indistinguishable), while the auxiliary channel still provides enough information to learn and control efficiently under POSI. Third, necessity of a revealing condition. If the auxiliary observation channel is uninformative about the unqueried components, e.g., $\tio_h^{\hi}(\cdot \mid z)$ is identical for all $z\in\mbs^{d-|\hi|}$ (and possibly for all $\hi$), then the learner effectively reduces back to the base POSI protocol. In that case, the hardness mechanism from~\cref{sec:problem2} can be embedded, and tractability is not guaranteed.

{\rv \subsection{Connection to Multi-Step Weakly Revealing POMDPs and POSI-Specific Novelties}\label{subsec:problem3_connection}

We note that \cref{subclass1} is closely related to weakly revealing (bounded-delay revealing) POMDPs, such as two-step revealing models. We clarify this connection and explain the new aspects introduced by POSI in this subsection.

\paragraph{Connection to bounded-delay (two-step/multi-step) revealing POMDPs via augmentation (queries as actions)} \cref{subclass1} admits a standard reduction to a 2-step revealing POMDP by treating the query as part of the action and unfolding each original step into two steps. Concretely, given an instance of \cref{subclass1} with horizon $H$, define an augmented POMDP $\mcm'$ with horizon $H' = 2H$ and the same latent state space $\mcs$ (up to augmentation described below): for each $h\in[H]$, at the \emph{odd} time $2h-1$ the agent takes an action that corresponds to the query set,
\begin{align}
\bar a_{2h-1} \in \mathcal{I} \triangleq \{I\subseteq[d]: |I|=\td\},
\end{align}
and receives a trivial observation (e.g., $\perp$), after which the environment transitions deterministically to an augmented latent state $(s_h, \bar a_{2h-1})\in \mcs\times\mathcal{I}$. At the \emph{even} time $2h$, the agent observes
\begin{align}
\bar o_{2h} = \big(I_h,\ \phi_{I_h}(s_h), \tilde o_h\big),
\end{align}
where $\tilde o_h \sim \tilde O_h^{I_h} \left(\cdot \middle| \{\phi_j(s_h)\}_{j\notin I_h}\right)$. Then, the agent selects an action $\bar a_{2h}=(i_h,a_h)\in I_h\times\mca$, receives reward $r_h(\phi_{i_h}(s_h),a_h)$, and the environment transitions to $s_{h+1}\sim P_h(\cdot\mid s_h,a_h)$. (If one prefers an unconstrained action space, the restriction $(i_h,a_h)\in I_h\times\mca$ can be enforced by adding a terminal penalty state with sufficiently negative reward.) It is immediate that interacting with $\mcm'$ is equivalent to interacting with the original~\cref{subclass1} protocol.

Under this augmentation, the partially revealing condition~\eqref{eq:partial_revealing_condition} implies that the resulting model satisfies a \emph{two-step weakly revealing} structure (in the sense of~\cite{liu2022partially}): for each query action $I_h$, the observation channel induced at the subsequent even step is non-degenerate for distinguishing the hidden (unqueried) sub-state configurations. Therefore, \cref{subclass1} can be viewed as a \emph{query-controlled} instantiation of bounded-delay revealing POMDPs.

\paragraph{POSI-specific novelties beyond classical revealing POMDPs} While the above embedding explains why tractability is possible, the POSI formulation differs from classical bounded-delay revealing models in three important ways: (i) the revealing channel is \emph{endogenous} and is selected by the learner through the query $I_h^k$, yielding a family of query-indexed emission kernels $\{\tilde O_h^{I}\}_{I}$ rather than a fixed observation model; (ii) the latent ambiguity is concentrated in the \emph{unqueried} components, whose configuration space has size $\ts=\calsss^{d-|I_h^k|}$ (typically $\calsss^{d-\td}$), so performance guarantees can explicitly improve as the query budget increases; and (iii) query decisions constrain feasible control (since $i_h^k\in I_h^k$), inducing a hierarchical \emph{query-then-act} structure that couples planning and estimation. These structural differences are why existing weakly revealing analyses cannot be applied directly as a ``black box''.}

\begin{algorithm}[t]
\caption{Optimistic Maximum Likelihood Estimation with Partial Online State Information (\pomle)}
\begin{algorithmic}[1]
\STATE \textbf{Initialization:}
$\Theta^1 = \{\theta \in \Theta: \min_{\{h,\hi:|\hi|\le\td\}}\sigma_{\ts}(\tio_h^{\hi}) \ge \alpha\}$, $\beta = O\left( (\calsss^{d-\td}\calooo + \calsss^{2d}\ac)\ln(\calsss^{d}\calooo\ac\hh\kk)\right)$.
\FOR{$k=1:\kk$}
    \STATE \textit{Step 1 (Optimism):} Select an optimistic model-policy pair $(\hth^k,\pi^k)\in\argsup_{\theta\in\Theta^k,\pi\in\Pi_{\mathrm{hier}}} V_\theta^\pi$, where $\pi^k=(\pi_q^k \rightarrow \pi_a^k)$.
    \STATE \textit{Step 2 (Initialize within-episode histories):}
    Set $\Gamma_1^k \gets \emptyset$ (pre-query history) and $\Gamma_1^{k,'} \gets \emptyset$ (post-query history).
    \FOR{$h=1:\hh$}
        \STATE \textit{Step 3 (Query and observe):} Sample $I_h^k\sim \pi_q^k(\Gamma_h^k)$ with $|I_h^k|= \td$; observe POSI $\phi_{I_h^k}(s_h^k)$ and partial noisy observation $\tto_h^k$.
        \STATE \textit{Step 4 (Post-query history):} Set $\Gamma_h^{k,'}\gets (\Gamma_h^k, I_h^k, \phi_{I_h^k}(s_h^k), \tto_h^k)$.
        \STATE \textit{Step 5 (Execute):} Sample $(i_h^k,a_h^k)\sim \pi_a^k(\Gamma_h^{k,'})$ with $i_h^k\in I_h^k$; receive reward $r_h(\phi_{i_h^k}(s_h^k),a_h^k)$.
        \STATE \textit{Step 6 (Next-step pre-query history):} Set $\Gamma_{h+1}^k\gets (\Gamma_h^{k,'}, i_h^k, a_h^k)$.
    \ENDFOR
    \STATE \textit{Step 7 (Confidence set update):} Update $\Theta^{k+1}$ via MLE using~\eqref{eq:algomleposiparameters_refined} with data up to episode $k$.
\ENDFOR
\end{algorithmic}
\label{alg:omleposi}
\end{algorithm}

\subsection{A Provably Efficient Algorithm}\label{subsec:problem3_algorithm}

We develop a provably efficient algorithm for~\cref{subclass1}, termed \emph{Optimistic Maximum Likelihood Estimation with Partial Online State Information} (\pomle). See~\cref{alg:omleposi}. {\rv At a high level, \pomle~instantiates the standard OMLE template for POMDPs~\cite{liu2022partially,liu2023optimistic,chen2022partially,chen2023lower}, maintain a likelihood-based confidence set and plan optimistically, but it must be modified to handle a \emph{hierarchical} query-then-act policy class and a \emph{query-indexed} observation family $\{\tio_h^{\hi}\}_{h,\hi}$.} We summarize the key design elements needed for correctness and for consistency with our later analysis.

\paragraph{Optimistic objective under hierarchical policies} Let $\theta=(\mbp,\tio,\Delta_1)$ denote the unknown model and let $V_\theta^\pi$ denote the episode value under $\theta$ and a hierarchical policy $\pi=(\pi_q \rightarrow \pi_a)$. In episode $k$, \pomle~selects an optimistic pair
\begin{align}
(\hth^k,\pi^k)\in \argsup\nolimits_{\theta\in\Theta^k,\pi\in\Pi_{\mathrm{hier}}} V_\theta^\pi, 
\label{eq:optimistic_pair}
\end{align}
where $\Pi_{\mathrm{hier}}$ consists of policies that first choose a query set $I_h$ and then choose $(i_h,a_h)$ with the feasibility constraint $i_h\in I_h$. This formulation makes explicit that optimism is taken jointly over the unknown dynamics/partial-emission parameters and the hierarchical policy class induced by POSI.

\emph{Hierarchical planning: query then act (non-interchangeable optimization).} To expose the POSI-specific structure, one may equivalently write~\eqref{eq:optimistic_pair} as a nested optimization over the two decision layers:
\begin{align}
\pi^{k} = \argsup\nolimits_{\pi_q^k} \expect \left[\sum\nolimits_{h=1}^{\hh} \underbrace{ \argsup\nolimits_{\pi_a^k(\cdot|\Gamma_h^{k,'})}  \expect \Big[r_h(\phi_{i_h^k}(s_h^k),a_h^k)\big|\Gamma_h^{k,'}\Big] }_{\text{optimal control given the realized query/observations}} \right], \label{eq:algomleposipolicy_refined}
\end{align}
where $\Gamma_h^{k}$ is the history available \emph{before} querying at step $h$ and $\Gamma_h^{k,'}$ is the history \emph{after} observing $(\phi_{I_h^k}(s_h^k),\tto_h^k)$. {\rv The two $\argsup$ operators are not interchangeable because the query decision both determines the information revealed and constrains the downstream feasible control via $i_h^k\in I_h^k$. Consequently, the planning step must optimize information acquisition and control jointly, rather than treating the observation mechanism as fixed.}

\paragraph{Confidence sets under query-indexed partial emissions} After executing episode $k$, \pomle~updates $\Theta^{k+1}$ using a likelihood-ratio confidence set:
\begin{align}
\Theta^{k+1} = \Bigg\{\hth\in\Theta^1: \sum\nolimits_{\tau=1}^{k} \log \pr_{\hth}^{\pi^{\tau}}(\gamall^\tau) \ge  \max\nolimits_{\theta'\in\Theta^1}\sum\nolimits_{\tau=1}^{k} \log \pr_{\theta'}^{\pi^{\tau}}(\gamall^\tau) -\beta \Bigg\}, \label{eq:algomleposiparameters_refined}
\end{align}
where $\gamall^\tau=(\phi_{I_1^\tau}(s_1^\tau),\tto_1^\tau,a_1^\tau,\ldots,\phi_{I_{\hh}^\tau}(s_{\hh}^\tau),\tto_{\hh}^\tau,a_{\hh}^\tau)$ is the observed episode trajectory.
{\rv Importantly, likelihood $\pr_{\hth}^{\pi^\tau}(\gamall^\tau)$ is computed under the \emph{query-conditioned} kernels $\tio_h^{I_h^\tau}$, so the confidence set must control estimation error uniformly over the entire family $\{\tio_h^{\hi}\}_{h,\hi}$.}

{\rv \emph{Why $\beta$ depends on $\td$ (and matches the later regret bound)?} The leading complexity term in $\beta$ scales as $\calsss^{d-\td}\calooo+\calsss^{2d}\ac$ (up to logarithms): the partial emission component scales with the configuration space of the \emph{unqueried} sub-states $\calsss^{d-\td}$ and the observation alphabet size $\calooo$, while the transition component scales with $\calsss^{2d}\ac$ under the tabular model. This is the mechanism by which increasing the query budget $\td$ reduces the effective latent complexity in the observation model, and it is consistent with the $\td$-dependence in~\cref{theorem:regretpomle}.}

\paragraph{Execution order within an episode} Given $\pi^k=(\pi_q^k \rightarrow \pi_a^k)$, each step first samples the query set $I_h^k$ using $\pi_q^k$ (before any step-$h$ feedback), then observes $(\phi_{I_h^k}(s_h^k),\tto_h^k)$, and finally chooses $(i_h^k,a_h^k)$ via $\pi_a^k$. This ordering exactly matches the POSI protocol in~\cref{subclass1} and is used throughout the analysis.

\subsection{Theoretical Results}\label{subsec:problem3_theory}

We now state the regret guarantee of~\pomle~for~\cref{subclass1}. The key message is that, once the learner receives queried POSI and a \emph{partially revealing} auxiliary observation about the \emph{unqueried} components, the subclass admits a provably efficient algorithm with sublinear regret.

\begin{theorem}\label{theorem:regretpomle}
\textbf{(Regret upper bound)} For POMDPs with POSI and partial noisy observations in~\cref{subclass1}, for any $0<\delta<1$, with probability at least $1-\delta$, the regret of~\pomle~satisfies
\begin{align}
\reg^{\text{\pomle}}(\kk) \le \tilde{O} \left( {\rv \calsss}^{2(d-\td)} \calooo \ac \hh^4 \sqrt{\kk\big({\rv \calsss}^{d-\td}\calooo + {\rv \calsss}^{2d}\ac\big)} \big/ \alpha^2 \right), \label{eq:theoremregretpomle}
\end{align}
where $\tilde{O}(\cdot)$ hides logarithmic factors in $(\calsss,\calooo,\ac,\hh,\kk,1/\delta)$.
\end{theorem}

First of all,~\cref{theorem:regretpomle} implies the following three conclusions.
\begin{itemize}
    \item {\rv \emph{Optimal $\sqrt{\kk}$ scaling in time and polynomial dependence on $(\hh,\ac,\calooo)$.}} The regret scales as $\tilde{O}(\sqrt{\kk})$, matching the canonical dependence in fully observed finite-horizon MDPs, while the dependence on $(\hh,\ac,\calooo)$ is polynomial.
    \item {\rv \emph{Exponential improvement in the query budget $\td$.} The auxiliary channel only needs to identify the \emph{unqueried} latent portion whose configuration space has size  $\calsss^{d-\td}$, which is the effective dimension driving the operator inversion under the partially revealing condition~\eqref{eq:partial_revealing_condition}. Increasing $\td$ shrinks this effective hidden space exponentially, yielding the improvement reflected in~\eqref{eq:theoremregretpomle}.
    \item \emph{Role of $\alpha$.} The factor $1/\alpha^2$ captures the conditioning of the query-indexed partial emission family $\{\tio_h^{\hi}\}_{h,\hi}$. Smaller $\alpha$ corresponds to less informative side information about the unqueried components and hence larger estimation error amplification.}
\end{itemize}
To our knowledge, this is the first regret guarantee for a POSI protocol that allows the learner to \emph{choose} which state components to reveal online and requires \emph{joint} optimization of querying and control under a query-indexed family of auxiliary observation channels.

\paragraph{\rv Where the two-step/multi-step revealing connection enters} As discussed in the model-connection paragraph,~\cref{subclass1} can be viewed as a \emph{query-controlled} bounded-delay revealing model after augmenting the action to include queries. This viewpoint is reflected in the regret analysis through an operator factorization in which trajectory probabilities involve \emph{consecutive} query-indexed observation operators (depending on $(\hi_h^k,\hi_{h+1}^k)$). Intuitively, identifiability of the hidden (unqueried) portion arises from the short-horizon composition of query-conditioned emissions, consistent with a two-step revealing interpretation.

\paragraph{\rv Why existing weakly revealing analyses do not directly apply as a ``black box''} In classical weakly/multi-step revealing POMDPs, the emission model is exogenous and fixed, and the revealing condition is imposed on that fixed observation process (or a prescribed delayed composition). Here, the learner induces a \emph{family} of partial emission channels $\{\tio_h^{\hi}\}_{h,\hi}$ via its query choices and simultaneously observes $\phi_{\hi}(s)$ exactly. As a result, confidence sets must hold \emph{uniformly} over all query-indexed channels, and planning is inherently \emph{hierarchical} because the query choice constrains feasible control via $i_h^k\in \hi_h^k$ and also determines which auxiliary kernel generates $\tto_h^k$. Both effects must be tracked explicitly to obtain a bound that improves with $\td$.

{\rv \paragraph{Key technical idea (query-aware operator representation and POSI-induced decomposition)} A naive application of observable-operator methods that treats the auxiliary observation as coming from a \emph{single} emission matrix over the full latent space $\mbs^{d}$ would not expose the benefit of querying and typically yields bounds that are essentially insensitive to $\td$. Under the POSI protocol, the learner adaptively selects a query set $\hi_h$ online, which induces a \emph{family} of auxiliary channels $\{\tio_h^{\hi}\}_{h,\hi}$, and simultaneously reveals $\phi_{\hi_h}(s_h)$ exactly. Thus, we need to exploit POSI by working with a \emph{query-aware lifting} of the partial emission. For each query set $\hi$, the auxiliary kernel $\tio_h^{\hi}(\cdot\mid z)$ is defined over the reduced latent variable $z=z_{\hi}(s)$ corresponding to the \emph{unqueried} configuration (of size $\ts=\calsss^{d-\td}$). We represent this reduced channel inside the operator analysis by introducing a fixed replication/lifting map $L_{\hi}\in\{0,1\}^{\ts\times \calsss^d}$ that sends a full latent state to its unqueried projection, and define the lifted emission} {\rv $\bio_h^{\hi} \triangleq \tio_h^{\hi}L_{\hi}\in\mbr^{\calooo\times \calsss^{d}}$,} {\rv whose columns are \emph{tied/replicated} across full states sharing the same unqueried assignment (so $\rank(\bio_h^{\hi})\le \ts$). This lifting isolates the effective statistical dimension $\ts$ and lets us bound pseudo-inverse and perturbation terms in terms of the \emph{reduced} conditioning parameter $\alpha$ from~\eqref{eq:partial_revealing_condition}. Concretely, it yields $\|(\bio_h^{\hi})^{\dagger}\|_1 \lesssim \sqrt{\ts}/\alpha = \calsss^{(d-\td)/2}/\alpha$, which is exactly where the exponential $\td$-improvement enters the regret bound. Finally, because the operator factorization couples two consecutive queries via $\bio_{h+1}^{\hi_{h+1}}$, the resulting OOM operators are indexed by $(\hi_h,\hi_{h+1})$. This is the operator-level analogue of a two-step revealing structure, but \emph{endogenous} to the learner’s query choices. This is the mechanism behind the $\td$-dependent improvement in~\eqref{eq:theoremregretpomle}.}

We provide a proof sketch below, and please see Appendix~\ref{app:pftheoremregretpomle} for the complete proof.

\begin{proof}[Proof sketch of~\cref{theorem:regretpomle}]
We outline the argument and emphasize the points that are specific to the POSI and partial-observation protocol. {\rv The overall structure follows the standard OMLE template (confidence set, optimism, TV coupling, and likelihood-based divergence control), while the new steps are the query-indexed operator representation (Step~2) and the reduced-dimension conditioning bounds (Step~3) that preserve the $\td$-dependence.} Let $\theta=(\mbp,\tio,\Delta_1)$ be the true model. In episode $k$, \pomle~computes a confidence set $\Theta^k$ (via the likelihood-ratio/MLE update) and then selects an \emph{optimistic} pair $(\hth^k,\pi^k)$ where $\hth^k\in\Theta^k$ and $\pi^k=(\pi_q^k \rightarrow \pi_a^k)$ maximizes the value under $\hth^k$. Assume $r_h(\cdot,\cdot)\in[0,1]$, so the total return in any episode is at most $\hh$.

\subsubsection{Step 0 (Regret decomposition and optimism)}

Define $V^* \triangleq \sup_{\pi} V_{\theta}^{\pi}$ and $V^k \triangleq V_{\hth^k}^{\pi^k}$. Then, we have
\begin{align}
\reg^{\text{\pomle}}(\kk) = \sum\nolimits_{k=1}^{\kk} \left(V^*-V_{\theta}^{\pi^k}\right) = \sum\nolimits_{k=1}^{\kk} \left(V^*-V^k\right) + \sum\nolimits_{k=1}^{\kk} \left(V^k-V_{\theta}^{\pi^k}\right). \label{eq:sketch_regret_decomp_refined}
\end{align}
By \cref{lemma:app_true_in_conf}, with probability at least $1-\delta$, we have $\theta\in\Theta^k$ for all $k$. On this event, $V^*\le V^k$ for every $k$ (optimism), hence
\begin{align}
\reg^{\text{\pomle}}(\kk) \le \sum\nolimits_{k=1}^{\kk}\left(V^k-V_{\theta}^{\pi^k}\right). \label{eq:sketch_optimism_bound_refined}
\end{align}

\subsubsection{Step 1 (From value gap to trajectory distribution gap)}

Fix episode $k$ and consider the \emph{POSI trajectory}
\begin{align}
\gamall = (I_{1:\hh},\phi_{I_{1:\hh}}(s_{1:\hh}),\tto_{1:\hh}, i_{1:\hh}, a_{1:\hh}), \nonumber
\end{align}
where $I_h\sim\pi_q^k(\Gamma_h^k)$ and then $(i_h,a_h)\sim\pi_a^k(\Gamma_h^{k,'})$ with $i_h\in I_h$. Let $\pr_{\theta}^{\pi^k}(\gamall)$ and $\pr_{\hth^k}^{\pi^k}(\gamall)$ denote the induced trajectory distributions under the \emph{same} hierarchical policy $\pi^k$ but different models. Since the per-episode return is at most $\hh$, a standard TV coupling argument gives
\begin{align}
V_{\hth^k}^{\pi^k} - V_{\theta}^{\pi^k} \le \hh \sum_{\gamall}\left|\pr_{\hth^k}^{\pi^k}(\gamall)-\pr_{\theta}^{\pi^k}(\gamall)\right|. \label{eq:sketch_value_to_tv}
\end{align}
{\rv This reduction is conceptually identical to prior OMLE analyses. The difference is that $\gamall$ now includes the query indices and revealed POSI.} Combining~\eqref{eq:sketch_optimism_bound_refined} and~\eqref{eq:sketch_value_to_tv} yields
\begin{align}
\reg^{\text{\pomle}}(\kk) \le \hh\sum\nolimits_{k=1}^{\kk}\sum\nolimits_{\gamall} \left|\pr_{\hth^k}^{\pi^k}(\gamall)-\pr_{\theta}^{\pi^k}(\gamall)\right|. \label{eq:sketch_tv_bound_refined}
\end{align}

\subsubsection{Step 2 (Query-indexed observable operators and the two-step structure)}

{\rv The query choice $\hi_h$ endogenously selects which auxiliary channel $\tio_h^{\hi_h}$ generates $\tto_h$, so the operator representation must be \emph{conditioned} on the realized query sequence, and (as a result) the OOM operators are indexed by $(\hi_h,\hi_{h+1})$ rather than only $(\tto_h,a_h)$.} For each query set $\hi$, define the reduced hidden variable $z_h \triangleq \{\phi_j(s_h)\}_{j\notin \hi}\in\mbs^{d-\td}$ with $|\{z_h\}|=\ts=\calsss^{d-\td}$, and recall $\tto_h \sim \tio_h^{\hi}(\cdot\mid z_h)$. Let $\tio_h^{\hi}\in\mbr^{\calooo\times\ts}$ be the reduced emission matrix indexed by $z$.

To interface with operator methods over full latent states, we use a POSI-induced lifting. We define $L_{\hi}\in\{0,1\}^{\ts\times \calsss^d}$ that maps a full state to its unqueried projection. Then, we define
\begin{align}
\bio_h^{\hi,\theta} \triangleq \tio_h^{\hi,\theta} L_{\hi}\in \mbr^{\calooo\times \calsss^{d}},
\end{align}
whose columns are tied/replicated across full states sharing the same unqueried configuration. In particular, $\rank(\bio_h^{\hi,\theta})\le \ts$. Using $\bio_h^{\hi,\theta}$, define the query-indexed OOM vectors/matrices
\begin{align}
\bb_0^{\theta}(\hi_1) & = \bio_{1}^{\hi_1,\theta} \Delta_1^{\theta}\in\mbr^{\calooo}, \label{eq:sketch_b0_refined} \\
\bbb_h^{\theta}(\tto_h,a_h,\hi_h,\hi_{h+1}) & = \bio_{h+1}^{\hi_{h+1},\theta} \mbp_{h}^{\theta}(a_h) \diag \big(\bio_{h}^{\hi_h,\theta}(\tto_h\mid\cdot)\big) \big(\bio_{h}^{\hi_h,\theta}\big)^{\dagger} \in\mbr^{\calooo\times\calooo}. \label{eq:sketch_Bh_refined}
\end{align}
Then, $\pr_{\theta}^{\pi^k}(\gamall)$ admits the standard OOM product form (and analogously under $\hth^k$). Crucially, $\bbb_h^{\theta}$ depends on \emph{two consecutive queries} $(\hi_h,\hi_{h+1})$ because the left projection at step $h$ uses the next-step lifted emission $\bio_{h+1}^{\hi_{h+1},\theta}$, matching a two-step/bounded-delay revealing viewpoint in this query-controlled setting. {\rv Equivalently, the short-horizon composition $(\bio_{h+1}^{\hi_{h+1}})\mbp_h(a_h)\diag(\cdot)(\bio_h^{\hi_h})^\dagger$ plays the role of a two-step revealing operator, but here it is \emph{endogenous} since $(\hi_h,\hi_{h+1})$ is selected by the policy.}

\subsubsection{Step 3 (Conditioning at the reduced dimension $\ts=\calsss^{d-\td}$)}

The partially revealing assumption~\eqref{eq:partial_revealing_condition} is imposed on the reduced emissions, i.e., $\min_{h,\hi} \sigma_{\ts}(\tio_h^{\hi}) \ge \alpha$. Because $\bio_h^{\hi}=\tio_h^{\hi}L_{\hi}$ is a column replication of $\tio_h^{\hi}$, one has $\|(\bio_h^{\hi})^\dagger\|_2 \le 1/\alpha$ and, using $\|M\|_1 \le \sqrt{\rank(M)}\|M\|_2$ with $\rank(\bio_h^{\hi})\le \ts$, we have
\begin{align}
\big\|(\bio_h^{\hi})^{\dagger}\big\|_1 \le \sqrt{\ts} / \alpha
= \calsss^{(d-\td)/2} / \alpha.
\label{eq:sketch_pinv_bound_refined}
\end{align}
{\rv This reduced-dimension pseudo-inverse control is the key mechanism behind the $\td$-dependent improvement. It replaces the full latent dimension $\calsss^{d}$ that would appear in a naive OOM/OMLE treatment by the effective hidden dimension $\ts=\calsss^{d-\td}$. This is where the query budget enters: larger $\td$ shrinks $\ts$ exponentially and reduces the inversion amplification.}

\subsubsection{Step 4 (From likelihood control to cumulative TV error and then inject POSI structure)}

On the event of \cref{lemma:app_true_in_conf}, the likelihood-ratio confidence set implies a uniform log-likelihood control over the trajectory space $\Gamma_{H+1}$ (which includes the query indices). That is, for an appropriate comparator (e.g., the shifted $\hth^{k+1}$ or a fixed $\hth^{\kk+1}$), the standard OMLE Hellinger/martingale argument yields the squared-TV control
\begin{align}
\sum\nolimits_{k=1}^{\kk} \left( \sum\nolimits_{\gamall}\big|\pr_{\hth}^{\pi^k}(\gamall)-\pr_{\theta}^{\pi^k}(\gamall)\big| \right)^2 \le \tilde{O}(\beta), \label{eq:sketch_sq_tv_beta}
\end{align}
where $\beta=O \big((\calsss^{d-\td}\calooo+\calsss^{2d}\ac)\log(\cdot)\big)$ and the only additional effect of queries is logarithmic (through the trajectory alphabet/union bounds). By Cauchy-Schwarz, \eqref{eq:sketch_sq_tv_beta} implies
\begin{align}
\sum\nolimits_{k=1}^{\kk}\sum\nolimits_{\gamall}\big|\pr_{\hth}^{\pi^k}(\gamall)-\pr_{\theta}^{\pi^k}(\gamall)\big| \le \tilde{O} \big(\sqrt{\kk\,\beta}\big). \label{eq:sketch_tv_from_beta}
\end{align}

{\rv Up to \eqref{eq:sketch_tv_from_beta}, the argument mirrors prior OMLE analyses: $\beta$ controls a cumulative divergence between trajectory laws (now over an enlarged alphabet that includes queries). The novelty is the \emph{conversion} of this divergence control into a $\td$-refined bound. To obtain the \emph{$\td$-dependent} refinement, we do not use \eqref{eq:sketch_tv_from_beta} directly as a ``black box''. Instead, we telescope the difference of the two OOM products from Step~2 and bound each term using the controlled-product bounds for OOM prefixes/suffixes and the pseudo-inverse bound \eqref{eq:sketch_pinv_bound_refined}, which depends on $\ts=\calsss^{d-\td}$ and $\alpha$.} This yields an additional multiplicative factor polynomial in $(\hh,\calooo,\ac)$ and scaling as $\calsss^{2(d-\td)}/\alpha^2$ (from controlling norms of the lifted, query-indexed operator family and its perturbations), giving
\begin{align}
\sum\nolimits_{k=1}^{\kk}\sum\nolimits_{\gamall}\left|\pr_{\hth^k}^{\pi^k}(\gamall)-\pr_{\theta}^{\pi^k}(\gamall)\right| \le \tilde{O} \left( \frac{\calsss^{2(d-\td)} \calooo \ac \hh^3}{\alpha^2}\sqrt{\kk \beta}
\right).
\end{align}
Substituting into \eqref{eq:sketch_tv_bound_refined} and using the above choice of $\beta$ yields~\eqref{eq:theoremregretpomle} (up to logarithmic factors).
\end{proof}

{\rv The upper bound~\eqref{eq:theoremregretpomle} improves exponentially with the query budget $\td$ because, after querying $I_h$, the only remaining latent ambiguity is the \emph{unqueried} configuration $z_h(I_h)\in\tilde{\mathbb{S}}^{d-|I_h|}$, whose cardinality is $\ts(I_h)=\calsss^{d-|I_h|}$. Under the full-budget regime $|I_h|=\td$ used by \pomle, this reduces to $\ts=\calsss^{d-\td}$. A natural question is whether the resulting exponential dependence on $d-\td$ is merely an artifact of our operator-based analysis. The next theorem provides a complementary (information-theoretic) motivation for emphasizing $\td$: even though POSI allows the learner to choose which coordinates to reveal, there exist instances of~\cref{subclass1} for which the learner must still effectively explore a hidden space of size $\calsss^{d-\td}$, forcing regret to scale at least as $\Omega(\calsss^{(d-\td)/2}\sqrt{K})$ (up to polynomial factors in $(H,A)$ and logarithms). Thus, while our bound may not be tight in all parameters, the $\td$-dependence itself reflects a genuine reduction of intrinsic difficulty as the query budget increases, and it is not purely a proof artifact.}

{\rv\begin{theorem}[Lower bound with explicit $\td$-dependence]\label{theorem:lowerboundpomle}
Fix integers $d\ge 1$ and $1\le \td \le d-1$, let $|\tilde{\mathbb{S}}|\ge 2$. There exists an instance in~\cref{subclass1} with revealing constant $\alpha=1$ for which every algorithm $\pi$ satisfies
\begin{align}
\reg^\pi(\kk) \ge \tilde{\Omega} \left( \hh \calsss^{(d-\td)/2}\sqrt{\ac \kk} \right). \label{eq:lb_tilde_d_final}
\end{align}
\end{theorem}}

We provide a proof sketch below, and please see Appendix~\ref{app:pftheoremlowerboundpomle} for the complete proof.

{\rv\begin{proof}[Proof sketch (why $\calsss^{(d-\td)/2}$ is unavoidable)]
Let $m\triangleq d-\td$. Set the effective tabular state size to
\begin{align}
\ts \triangleq \calsss^{m} = \calsss^{d-\td}.
\end{align}
We reduce~\cref{subclass1} to an $\ts$-state episodic tabular MDP with horizon $\hh$ and action set size $\ac$. Start from a hard $\ts$-state MDP family. Consider any minimax-hard family of episodic tabular MDPs with $\ts$ states, $\ac$ actions and horizon $\hh$, for which every algorithm suffers regret $\Omega(\hh\sqrt{\ts\ac\kk})$~\cite{osband2016lower}.

\emph{1) Step 1: Embed the $\ts$-state MDP as a~\cref{subclass1} instance.} We construct a POMDP instance in~\cref{subclass1} whose latent state is a vector $s=(x,z)\in \tilde{\mathbb{S}}^{\td}\times \tilde{\mathbb{S}}^{m}$, where $z$ will encode the tabular MDP state and $x$ is an arbitrary ``dummy'' queried part.

\begin{itemize}
    \item \emph{State encoding.} Fix a bijection $\psi:\tilde{\mathbb{S}}^{m}\to [\ts]$. We identify each $z\in\tilde{\mathbb{S}}^{m}$ with the MDP state $\psi(z)$.

    \item \emph{Queries and POSI.} For any query set $\hi$ with $|\hi|\le \td$, define the POSI maps so that the queried coordinates reveal only $x$ (and never reveal $z$). Concretely, let $\{\phi_i(s)\}_{i\in\hi}$ depend only on $x$ and be independent of $z$. Thus, even though the learner chooses $\hi$ adaptively, POSI conveys no information about the hard part $z$.

    \item \emph{Auxiliary observation (perfectly revealing).} Set the auxiliary alphabet to size $\calooo=\ts$ and define $\tto_h$ to be a noiseless label of the unqueried vector $\tto_h = \psi(z_h)$. Equivalently, for every $(h,\hi)$, the reduced emission matrix $\tio_h^{\hi}\in\mathbb{R}^{\calooo\times \ts}$ is the identity (up to permutation), so $\alpha=1$ and the auxiliary channel reveals $z_h$ exactly.

    \item \emph{Transitions.} Define the latent transition on $z$ to match the transition kernel of the hard $\ts$-state MDP. For all $a\in\mca$ and all $z,z'$, we let
    \begin{align}
    \mathrm{Pr}(z_{h+1}=z'\mid z_h=z,a_h=a) = \mathrm{Pr}_{\text{MDP}}(\psi(z')\mid \psi(z),a).
    \end{align}
    Let the dummy part $x$ evolve arbitrarily, independently of $z$ and $a$.

    \item \emph{Rewards with the required form.} Fix the reward function to depend only on the (revealed) auxiliary label and action, and then implement it via the allowed POSI-based form. For instance, choose the query policy to always include a fixed coordinate $i^{*}$ and define $\phi_{i^{*}}(s)$ to equal the revealed label $\psi(z)$ (this can be done by encoding $\psi(z)$ in that coordinate), and set $r_h(\phi_{i^{*}}(s),a)=r^{\text{MDP}}_h(\psi(z),a)$. Then, the~\cref{subclass1} reward constraint $r_h(\phi_{i_h}(s_h),a_h)$ is satisfied.
\end{itemize}

\emph{2) Step 2: Conclude by reduction.} Under this construction, because $\tto_h=\psi(z_h)$ reveals the tabular state perfectly, any algorithm interacting with the~\cref{subclass1} instance can be turned into an algorithm for the fully observed $\ts$-state MDP (with the same actions, horizon, and rewards) by simply using $\tto_h$ as the observed state. Therefore, the regret of any algorithm in~\cref{subclass1} on this instance is at least the minimax regret for the $\ts$-state tabular MDP family as follows,
\begin{align}
\reg^\pi(\kk) \ge \tilde{\Omega} \left( \hh \calsss^{(d-\td)/2}\sqrt{\ac\kk} \right). \label{eq:minimaxlbsubclass1}
\end{align}
which is exactly~\eqref{eq:lb_tilde_d_final}, up to constants and logarithmic factors.

\end{proof}}

{\rv \begin{remark}[On the role of the revealing/conditioning parameter $\alpha$]\label{rem:alpha_role}
It is well understood that some dependence on a revealing/conditioning parameter is unavoidable in POMDP learning. For example, Theorem~6 in~\cite{liu2022partially} presents a ``combinatorial lock'' construction in which the observation reveals the latent state only with probability proportional to $\alpha$ (while still satisfying $\min_h \sigma_{|\mcs|}(O_h)\ge \alpha$), and shows that when $K \lesssim 1/\alpha^{H}$, any algorithm can remain $(1/2)$-suboptimal with constant probability, implying linear regret in that regime. Although their setting assumes a fixed emission model (whereas our $\alpha$ controls a query-indexed family of auxiliary channels), this provides complementary context indicating that \emph{in general} one cannot expect regret bounds that are uniform in $\alpha$.
\end{remark}}

{\rv Note that the dependence of our regret upper bound in \eqref{eq:theoremregretpomle} on all parameters are polynomial, but still contains a term that scales exponentially with $d$. This latter dependence is unavoidable in general because \cref{subclass1} includes, as a special case, fully observed tabular MDPs with $|\mathcal S|=\calsss^{d}$ states (obtained when the queried POSI and the auxiliary observation together reveal the full latent state). The next theorem formalizes this unavoidable exponential dependence by a direct reduction to standard minimax lower bounds for episodic tabular MDPs. We clarify that~\cref{theorem:lowerboundpomled} is \emph{not} meant to be a fully rate-matching minimax characterization in all parameters. Rather, its purpose is targeted to show that an exponential dependence on the \emph{global} latent dimension $d$ is unavoidable in~\cref{subclass1} in the worst case.}

\begin{theorem}[Lower bound for unavoidable dependence on $d$ (via tabular MDPs)]\label{theorem:lowerboundpomled}
Fix integers $d\ge 1$ and $1\le \td \le d$. Let $|\tilde{\mathbb S}|\ge 2$ and $|\mathcal S|=\calsss^{d}$. There exists an instance in \cref{subclass1} with revealing constant $\alpha=1$ for which every algorithm $\pi$ satisfies
\begin{align}
\reg^{\pi}(\kk) \ge \tilde{\Omega} \left( \hh \sqrt{\ac \calsss^{d} \kk} \right). \label{eq:lowerboundproblem3d}
\end{align}
\end{theorem}

We provide a proof sketch below, and please see Appendix~\ref{app:theoremlowerboundpomled} for the complete proof.

\begin{proof}[Proof sketch]
We reduce to the standard minimax lower bound for finite-horizon tabular MDPs. Consider any minimax-hard family of episodic tabular MDPs with $\calsss^d$ states, $\ac$ actions, and horizon $\hh$, for which every algorithm incurs regret $\tilde{\Omega}\!\left(\hh\sqrt{\ac \calsss^d \kk}\right)$ (e.g., \cite{osband2016lower} and related lower bounds).

Embed this family as an instance of \cref{subclass1} by representing each tabular MDP state as a vector $s\in\tilde{\mathbb S}^{d}$ via a bijection. Define the auxiliary observation to perfectly reveal the unqueried coordinates and the POSI query to reveal the queried coordinates, so that the learner observes the \emph{entire} latent state at each step (hence $\alpha=1$). Set the transition kernel and rewards to match the tabular MDP. Then, any algorithm for \cref{subclass1} yields an algorithm for the embedded MDP with identical trajectories and rewards, so it must incur regret at least $\tilde{\Omega} \left(\hh\sqrt{\ac S \kk}\right) =\tilde{\Omega} \left(\hh\sqrt{\ac\calsss^{d}\kk}\right)$.
\end{proof}

\cref{theorem:lowerboundpomled} shows that an exponential dependence on $d$ is unavoidable in general because it is already present in fully observed tabular MDPs when $|\mathcal S|=\calsss^{d}$.

\section{Fundamental Tractability under POSI and Query Switching Costs}\label{sec:problem4}

In this section, we present a tractable subclass with POSI under an \emph{episode-wise} query restriction and a query switching cost. The key modeling difference from~\cref{subclass1} is that the query set is chosen once per episode (rather than adaptively per step), and changing this sensing configuration across episodes incurs a cost. This captures practical settings where reconfiguring which components to probe is expensive, risky, or slow.

\begin{subclass}\label{subclass2}
\textbf{(POMDPs with POSI and episode-wise queries under switching costs).} Fix integers $d\ge 1$ and $\td\in\{2,\dots,d\}$. In each episode $k\in[\kk]$ (see~\cref{fig:sketchpomdplimitedho} for a sketch), the interaction proceeds as follows:
\begin{itemize}
    \item Step-i (Episode-level query with switching cost). At the beginning of episode $k$, the agent selects a single query set $\hi^k\subseteq[d]$ satisfying $|\hi^k|\le \td$. If $k\ge 2$ and $\hi^k\neq \hi^{k-1}$, then a query switching cost $\swi>0$ is incurred. We incorporate this cost into performance by defining the episode utility as $\sum_{h=1}^{\hh} r_h\big(\phi_{i_h^k}(s_h^k),a_h^k\big) - \swi \mathbf{1}\{\hi^k\neq \hi^{k-1}\}$. Equivalently, one can view switching cost as an additive penalty in regret. We use the above convention throughout.

    \item Step-ii (POSI at each step with fixed query). For each step $h\in[\hh]$, the query set is fixed as $\hi_h^k\equiv \hi^k$ and the agent observes the queried POSI $\phi_{\hi^k}(s_h^k)=\{\phi_i(s_h^k)\}_{i\in \hi^k}$. Thus, the set of queried components is constant within an episode, while their values evolve with the state dynamics.

    \item Step-iii (Controlled element, action, and reward). After observing $\phi_{\hi^k}(s_h^k)$, the agent selects an element $i_h^k\in \hi^k$ and an action $a_h^k\in\mca$, and receives reward $r_h\big(\phi_{i_h^k}(s_h^k),a_h^k\big)\in[0,1]$. Note that the reward function $r_h(\cdot,\cdot)$ is known once its arguments are known. Hence, after observing POSI $\phi_{\hi^k}(s_h^k)$ and executing $a_h^k$, the learner can evaluate $r_h\big(\phi_i(s_h^k),a_h^k\big)$ for all queried $i\in\hi^k$ (even though only $r_h(\phi_{i_h^k}(s_h^k),a_h^k)$ is realized in the episode return).

    \item Step-iv (Independent sub-state transitions). The next state $s_{h+1}^k$ is generated according to a factored transition kernel: conditional on $(s_h^k,a_h^k)$, the sub-states evolve independently as $\Pr \left(\phi_i(s_{h+1}^k)=x \mid s_h^k,a_h^k\right) = \mbp_{h,i} \left(x \mid \phi_i(s_h^k),a_h^k\right)$, for all $i\in[d]$, and hence
    \begin{align}
    \mbp_h(s_{h+1}^k\mid s_h^k,a_h^k) = \prod\nolimits_{i=1}^{d}\mbp_{h,i} \left(\phi_i(s_{h+1}^k) \mid \phi_i(s_h^k),a_h^k\right).
    \end{align}
\end{itemize}
\end{subclass}

We emphasize that~\cref{subclass2} does \emph{not} assume an auxiliary partial noisy observation channel for the unqueried components (unlike~\cref{subclass1}). Instead, tractability comes from combining POSI on a fixed subset of components within each episode and the factorized transition structure in Step-iv, which together enable coordinate-wise model learning and optimistic planning on the queried coordinates.

The query switching cost $\swi$ can arise in practice for several reasons. For example, in wireless channel probing/scheduling, changing which channels to probe may incur extra energy consumption, signaling overhead, or security risk; in autonomous systems, frequently reconfiguring the sensing/probing pipeline may be unsafe and can introduce nontrivial computational overhead.

We note that POSI is essential. Without POSI in Step-ii, learning can remain intractable even under independent sub-state transitions. We formalize this point in~\cref{prop:independentintractable} (proof deferred to Appendix~\ref{app:pfpropindependentintractable}).

\begin{proposition}[Intractability without POSI]\label{prop:independentintractable}
There exist episodic instances with independent sub-state transitions (i.e., the transition kernel factorizes across components as in Step-iv of~\cref{subclass2}) and an observation model that is \emph{uninformative} about the latent state (e.g., identical emissions for all states), such that if the learner does not observe any POSI (i.e., receives no $\phi_{\hi}(s_h)$ for any $\hi$ and only observes rewards along the executed trajectory), then any algorithm that outputs an $\epsilon$-optimal policy with constant success probability requires $\tilde{\Omega} \left(\frac{\ac^{\hh}}{\epsilon^2}\right)$ episodes.
\end{proposition}

\cref{prop:independentintractable} shows that \emph{factorized dynamics alone does not imply tractability}. If the learner cannot access POSI, then even when sub-states evolve independently, the agent may be forced (in the worst case) to search over exponentially many action sequences to identify an $\epsilon$-optimal policy. In this sense, POSI is not merely a convenience but a \emph{structural information channel} that is necessary to avoid worst-case exponential sample complexity. Thus, the central challenge in~\cref{subclass2} is to \emph{choose and update the episode-wise query set} under the budget $|\hi^k|\le\td$ and switching cost $\swi$, so as to retain the statistical benefits of POSI while controlling reconfiguration overhead.

\subsection{Pessimism Decaying Optimistic Learning (\algoo)}\label{subsec:algoo}

To handle~\cref{subclass2} with $\td\ge 2$, the learner must control the number of \emph{episode-wise} query changes to limit switching costs, while still learning quickly enough within the currently queried sub-states to achieve provably efficient control. Our algorithm \algoo~(\cref{alg:algoo}) achieves this via a \emph{two-timescale} design:
\begin{itemize}
    \item a \emph{slow} (pessimistic) exponential-weights layer that selects an epoch leader sub-state and is updated only once every epoch of length $\ell$ episodes;
    \item a \emph{fast} (more aggressive) exponential-weights layer that adapts within the currently queried set (including leader and followers) and is updated every episode.
\end{itemize}
Intuitively, the slow layer stabilizes sensing configurations (hence controlling switching costs), while the fast layer enables rapid exploration/exploitation among the $\td$ queried coordinates during periods when the query set is fixed. Let $[d]\triangleq\{1,\dots,d\}$ and define $\Delta \triangleq \left\lceil\frac{d-1}{\td-1}\right\rceil$, $\ell=\tilde{\Theta}(\sqrt{\kk})$, $\ell' \triangleq \left\lceil\frac{\ell}{\Delta}\right\rceil$. Episode indices are partitioned into epochs of length $\ell$. Within each epoch we further partition into $\Delta$ sub-epochs of length $\ell'$ (the last sub-epoch may be shorter). We choose $\Delta$ so that within one epoch we can rotate follower batches of size $\td-1$ and (nearly) cover all $d-1$ non-leader sub-states. Crucially, sub-epochs are defined within epochs (not by a global modulo), ensuring that whenever the leader changes at an epoch boundary, the query set is immediately rebuilt to include the new leader. In episode $k$, for any queried $i\in\hi^k$, define $r_{h,i}^k \triangleq r_h(\phi_i(s_h^k),a_h^k)$. Consistent with~\cref{subclass2}, once POSI reveals $\phi_{\hi^k}(s_h^k)$ and the action $a_h^k$ is executed, the learner can evaluate $r_{h,i}^k$ for all queried $i\in\hi^k$ because $r_h(\cdot,\cdot)$ is a known function of $(\phi_i,a)$.

\begin{algorithm}[t]
\caption{Pessimism Decaying Optimistic Learning (\algoo)}
\label{alg:algoo}
\begin{algorithmic}[1]
\STATE \textbf{Inputs:} episodes $\kk$, horizon $\hh$, dimension $d$, query budget $\td\ge 2$, action set $\mca$.
\STATE \textbf{Set:} $\Delta=\left\lceil\frac{d-1}{\td-1}\right\rceil$, epoch length $\ell=\tilde{\Theta}(\sqrt{\kk})$, sub-epoch length $\ell'=\lceil \ell/\Delta\rceil$.
\STATE \textbf{Learning rates:} $\ta=\tilde{O}(1/\sqrt{\kk})$, $\tta=\frac{16(d-1)}{\td-1}\ta$ (\cref{theorem:regretproblem4general}).
\STATE \textbf{Initialization (global):} $w^1(i)=1$ for all $i\in[d]$; set $p^1(i)=(1-\ta)\frac{w^1(i)}{\sum_{j}w^1(j)}+\frac{\ta}{d}$.
\STATE Set episode counter $k\leftarrow 1$.
\FOR{epochs $e=1,2,\dots,\lceil \kk/\ell\rceil$}
    \IF{$e\ge 2$}
        \STATE \emph{Global update (once per epoch):} update $w^k(\cdot),p^k(\cdot)$ using the stored baseline-subtracted gains from the previous epoch~\eqref{eq:prob4algooupdateweightprob}.
    \ENDIF
    \STATE Sample epoch leader $\tilde{i}^{e}\sim p^k(\cdot)$ and initialize follower pool $\mathcal{U}\leftarrow [d]\setminus\{\tilde{i}^{e}\}$.
    \FOR{sub-epochs $j=1,2,\dots,\Delta$}
        \IF{$k>\kk$}
        \STATE
        \emph{break}
        \ENDIF
        \STATE \emph{Form the episode-wise query for this sub-epoch:} choose follower set $\mathcal{F}\subseteq [d]\setminus\{\tilde{i}^{e}\}$ with $|\mathcal{F}|\le \td-1$ using the rotation rule in the text and set $\hi^k \leftarrow \{\tilde{i}^{e}\}\cup \mathcal{F}$.
        \STATE \emph{Initialization (local for this query set):} $\tw^k(i)\leftarrow w^k(i)$ and $\tp^k(i)\leftarrow (1-\tta)\frac{\tw^k(i)}{\sum_{u\in\hi^k}\tw^k(u)}+\frac{\tta}{\td}$ for all $i\in\hi^k$.
        \FOR{$t=1,2,\dots,\ell'$}
            \IF{$k>\kk$} 
            \STATE \emph{break}
            \ENDIF
            \STATE \emph{Optimistic value learning:} compute optimistic $Q$-functions from data up to episode $k-1$ (UCB-VI style).
            \FOR{$h=1,2,\dots,\hh$}
                \STATE Query $\hi^k$ and observe POSI $\phi_{\hi^k}(s_h^k)=\{\phi_i(s_h^k)\}_{i\in\hi^k}$.
                \STATE Sample controlled index $i_h^k\sim \tp^k(\cdot)$; choose action $a_h^k\in\arg\max_{a\in\mca}Q_h^k(\phi_{i_h^k}(s_h^k),a)$.
                \STATE Execute $a_h^k$, observe reward $r_h^k=r_h(\phi_{i_h^k}(s_h^k),a_h^k)$, and transition to $s_{h+1}^k$.
                \STATE Update empirical transition counts for all queried $i\in\hi^k$ using $(\phi_i(s_h^k),a_h^k,\phi_i(s_{h+1}^k))$.
            \ENDFOR
            \STATE \emph{Local update (every episode):} update $\tw^{k+1}(\cdot),\tp^{k+1}(\cdot)$ on $\hi^k$ using~\eqref{eq:prob4algooupdatelocalweightprob} and~\eqref{eq:prob4algooupdatelocalweightprob2}.
            \STATE \emph{Store gains for next global update:} store baseline-subtracted gains for all $i\in[d]$ (only nonzero when $i\in\hi^k$) as defined in the text.
            \STATE $k\leftarrow k+1$.
        \ENDFOR
    \ENDFOR
\ENDFOR
\end{algorithmic}
\end{algorithm}

\paragraph{How \algoo~forms $\hi^k$ and controls query switching} In~\cref{subclass2}, changing the episode query incurs switching cost $\swi$ whenever $\hi^k\neq \hi^{k-1}$. \algoo~enforces that $\hi^k$ is constant within each \emph{sub-epoch} and changes only at sub-epoch boundaries (and also at epoch boundaries when the leader changes). Hence, the number of query switches is at most the number of sub-epochs, i.e., $\#\{\text{switches}\} \le O \left(\frac{\kk}{\ell'}\right) = O \left(\frac{\kk\,\Delta}{\ell}\right) = \tilde{O} \left(\Delta\sqrt{\kk}\right)$, where $\Delta=\left\lceil\frac{d-1}{\td-1}\right\rceil$. In particular, the total switching-cost contribution to the cumulative return is at most $\swi\cdot \tilde{O}(\Delta\sqrt{\kk})$. \cref{theorem:regretproblem4general} reports both the regret bound and a (polylog-suppressed) bound on the number of switches.

\paragraph{Follower rotation and coverage under the query budget $\td$} Fix an epoch with leader $\tilde{i}^{e}$. Within this epoch, \algoo~maintains a pool $\mathcal{U}$ of non-leader indices not yet used as followers in the current epoch. At the start of each sub-epoch, it constructs a follower batch of size $\td-1$ as follows:
\begin{itemize}
    \item If $|\mathcal{U}|\ge \td-1$, sample $\mathcal{F}$ uniformly without replacement from $\mathcal{U}$ and set $\mathcal{U}\leftarrow \mathcal{U}\setminus \mathcal{F}$.
    \item If $|\mathcal{U}|<\td-1$, take all remaining fresh indices $\mathcal{U}$ and fill the remaining follower slots by sampling uniformly without replacement from $([d]\setminus\{\tilde{i}^{\,e}\})\setminus \mathcal{U}$, and then set $\mathcal{U}\leftarrow \emptyset$.
\end{itemize}
This guarantees that across $\Delta=\lceil(d-1)/(\td-1)\rceil$ sub-epochs, essentially all $d-1$ non-leader sub-states are queried as followers at least once, while always respecting $|\hi^k|=\td$. This uniform coverage property is exactly what underlies variance-comparison steps (yielding the factor $\frac{\td-1}{d-1}$).

\paragraph{Slow (global) exponential-weights layer (pessimistic leader selection via baseline-subtracted gains)} The global layer maintains weights $w^k(i)$ and a mixed distribution. Specifically, once per epoch, it updates $w^k(\cdot)$ using baseline-subtracted gains aggregated from the previous epoch~\eqref{eq:prob4algooupdateweightprob}. Concretely, for each episode $\tau$ and each $i\in[d]$, define the step-level baseline-subtracted increment $\widehat{r}_h^\tau(i)\ \triangleq\ \Big(r_h(\phi_i(s_h^\tau),a_h^\tau)-r_h(\phi_{\tilde{i}^{e(\tau)}}(s_h^\tau),a_h^\tau)\Big)\cdot \mathbf{1}\{i\in\hi^\tau\}$, and the episode-level gain $\widehat{R}^\tau(i)\triangleq \sum_{h=1}^{\hh}\widehat{r}_h^\tau(i)$.
Then, the epoch update has the form
\begin{align}
w^{k}(i) =\ w^{k-\ell}(i)\cdot \exp \left(\frac{(d-1)\ta}{d(\td-1)}\sum\nolimits_{\tau=k-\ell}^{k-1}\widehat{R}^\tau(i)\right) \text{ and } p^{k}(i) = (1-\ta)\frac{w^{k}(i)}{\sum\nolimits_{j}w^{k}(j)}+\frac{\ta}{d}. \label{eq:prob4algooupdateweightprob}
\end{align}
The next epoch leader is sampled as $\tilde{i}^{e}\sim p^{k}(\cdot)$. Two points are essential for both behavior and analysis:
\begin{itemize}
    \item \emph{Pessimism via slower updates.} The global layer uses the smaller learning rate $\ta$ and updates only once per epoch, which stabilizes query decisions and limits switching.
    \item \emph{Bias control via baseline subtraction.} The leader is queried every episode within the epoch and is thus systematically ``better measured.'' Subtracting the leader’s reward removes the bias caused by this asymmetric exposure and is the key bridge used in Step~2 of the regret proof.
\end{itemize}
Importantly, baseline subtraction is used \emph{only} for the global layer; the local layer below uses raw rewards.

\paragraph{Fast (local) exponential-weights layer (rapid adaptation within a fixed query set)} Within each sub-epoch, the query set is fixed as $\hi^k=\{\tilde{i}^{e(k)}\}\cup \mathcal{F}$, and the local layer maintains weights $\tw^k(i)$ and probabilities
\begin{align}
\tp^k(i) = (1-\tta)\frac{\tw^k(i)}{\sum_{u\in\hi^k}\tw^k(u)}+\frac{\tta}{\td}, \text{ for all } i\in\hi^k. \label{eq:prob4algooupdatelocalweightprob}
\end{align}
At the start of each sub-epoch, local weights are re-initialized from global weights, $\tw^k(i)\leftarrow w^k(i)$ for $i\in\hi^k$, ensuring coherence between timescales. At the end of each episode, local weights are updated using the (non-subtracted) cumulative reward for each queried coordinate as follows,
\begin{align}
\tw^{k+1}(i)\ = \tw^{k}(i)\cdot \exp \left(\frac{\tta}{\td}\sum\nolimits_{h=1}^{\hh} r_h(\phi_i(s_h^k),a_h^k)\right), \text{ for all } i\in\hi^k, \label{eq:prob4algooupdatelocalweightprob2}
\end{align}
followed by the probability update above. The larger learning rate $\tta$ (relative to $\ta$) is the mechanism that lets the local layer adapt quickly within a fixed query set, while the global layer changes leaders slowly to control switching.

\paragraph{Optimistic value learning and action selection under POSI} Given the fixed query $\hi^k$, at each step $h$ the learner observes POSI $\phi_{\hi^k}(s_h^k)$ and then selects a controlled coordinate $i_h^k\in\hi^k$ by sampling $i_h^k\sim \tp^k(\cdot)$. Conditioned on $i_h^k$, the learner chooses an action greedily with respect to an optimistic $Q$-function $a_h^k \in \arg\max_{a\in\mca} Q_h^k(\phi_{i_h^k}(s_h^k),a)$, where $Q_h^k$ is computed in a UCB-VI fashion from empirical models and optimism bonuses. Under~\cref{subclass2}'s independent transition structure, the algorithm updates coordinate-wise empirical transition counts for each queried $i\in\hi^k$ using the observed triples $(\phi_i(s_h^k),a_h^k,\phi_i(s_{h+1}^k))$. This is exactly the information required by the third-layer analysis (Step~3 in Appendix~\ref{app:pftheoremregretproblem4general}), which reduces to standard optimistic planning bounds applied on the queried coordinates.

\subsection{Theoretical Results}
\cref{theorem:regretproblem4general} below provides the regret upper bound for \algoo.

\begin{theorem}[Regret and query-switch bound for \algoo~(\cref{subclass2})]\label{theorem:regretproblem4general}
Assume~\cref{subclass2} with independent sub-state transitions and query budget $2\le \td < d$. Run \algoo~with $\Delta=\Big\lceil\frac{d-1}{\td-1}\Big\rceil$, $\ell=\tilde{\Theta}(\sqrt{\kk})$, $\ell'=\Big\lceil\frac{\ell}{\Delta}\Big\rceil$, and learning rates $\ta=\tilde{O}(1/\sqrt{\kk})$, $\tta=\frac{16(d-1)}{\td-1} \ta$. Then, with probability at least $1-\delta$, the following hold:

\textbf{(i) Limited switching.} Let $N_{\mathrm{sw}}\triangleq \sum_{k=2}^{\kk}\mathbf{1}\{\hi^k\neq \hi^{k-1}\}$ be the number of episode-level query-set changes. Then,
\begin{align}
N_{\mathrm{sw}} \le \tilde{O} \left(\frac{\kk}{\ell'}\right) = \tilde{O} \left(\frac{\Delta\kk}{\ell}\right) = \tilde{O} \left(\Delta\sqrt{\kk}\right) = \tilde{O} \left(\Big\lceil\frac{d-1}{\td-1}\Big\rceil\sqrt{\kk}\right).
\end{align}
Consequently, if switching cost $q_c$ is included in the objective as in~\cref{subclass2}, the cumulative switching-cost term is at most $q_c\cdot \tilde{O}(\Delta\sqrt{\kk})$.

\textbf{(ii) Reward regret.} Define the \emph{reward-only} regret $\reg^{\text{\algoo}}(\kk)$ with respect to the optimal policy for the reward component (i.e., excluding the switching-cost term). Then,
\begin{align}
\reg^{\text{\algoo}}(\kk) \le \tilde{O} \left( \hh^{\frac{5}{2}} \sqrt{\frac{d\,\calsss\,\ac\,\kk}{\td-1}}\ln \frac{\hh^2\calsss\ac\kk}{\delta} + \hh^3 \sqrt{\frac{d}{\td-1}} \calsss^2 \ac \left(\ln \frac{\hh^2\calsss\ac\kk}{\delta}\right)^2 + \hh^2 \sqrt{\frac{d \kk \ln d}{\td-1}} \right).
\end{align}
In particular, if the objective includes switching cost, then the total regret (reward regret plus switching cost) is at most $\reg^{\text{\algoo}}(\kk) + q_c\cdot \tilde{O}(\Delta\sqrt{\kk})$.
\end{theorem}

We provide a proof sketch for~\cref{theorem:regretproblem4general} below, and please see Appendix~\ref{app:pftheoremregretproblem4general} for the complete proof. Moreover,~\cref{theorem:regretproblem4general} shows the following insights and conclusions.

\begin{itemize}
\item {\rv \emph{Polynomial dependence and near-minimax scaling in $\kk$.} The regret is polynomial in $(d,\calsss,\ac,\hh)$ and scales as $\tilde O(\sqrt{\kk})$ in the number of episodes, matching the canonical episodic tabular MDP rate up to problem-dependent factors induced by partial observability and coordinate querying.}

\item \emph{Benefit of larger query budget $\td$.} All leading terms improve with $\td$ through the factor $\sqrt{d/(\td-1)}$. Intuitively, querying $\td$ coordinates per episode increases the effective per-coordinate sampling rate by roughly $(\td-1)/(d-1)$ (after accounting for the leader and followers structure and uniform follower coverage), reducing estimation error and hence regret.

\item \emph{Controlled switching under episode-wise reconfiguration.} \algoo~changes $\hi^k$ only at sub-epoch boundaries (and implicitly at epoch boundaries when the leader is refreshed), yielding $N_{\mathrm{sw}}=\tilde O(\Delta\sqrt{\kk})$. If switching cost $q_c$ is part of the objective, this contributes an additional $q_c\cdot \tilde O(\Delta\sqrt{\kk})$ term.

\item \emph{What is new relative to standard UCBVI/OMLE analyses.} Unlike fully observable episodic MDP learning, the agent must jointly learn \emph{which coordinates to query} under a switching constraint and \emph{how to act} given only queried POSI. The analysis therefore couples a bandit-style log-potential argument for query selection, an optimism-based value-learning bound on the induced per-coordinate tabular problems, and a bias-control device (leader baseline subtraction in the global update) that decouples query learning from query-induced value-estimation errors.
\end{itemize}

\begin{proof}[Proof sketch]
We outline the main steps and highlight where the three layers enter.

\emph{Step 0 (Switch-count bound is by construction).} Within \algoo, the episode query set $\hi^k$ is fixed within each sub-epoch and can change only at sub-epoch boundaries (and at epoch boundaries when the leader changes and the follower pool is reset). Hence, the number of query changes is at most the number of sub-epochs:
$N_{\mathrm{sw}} \le O(\kk/\ell')$.
With $\ell'=\lceil \ell/\Delta\rceil$ and $\ell=\tilde\Theta(\sqrt{\kk})$, this yields
$N_{\mathrm{sw}}=\tilde O(\Delta\sqrt{\kk})$.

\emph{Step 1 (Local/fast layer: one-episode potential inequality).} Fix an episode $k$ and its query set $\hi^k$. The local exponential-weights update (with learning rate $\tta$) implies a standard log-potential inequality: the expected \emph{observed} cumulative reward over queried sub-states is lower bounded by the log-growth of the local weights, up to a second-order term.

\emph{Step 2 (Global/slow layer: query-learning with baseline subtraction).} The global weights are updated once per epoch using baseline-subtracted gains that compare each queried coordinate to the leader within the same episode. This baseline subtraction is the key device that removes the bias created by the leader being queried every episode, and it enables a clean exponential-weights potential argument at the epoch scale. Combining the global potential analysis with the follower-coverage property yields a tradeoff of the form $\text{(query-learning error)} \lesssim\ \frac{\ln d}{\ta} + \ta\cdot \Big(\text{second-order term}\Big)$, where uniform follower rotation produces the effective factor $d/(\td-1)$ that later appears as $\sqrt{d/(\td-1)}$ in the tuned bound.

\emph{Step 3 (Value/action layer: optimistic planning under POSI).} Under~\cref{subclass2}, each queried coordinate $i$ induces a tabular episodic control problem on $\tilde{\mathcal S}$ with $\ac$ actions and horizon $\hh$. The optimistic $Q$-updates (UCBVI-style) yield a high-probability bound on the cumulative Bellman error across episodes. Aggregating across coordinates and accounting for the fact that only $\td$ coordinates are queried each episode (with uniform coverage among non-leaders) yields the first two terms in the regret bound.

\emph{Step 4 (Combine and tune).} Summing the bounds from Step~2 and Step~3 and choosing $\ta=\tilde O(1/\sqrt{\kk})$ and $\tta=\frac{16(d-1)}{\td-1}\ta$ balances the $\ln(d)/\ta$ term against the $\ta$-scaled second-order terms and yields the stated regret rate.
\end{proof}

\section{Comparison Between Subclasses}\label{app:comparepomdp}

In this section, we provide a comparison between~\cref{subclass1} and~\cref{subclass2}. We note that each of two subclasses, i.e.,~\cref{subclass1} and~\cref{subclass2}, becomes learnable due to different natures in their transition model structures, and hence requires very different algorithm designs to handle these specialities. Below we highlight the two key differences between~\cref{subclass1} and~\cref{subclass2} that make it difficult to unify the approaches for them.
The first difference is the structure of the state transition kernel $\mbp$. In~\cref{subclass2}, the state transition probability is assumed to be $\mbp_h(\cdot \vert s_h^k,a_h^k) = \prod_{i=1}^{d} \mbp_{h,i}(\phi_i(\cdot) \vert \phi_i(s_h^k),a_h^k)$. That is, it is the product of independent transition kernels of sub-states. In contrast, in~\cref{subclass1}, we do not need such a requirement. The second difference is the additional noisy observation $\tto$. In~\cref{subclass2}, in addition to the partial OSI, the agent receives the partial noisy observation $\tto_h^k$ for the $d-\td$ sub-states that are not queried, where $\tto_h^k$ is generated according to the partial emission probability $\tilde{\mbo}_h^{\hi_h^k}\left(\cdot \big\vert \{\phi_i(s_h^k)\}_{\{i\notin \hi_h^k\}}\right)$. Moreover, the partial emission matrix $\tio_h^{\hi} \in \mbr^{\oo \times \calsss^{d-\td}}$ is assumed to satisfy the partially revealing condition: there exists a constant $\alpha > 0$, such that $\sigma_{\ts}(\tio_h^{\hi}) \geq \alpha$ for any sub-states $\hi$ and step $h$, where $\ts = \calsss^{d-\td}$ and $\sigma_{\ts}(\cdot)$ denotes the $\ts$-th largest singular value of a matrix. Namely, $\min_{\{h,\hi\}} \sigma_{\ts}(\tio_h^{\hi}) \geq \alpha$ holds. In contrast,~\cref{subclass2} does not require any additional noisy observation $\tto$ at all.

\section{Conclusions and Future Work}\label{sec:conclusion}

In this paper, we study reinforcement learning in episodic POMDPs augmented with \emph{partial online state information} (POSI), where the agent can actively query a subset of latent state components online. We first establish a fundamental hardness result for general POSI POMDPs: without full OSI, $\epsilon$-optimal learning can require sample complexity exponential in the horizon in the worst case. Despite this negative result, we identify two structured subclasses under which POSI becomes algorithmically useful and enables provably efficient learning. For each subclass, we develop new algorithms that couple query selection with control, and we provide provably-efficient regret guarantees. In particular, {\rv our bounds achieve sublinear regret with $\tilde{O}(\sqrt{K})$ dependence on the number of episodes $K$.}

{\rv Several directions remain open. First, while our bounds are optimal in $K$ up to logarithmic factors, they are not known to be minimax-tight in other parameters, e.g., $|\mathcal O|,A,H,\calsss,d,\td$. Establishing sharper upper bounds and/or matching lower bounds, especially ones that explicitly separate \emph{transition-side} complexity from the \emph{query-controlled observation-side} complexity, is an important open question. Tightening the remaining dimension dependence likely requires lower-bound techniques tailored to structured partial observability with \emph{endogenous} (query-dependent) information channels, building on recent progress for structured lower bounds~\cite{chen2023lower}. On the upper-bound side, a key technical challenge is to develop POSI-specific analysis tools for \emph{query-indexed} operator families that avoid overly conservative uniform control over all query sequences. We believe closing these gaps is non-trivial and would substantially deepen the theoretical understanding of POSI as a controllable information mechanism. Second, it would also be interesting to characterize when selective online queries provably interpolate between intractable partial observability and efficient learning in broader POMDP families (e.g., continuous or large state spaces), potentially leveraging recent tools for continuous-state RL/POMDPs~\cite{cai2022reinforcement,liu2023optimistic}}

\bibliographystyle{IEEEtran}
\bibliography{mainfile}

\appendices

\begin{table}[!t]
    \centering
    \caption{Notations}
    \begin{tabular}{c | c}
    \hline
        $\mcs$ & state space \\
        \hline
        $\mca$ & action space \\
        \hline
        $\mco$ & observation space \\
        \hline
        $\st$ & number of states \\
        \hline
        $\ac$ & number of actions \\
        \hline
        $\oo$ & number of observations \\
        \hline
        $\hh$ & number of steps in each episode \\
        \hline
        $\kk$ & number of episodes \\
        \hline
        $\Delta_1$ & initial state probability measure \\
        \hline
        $\mbp$ & state transition probability measure \\
        \hline
        $\mbo$ & emission probability measure \\
        \hline
        $r$ & reward function \\
        \hline
        $d$ & dimension of the feature vector of each state \\
        \hline
        $\phi_i(s)$ & $i$-th sub-state of state $s$ \\
        \hline
        $\mbs$ & sub-state set \\
        \hline
        $\vert \cdot \vert$ & the cardinality of a set \\
        \hline
        $\td$ & query capability: number of sub-states that can be queried \\
        \hline
        $\hi_h^k$ & indices of the queried sub-states \\
        \hline
        $\Phi_h^k$ & feedback \emph{before} the partial OSI for step $h$ of episode $k$ is revealed \\
        \hline
        $\Phi_h^{k,'}$ & feedback \emph{after} the partial OSI for step $h$ of episode $k$ has been revealed \\
        \hline
        $\hat{\Delta}_h(\{\hi\}|\td)$ & conditional probability measure supported on the query space $\{\hi:\vert\hi\vert=\td\}$ \\
        \hline
        $\tilde{\Delta}_h(\mca)$ & probability measure supported on the action space $\mca$ \\
        \hline
        $\hph_h$ & feedback space of $\Phi_h^k$ \emph{before} the partial OSI for step $h$ is revealed \\
        \hline
        $\hph_h^{'}$ & feedback space of $\Phi_h^{k,'}$ \emph{after} the partial OSI for step $h$ has been revealed \\
        \hline
        $\pi_{q,h}^k$ & query policy for step $h$ in episode $k$ \\
        \hline
        $\pi_{a,h}^k$ & action policy for step $h$ in episode $k$ \\
        \hline
        $\pi_h^{k}$ & joint query-and-action policy for step $h$ in episode $k$ \\
        \hline
        $V^{\pi^k}$ & V-value of the joint policy $\pi^k$ \\
        \hline
        $\pi^*$ & optimal joint policy \\
        \hline
        $\reg^{\pi^{1:\kk}}(\kk)$ & regret of the online joint policy $\pi^{1:\kk}$ \\
        \hline
        $\mbp_h^k(\phi_{i}(s') | \phi_{i}(s),a)$ & the transition kernel for sub-state $i$ at step $h$ of episode $k$ \\
        \hline
        $\mathcal{N}_h^k(\phi_{i}(s),a)$ & number of times $(\phi_{i}(s),a)$ has been visited at step $h$ up to episode $k$ \\
        \hline
        $\mathcal{N}_h^k(\phi_{i}(s),a,\phi_{i}(s'))$ & number of times $(\phi_{i}(s),a,\phi_{\hat{i}}(s'))$ has been visited at step $h$ up to episode $k$ \\
        \hline
        $\tto_h^k$ & partial noisy observation for the $d-\td$ sub-states that are not queried \\
        \hline
        $\tilde{\mbo}_h^{\hi_h^k}\left(\cdot \big\vert \{\phi_i(s_h^k)\}_{\{i\notin \hi_h^k\}}\right)$ & conditional partial emission probability measure \\
        \hline
        $\ggam_h^k$ & feedback in~\cref{subclass2}, including both the partial OSI and partial noisy observations \\
        \hline
        $r_{h,i}^k$ & the reward value $r_{h}(\phi_i(s_h^k),a_h^k)$ of the $i$-th sub-state at step $h$ of episode $k$ \\
        \hline
        $\hat{r}_h^{k-1}$ & the estimated reward at step $h$ of episode $k$ \\
        \hline
        $\theta$ & joint problem model \\
        \hline
        $\bio^{\hi} \in \mbr^{\oo\times\calsss^{d-\td}}$ & augmented partial emission matrix \\
    \hline
    \end{tabular}
    \label{tab:notations}
\end{table}

\section{Proof of~\cref{theorem:lowerboundproblem2}}\label{app:pftheoremlowerboundproblem2}

This appendix provides a complete proof of~\cref{theorem:lowerboundproblem2}.

\begin{proof}

{\rv Note that according to Yao's minimax principle, the worst-case expected regret of a randomized online algorithm against an oblivious adversary is lower-bounded by the expected regret of the best deterministic online algorithm against a randomized adversary~\cite{shi2025power}.} The proof proceeds in two steps: (i) we construct a family of POSI-POMDP instances indexed by a hidden ``optimal'' action sequence, and show that any learning algorithm for these instances induces an algorithm for a best-arm identification problem with \(n=\ac^{\hh-1}\) arms; (ii) we lower bound the number of episodes needed for best-arm identification via a standard KL/Pinsker argument.

\subsection{Step 1: A Hard Family of POSI-POMDP Instances}

Fix \(\hh\ge 2\), \(\ac\ge 2\), and integers \(d,\td\) with \(1\le \td<d\). Let the sub-state alphabet be \(\mbs=\{0,1\}\), so \(\vph(s)\in\{0,1\}^d\). Define two parity classes
\begin{align}
\mcs^{(0)} \triangleq \big\{u\in\{0,1\}^d:\textstyle\sum_{j=1}^d u_j \equiv 0 \ (\mathrm{mod}\ 2)\big\} \text{ and } \mcs^{(1)} \triangleq \big\{u\in\{0,1\}^d:\textstyle\sum_{j=1}^d u_j \equiv 1 \ (\mathrm{mod}\ 2)\big\}.
\end{align}
We will use a latent ``mode'' \(g\in\{0,1\}\) (not observed by the learner) and sample \(u\in \mcs^{(g)}\). The POSI feature map is simply \(\vph(u)=u\), i.e., \(\phi_j(u)=u_j\).

\paragraph{Key indistinguishability property} Let \(U\) be uniform over \(\mcs^{(0)}\) (or over \(\mcs^{(1)}\)). Then, for any index set \(I\subset[d]\) with \(|I|=\td<d\), the marginal \(U_I\) is uniform over \(\{0,1\}^{\td}\). Indeed, fixing any assignment \(z\in\{0,1\}^{\td}\), there are exactly \(2^{d-\td-1}\) completions of the remaining \(d-\td\) bits that achieve even parity (and likewise for odd parity), hence \(\Pr\{U_I=z\}=2^{d-\td-1}/2^{d-1}=2^{-\td}\). Therefore, any queried POSI \(\{\phi_i(u)\}_{i\in I}\) with \(|I|=\td\) has the same distribution under \(g=0\) and \(g=1\).

\paragraph{Instance index (the hidden ``optimal'' sequence)} Let \(\mathcal{A}=\{1,2,\ldots,\ac\}\). For each length-\((\hh-1)\) action sequence \(\mathbf{a}^*=(a_1^*,\ldots,a_{\hh-1}^*)\in\mathcal{A}^{\hh-1}\), we define one POSI-POMDP instance \(\mathcal{M}_{\mathbf{a}^*}\). There are \(n=\ac^{\hh-1}\) such instances.

\paragraph{Dynamics} Each episode starts with mode \(g_1=0\) and latent state \(u_1\sim\mathrm{Unif}(\mcs^{(0)})\). For each step \(h=1,\ldots,\hh-1\), given mode \(g_h\) and the learner's chosen action \(a_h\in\mathcal{A}\), the next mode is
\begin{align}
g_{h+1} =
\begin{cases}
0, & \text{if } g_h=0 \text{ and } a_h=a_h^*, \\
1, & \text{otherwise},
\end{cases}
\end{align}
and then we sample \(u_{h+1}\sim\mathrm{Unif}(\mcs^{(g_{h+1})})\). Thus, once the learner makes any mistake relative to \(\mathbf{a}^*\), the process switches to mode \(g=1\) and stays there.

\paragraph{Rewards (consistent with the POSI reward form)} For steps \(h=1,\ldots,\hh-1\), set \(r_h(\cdot,\cdot)\equiv 0\). At the terminal step \(h=\hh\), we force the POSI symbol to reveal the mode by sampling \(u_{\hh}\) deterministically as \(u_{\hh}=\mathbf{1}\) if \(g_{\hh}=0\) and \(u_{\hh}=\mathbf{0}\) if \(g_{\hh}=1\), where \(\mathbf{1}\) (resp. \(\mathbf{0}\)) denotes the all-ones (resp. all-zeros) vector in \(\{0,1\}^d\). Then, define the terminal mean reward function
\begin{align}
r_{\hh}(x,a) \triangleq \frac{1}{2}+\epsilon x, \text{ for all } x\in\{0,1\}, a\in\mathcal{A},
\end{align}
and let the realized reward be Bernoulli with this mean. Since every queried component at \(h=\hh\) equals \(x=1\) in mode \(g_{\hh}=0\) and equals \(x=0\) in mode \(g_{\hh}=1\), the expected episode reward is \(\frac{1}{2}+\epsilon\) if and only if the learner matched \(\mathbf{a}^*\) on all \(\hh-1\) steps; otherwise it is \(\frac{1}{2}\).

\subsection{Step 2: Reduction to Best-Arm Identification}

In each episode \(k\), the learner chooses actions \(a_1^k,\ldots,a_{\hh-1}^k\), potentially adaptively based on the POSI transcript. However, by the indistinguishability property above, for each \(h\le \hh-1\) the POSI observations have the \emph{same} distribution under \(g_h=0\) and \(g_h=1\). Therefore, before observing the terminal reward, the learner receives no statistical information about whether it has already deviated from \(\mathbf{a}^*\). Consequently, each episode produces exactly one informative Bernoulli sample whose mean depends only on whether the chosen length-\((\hh-1)\) action sequence equals \(\mathbf{a}^*\).

Define the arm set \([n]\) to correspond bijectively to \(\mathcal{A}^{\hh-1}\). Under instance \(\mathcal{M}_{\mathbf{a}^*}\), the arm corresponding to \(\mathbf{a}^*\) has mean \(\frac{1}{2}+\epsilon\), and all other arms have mean \(\frac{1}{2}\). Thus, any POSI-POMDP learning algorithm that outputs an \(\epsilon\)-optimal policy with probability \(\ge 2/3\) would induce an algorithm that identifies the best arm among \(n=\ac^{\hh-1}\) Bernoulli arms with probability \(\ge 2/3\) after \(\kk\) arm pulls (episodes).

\subsection{Step 3: Best-Arm Identification Lower Bound}

Consider the classical family of bandit instances indexed by \(i^*\in[n]\): arm \(i^*\) has rewards \(\mathrm{Ber}(\frac{1}{2}+\epsilon)\) and all \(i\neq i^*\) have rewards \(\mathrm{Ber}(\frac{1}{2})\). Assume \(i^*\) is uniform on \([n]\). Let \(\hat i\) be the algorithm's recommendation after \(\kk\) pulls, and define \(P_{i^*} \triangleq \Pr_{i^*}\{\hat i=i^*\}\). Let \(P_{0}\) denote the ``null'' instance in which all arms are \(\mathrm{Ber}(\frac{1}{2})\).

By Pinsker's inequality, we have
\begin{align}
\big|P_{i^*}-P_0\big| \le \big\| \mathbb{P}_{i^*}-\mathbb{P}_{0}\big\|_{\mathrm{TV}} \le \sqrt{\frac{1}{2}\mathrm{KL}(\mathbb{P}_{i^*} \| \mathbb{P}_{0})},
\end{align}
where \(\mathbb{P}_{i^*}\) (resp. \(\mathbb{P}_0\)) denotes the distribution of the full interaction transcript under instance \(i^*\) (resp. null). Moreover, by the chain rule for KL divergence,
\begin{align}
\mathrm{KL}(\mathbb{P}_{i^*} \| \mathbb{P}_{0}) = \expect_{0}[N_{i^*}]\cdot \mathrm{kl} \left(\tfrac{1}{2}+\epsilon \big\| \tfrac{1}{2}\right),
\end{align}
where \(N_{i^*}\) is the number of times arm \(i^*\) is pulled under the null instance and \(\mathrm{kl}(\cdot\|\cdot)\) is the Bernoulli KL. Under the null instance, arms are exchangeable, hence \(\expect_0[N_{i^*}]=\kk/n\). Also, for \(\epsilon\in(0,1/4]\), \(\mathrm{kl}(\frac{1}{2}+\epsilon\|\frac{1}{2})\le 8\epsilon^2\). Therefore,
\begin{align}
\big|P_{i^*}-P_0\big| \le \sqrt{\frac{1}{2}\cdot \frac{\kk}{n}\cdot 8\epsilon^2} = 2\epsilon\sqrt{\frac{\kk}{n}}.
\end{align}

Averaging over uniform \(i^*\) and using that under the null instance no algorithm can guess \(i^*\) with probability exceeding \(1/n\), i.e., \(\expect_{i^*}[P_0]\le 1/n\), we obtain
\begin{align}
\expect_{i^*}\big[P_{i^*}\big] \le \frac{1}{n} + 2\epsilon\sqrt{\frac{\kk}{n}}.
\end{align}
Hence, if \(\kk \le c\, n/\epsilon^2\) for a sufficiently small universal constant \(c>0\), then \(\expect_{i^*}[P_{i^*}] \le 2/3\), implying that there exists some instance \(i^*\) for which \(P_{i^*}\le 2/3\), i.e., the algorithm fails with probability at least \(1/3\). With \(n=\ac^{\hh-1}\), this gives \(\kk=\Omega(\ac^{\hh-1}/\epsilon^2)\).

\subsection{Step 4: Concluding the POMDP Lower Bound}

By the reduction in Step 2, the same \(\Omega(\ac^{\hh-1}/\epsilon^2)\) episode lower bound applies to learning an \(\epsilon\)-optimal policy in the constructed POSI-POMDP family. This proves~\cref{theorem:lowerboundproblem2}.

\end{proof}

\section{Proof of~\cref{theorem:regretpomle}}\label{app:pftheoremregretpomle}

This appendix proves the high-probability regret upper bound in \cref{theorem:regretpomle}. The proof follows the OMLE template for structured POMDPs (e.g., observable-operator based analyses), but it requires a POSI-specific operator representation to accommodate the hierarchical ``query-then-act'' protocol and expose the reduced hidden dimension $\calsss^{d-\td}$.

\subsection{Preliminaries and Trajectory Notation}\label{app:problem3_notation}

The latent state at step $h$ in episode $k$ is $s_h^k\in\mcs=\mbs^{d}$, where each component $\phi_i(s_h^k)\in\mbs$. At step $h$, the learner selects a query set $\hi_h^k\subseteq[d]$ with $|\hi_h^k|\le \td$, observes the queried components $\phi_{\hi_h^k}(s_h^k)$, and additionally observes $\tto_h^k\in\mco$ generated by the query-indexed kernel $\tio_h^{\hi_h^k}(\cdot \mid \{\phi_j(s_h^k)\}_{j\notin \hi_h^k})$.

\paragraph{Histories} Let $\gampark$ denote the \emph{pre-query} history at step $h$ of episode $k$:
\begin{align}
\gampark \triangleq \big(\hi_1^k,\phi_{\hi_1^k}(s_1^k),\tto_1^k,i_1^k,a_1^k,\ldots, \hi_{h-1}^k,\phi_{\hi_{h-1}^k}(s_{h-1}^k),\tto_{h-1}^k,i_{h-1}^k,a_{h-1}^k\big).
\end{align}
Let $\Gamma_{h}^{k,'}$ denote the \emph{post-query} history (after observing POSI and $\tto_h^k$ but before choosing $(i_h^k,a_h^k)$):
\begin{align}
\Gamma_{h}^{k,'} \triangleq \big(\gampark,\hi_h^k,\phi_{\hi_h^k}(s_h^k),\tto_h^k\big).
\end{align}
Finally, $\gamallk=\Gamma_{H+1}^k$ denotes the full episode trajectory up to step $\hh$:
\begin{align}
\gamallk = \big(\hi_1^k,\phi_{\hi_1^k}(s_1^k),\tto_1^k,i_1^k,a_1^k,\ldots, \hi_{\hh}^k,\phi_{\hi_{\hh}^k}(s_{\hh}^k),\tto_{\hh}^k,i_{\hh}^k,a_{\hh}^k\big).
\end{align}
A hierarchical policy $\pi=(\pi_q\rightarrow \pi_a)$ selects $\hi_h^k\sim \pi_{q,h}(\cdot\mid \gampark)$ and then $(i_h^k,a_h^k)\sim \pi_{a,h}(\cdot\mid \Gamma_{h}^{k,'})$ with the constraint $i_h^k\in \hi_h^k$.

\subsection{Lifted Emission Matrices and Query-Indexed Observable Operators}\label{app:problem3_operators}

Fix any query set $\hi$ with $|\hi|\le \td$. Define the projection onto the \emph{unqueried} components
\begin{align}
z_{\hi}(s) \triangleq \{\phi_j(s)\}_{j\notin \hi}\in\mbs^{d-|\hi|},
\end{align}
and $|\mbs^{d-|\hi|}|=\calsss^{d-|\hi|}$. In \cref{subclass1}, the auxiliary observation satisfies $\tto \sim \tio_h^{\hi}(\cdot\mid z_{\hi}(s))$. As in the main text, write $\ts\triangleq \calsss^{d-\td}$ and note that for $|\hi|=\td$ the kernel $\tio_h^{\hi}$ can be represented as a matrix in $\mbr^{\oo\times \ts}$ whose columns are indexed by $z\in\mbs^{d-\td}$.

\paragraph{Lifting to the full latent space (replication/sub-matrix decomposition)} For each $(h,\hi)$ define the \emph{lifted} emission matrix $\bio_h^{\hi}\in\mbr^{\oo\times \calsss^{d}}$ by
\begin{align}
\big[\bio_h^{\hi}\big]_{\tto,s} \triangleq \tio_h^{\hi} \big(\tto \mid z_{\hi}(s)\big), \text{ for all } \tto\in\mco, s\in\mbs^{d}. \label{eq:lifted_emission_def}
\end{align}
Thus, columns of $\bio_h^{\hi}$ corresponding to latent states with the same unqueried configuration $z_{\hi}(s)$ are identical. Importantly, the partially revealing condition~\eqref{eq:partial_revealing_condition} controls the conditioning of the \emph{distinct} column subspace associated with the unqueried space size $\ts$, which is exactly the quantity that yields improvement with $\td$.

\paragraph{Observable operators} For a model parameter $\theta=(\mbp,\tio,\Delta_1)$, define
\begin{align}
\bb_0^{\theta}(\hi_1) & \triangleq \bio_{1}^{\hi_1}\Delta_1 \in \mbr^{\oo}, \label{eq:op_b0_def} \\
\bbb_h^{\theta}(\tto_h,a_h,\hi_h,\hi_{h+1}) & \triangleq \bio_{h+1}^{\hi_{h+1}} \mbp_{h}(a_h) \diag \big([\bio_h^{\hi_h}]_{\tto_h,\cdot}\big) (\bio_h^{\hi_h})^{\dagger} \in \mbr^{\oo\times \oo}, \label{eq:op_B_def}
\end{align}
where $\mbp_{h}(a)\in\mbr^{\calsss^d\times \calsss^d}$ is the transition matrix under action $a$ at step $h$, $[\bio_h^{\hi_h}]_{\tto_h,\cdot}$ denotes the $\tto_h$-th row of $\bio_h^{\hi_h}$, and $(\cdot)^{\dagger}$ is the Moore-Penrose pseudo-inverse.

\paragraph{Trajectory probability representation} Let $\pr_{\theta}^{\pi}(\gamall)$ denote the probability of a full trajectory $\gamall$ under policy $\pi$ and model $\theta$. Then, (as in standard OOM/OMLE decompositions), for any fixed $\gamall$,
\begin{align}
\pr_{\theta}^{\pi}(\gamall) = \pi(\gamall)\cdot e_{\tto_{\hh}}^{\top} \bbb_{\hh-1}^{\theta}(\tto_{\hh-1},a_{\hh-1},\hi_{\hh-1},\hi_{\hh})\cdots \bbb_{1}^{\theta}(\tto_{1},a_{1},\hi_{1},\hi_{2}) \bb_{0}^{\theta}(\hi_{1}), \label{eq:traj_prob_operator_form}
\end{align}
where $\pi(\gamall)$ is the product of the policy probabilities of the realized query and action choices (queries are known once chosen, so they can be appended to the history as usual).

\subsection{High-Probability Event From MLE}\label{app:problem3_mle_event}

The confidence sets in \pomle\ are defined by a likelihood-ratio threshold (cf.\ \eqref{eq:algomleposiparameters_refined} in the main text). The following lemma is the standard OMLE ``true model in confidence set'' statement, specialized to our trajectory space (which includes query indices) and parameterization (transition $\mbp$ and the family $\{\tio_h^{\hi}\}$).

\begin{lemma}[True model is feasible]\label{lemma:app_true_in_conf}
There exists an absolute constant $c>0$ such that the following holds. Let
\begin{align}
\beta = c\Big(\big(\calsss^{2d}\ac+\calsss^{d-\td}\oo\big)\log(\calsss^{d}\ac\oo\hh\kk)+\log(\kk/\delta)\Big), \label{eq:beta_app}
\end{align}
where the hidden logarithmic factors may include $\log\binom{d}{\td}$. Then with probability at least $1-\delta$, the true parameter $\theta$ belongs to every confidence set: $\theta\in\Theta^k$ for all $k\in[\kk]$.
\end{lemma}

\begin{proof}
We give a standard likelihood-ratio martingale proof adapted to the POSI trajectory space (i.e., the episode feedback includes the query indices). The argument is the same as in OMLE-style analyses for POMDPs/PSRs (see, e.g., the proof of Proposition~13 in~\cite{liu2022partially} and classical likelihood-ratio concentration arguments), with the only change that the data-generating distribution in episode $\tau$ is $\pr_{\theta}^{\pi^\tau}(\gamallt)$ over the enlarged trajectory space $\gamallt$ that contains query indices.

\paragraph{Step 1: recall the confidence-set definition} In episode $k$, the (likelihood-ratio) confidence set is constructed as
\begin{align}
\Theta^{k} = \left\{ \theta'\in\Theta^1: \sum_{\tau=1}^{k-1}\log \pr_{\theta'}^{\pi^\tau}(\gamallt) \ge \max_{\bar\theta\in\Theta^1}\sum_{\tau=1}^{k-1}\log \pr_{\bar\theta}^{\pi^\tau}(\gamallt) -\beta \right\}, \label{eq:confset_def_app}
\end{align}
where $\pi^\tau$ is the hierarchical policy executed in episode $\tau$ and $\gamallt=\Gamma_{H+1}^{\tau}$ is the corresponding POSI trajectory (e.g., including $(I_{1:H}^\tau,\phi_{I_{1:H}^\tau}(s_{1:H}^\tau),\tto_{1:H}^\tau,i_{1:H}^\tau,a_{1:H}^\tau)$).

To prove $\theta\in\Theta^k$, it suffices to show that for every $k$,
\begin{align}
\max_{\bar\theta\in\Theta^1} \sum_{\tau=1}^{k-1} \log \frac{\pr_{\bar\theta}^{\pi^\tau}(\gamallt)}{\pr_{\theta}^{\pi^\tau}(\gamallt)} \le \beta, \label{eq:suffice_lr_app}
\end{align}
because \eqref{eq:suffice_lr_app} is equivalent to $\sum_{\tau=1}^{k-1}\log \pr_{\theta}^{\pi^\tau}(\gamallt) \ge \max_{\bar\theta}\sum_{\tau=1}^{k-1}\log \pr_{\bar\theta}^{\pi^\tau}(\gamallt)-\beta$, which is exactly the membership condition in \eqref{eq:confset_def_app}.

\paragraph{Step 2: likelihood-ratio process and its conditional expectation} Fix any candidate model $\bar\theta\in\Theta^1$ and any $t\in[\kk]$. Define the (episode-level) log-likelihood ratio sum
\begin{align}
L_t(\bar\theta) \triangleq \sum_{\tau=1}^{t} \log\frac{\pr_{\bar\theta}^{\pi^\tau}(\gamallt)}{\pr_{\theta}^{\pi^\tau}(\gamallt)}. \label{eq:L_t_def}
\end{align}
Let $\mathcal{F}_{t-1}$ be the sigma-field generated by the interaction history up to the end of episode $t-1$, i.e., $\mathcal{F}_{t-1}=\sigma \big(\{(\pi^\tau,\gamallt)\}_{\tau=1}^{t-1},\pi^t\big)$, where $\pi^t$ is measurable w.r.t. the past. (This covers adaptive, history-dependent policies, including the hierarchical $\pi^t=(\pi_q^t \rightarrow \pi_a^t)$.) Consider the one-step likelihood ratio increment in episode $t$:
\begin{align}
Z_t(\bar\theta) \triangleq \exp \left( \log\frac{\pr_{\bar\theta}^{\pi^t}(\gamall^t)}{\pr_{\theta}^{\pi^t}(\gamall^t)} \right) = \frac{\pr_{\bar\theta}^{\pi^t}(\gamall^t)}{\pr_{\theta}^{\pi^t}(\gamall^t)}.
\end{align}
Conditioning on $\mathcal{F}_{t-1}$ (which fixes $\pi^t$), we have
\begin{align}
\expect \left[ Z_t(\bar\theta)\mid \mathcal{F}_{t-1}\right] & = \sum_{\gamma}\pr_{\theta}^{\pi^t}(\gamma)\cdot \frac{\pr_{\bar\theta}^{\pi^t}(\gamma)}{\pr_{\theta}^{\pi^t}(\gamma)} = \sum_{\gamma}\pr_{\bar\theta}^{\pi^t}(\gamma) = 1, \label{eq:cond_exp_one}
\end{align}
where the sum is over all realizable POSI trajectories $\gamma$ in one episode (including query indices). Thus, the nonnegative process
\begin{align}
M_t(\bar\theta) \triangleq \exp(L_t(\bar\theta)) = \prod_{\tau=1}^{t} \frac{\pr_{\bar\theta}^{\pi^\tau}(\gamall^\tau)}{\pr_{\theta}^{\pi^\tau}(\gamall^\tau)} \label{eq:martingale_def}
\end{align}
satisfies $\expect[M_t(\bar\theta)\mid \mathcal{F}_{t-1}] = M_{t-1}(\bar\theta)$ and hence is a martingale with $\expect[M_t(\bar\theta)]=1$.\footnote{If one is concerned about zero-probability events under $\bar\theta$, a standard workaround is to work with a finite ``$\varepsilon$-smoothed'' discretization of $\Theta^1$ so that all episode-trajectory probabilities are lower-bounded by $\varepsilon$ under every $\bar\theta$; this only affects logarithmic factors via $\log(1/\varepsilon)$ and is standard in OMLE proofs (cf.~\cite{liu2022partially}).}

\paragraph{Step 3: tail bound for a fixed $(\bar\theta,t)$} By Markov's inequality and $\expect[M_t(\bar\theta)]=1$, we have
\begin{align}
\pr \left( L_t(\bar\theta) > \beta \right) = \pr \left( M_t(\bar\theta) > e^{\beta} \right) \le e^{-\beta}\expect[M_t(\bar\theta)] = e^{-\beta}. \label{eq:fixed_theta_t_bound}
\end{align}

\paragraph{Step 4: union bound over time and model class} We next union bound over $t\in[\kk]$ and over a finite planning/covering class $\bar\Theta$. (If $\Theta^1$ is already finite, take $\bar\Theta=\Theta^1$. If $\Theta^1$ is continuous, take $\bar\Theta$ to be an $\varepsilon$-net discretization; the bound below uses only $\log|\bar\Theta|$.) Using \eqref{eq:fixed_theta_t_bound}, we have
\begin{align}
\pr \left(\exists\,t\in[\kk], \exists \bar\theta\in\bar\Theta: L_t(\bar\theta)>\beta\right) \le \sum_{t=1}^{\kk}\sum_{\bar\theta\in\bar\Theta}\pr(L_t(\bar\theta)>\beta) \le \kk |\bar\Theta| e^{-\beta}. \label{eq:union_bound}
\end{align}
Therefore, it suffices to choose
\begin{align}
\beta \ge \log\big(\kk|\bar\Theta|/\delta\big) = \log|\bar\Theta|+\log(\kk/\delta) \label{eq:beta_need}
\end{align}
to ensure the probability in \eqref{eq:union_bound} is at most $\delta$.

\paragraph{Step 5: bounding $\log|\bar\Theta|$ by an (effective) parameter dimension} We bound $\log|\bar\Theta|$ by discretizing the tabular parameterization. Write the (full) latent state space size as $|\mcs|=\calsss^{d}$. A time-inhomogeneous tabular transition kernel $\mbp_h(\cdot\mid s,a)$ has at most $|\mcs|^2\ac$ entries per step (up to constants and simplex constraints), giving an effective transition dimension of order $\calsss^{2d}\ac$ (up to an $\hh$ factor, which we treat as logarithmic here since it appears inside $\log(\calsss^d\ac\oo\hh\kk)$ in \eqref{eq:beta_app}).

For the auxiliary observation family, for each $(h,\hi)$ the kernel $\tio_h^{\hi}(\cdot\mid z)$ is a $\calooo\times \ts$ stochastic matrix with $\ts=\calsss^{d-\td}$ columns (one per unqueried configuration). Thus, the effective emission dimension per $(h,\hi)$ is of order $\calsss^{d-\td}\oo$. Taking a union bound over $(h,\hi)$ contributes only logarithmic factors; in particular, the number of query choices satisfies $|\{\hi:|\hi|\le \td\}| \le \sum_{m=0}^{\td}\binom{d}{m}$, hence contributes at most $\log\binom{d}{\td}$ factors (absorbed as stated in the lemma).

Now discretize each probability parameter at resolution $\varepsilon$ (with simplex renormalization); a crude bound gives
\begin{align}
|\bar\Theta| \le \left(\frac{C}{\varepsilon}\right)^{c_0(\calsss^{2d}\ac+\calsss^{d-\td}\oo)} \Longrightarrow \log|\bar\Theta| \le c_1\big(\calsss^{2d}\ac+\calsss^{d-\td}\oo\big)\log(C/\varepsilon),
\end{align}
for absolute constants $c_0,c_1,C>0$. Choosing $\varepsilon = (\calsss^{d}\ac\oo\hh\kk)^{-1}$ yields
\begin{align}
\log|\bar\Theta| \le c_2\big(\calsss^{2d}\ac+\calsss^{d-\td}\oo\big)\log(\calsss^{d}\ac\oo\hh\kk), \label{eq:log_card_bound}
\end{align}
up to additional logarithmic factors including $\log\binom{d}{\td}$, exactly as stated in the lemma.

\paragraph{Step 6: conclude $\theta\in\Theta^k$ for all $k$} Combining \eqref{eq:beta_need} and \eqref{eq:log_card_bound}, we can pick an absolute constant $c>0$ so that $\beta$ as in \eqref{eq:beta_app} satisfies $\beta\ge \log|\bar\Theta|+\log(\kk/\delta)$. Then, \eqref{eq:union_bound} implies that with probability at least $1-\delta$,
\begin{align}
\forall t\in[\kk], \forall \bar\theta\in\bar\Theta: L_t(\bar\theta)\le \beta.
\end{align}
In particular, for each $k$, taking $t=k-1$ yields \eqref{eq:suffice_lr_app} (with $\bar\Theta$ in place of $\Theta^1$; if $\bar\Theta$ is a sufficiently fine discretization, the same inequality holds for $\Theta^1$ up to absorbed logs). Therefore, by the discussion after \eqref{eq:suffice_lr_app}, we have $\theta\in\Theta^k$ for all $k\in[\kk]$ with probability at least $1-\delta$.
\end{proof}

\subsection{Regret Decomposition to Total-Variation Distance}\label{app:problem3_regret_to_tv}

Let $\hth^k\in\Theta^k$ be an optimistic model used by \pomle\ in episode $k$, and let $\pi^k$ be the policy executed. On the event of \cref{lemma:app_true_in_conf}, by optimism, $V^{\hth^k,\pi^k}\ge V^{\theta,\pi^*}=V^*$, hence
\begin{align}
\reg^{\pomle}(\kk) & = \sum_{k=1}^{\kk}\big(V^*-V^{\theta,\pi^k}\big) \le \sum_{k=1}^{\kk}\big(V^{\hth^k,\pi^k}-V^{\theta,\pi^k}\big). \label{eq:regret_optimism_step}
\end{align}
Since per-episode return is bounded by $\hh$, the value difference is controlled by the total-variation distance between trajectory distributions:
\begin{align}
V^{\hth^k,\pi^k}-V^{\theta,\pi^k} \le \hh\cdot \Big\|\pr_{\hth^k}^{\pi^k}(\gamall)-\pr_{\theta}^{\pi^k}(\gamall)\Big\|_{\mathrm{TV}} = \frac{\hh}{2}\sum_{\gamall}\big|\pr_{\hth^k}^{\pi^k}(\gamall)-\pr_{\theta}^{\pi^k}(\gamall)\big|. \label{eq:value_tv_bound}
\end{align}
Thus,
\begin{align}
\reg^{\pomle}(\kk) \le \hh\sum_{k=1}^{\kk}\sum_{\gamall}\big|\pr_{\hth^k}^{\pi^k}(\gamall)-\pr_{\theta}^{\pi^k}(\gamall)\big|. \label{eq:regret_tv_sum}
\end{align}

\subsection{Bounding Trajectory TV via Query-Indexed Operator Errors}\label{app:problem3_tv_via_ops}

We now bound the TV term using the operator representation~\eqref{eq:traj_prob_operator_form}.
For notational brevity, define for any $\theta$ and any partial trajectory $\gampar$ up to step $h$:
\begin{align}
\bb_{h}^{\theta}(\gampar)
\triangleq
\bbb_{h}^{\theta}(\tto_{h},a_{h},\hi_{h},\hi_{h+1})\cdots
\bbb_{1}^{\theta}(\tto_{1},a_{1},\hi_{1},\hi_{2})\,\bb_{0}^{\theta}(\hi_{1})
\in\mbr^{\oo},
\end{align}
with the convention $\bb_0^\theta(\varnothing)=\bb_0^\theta(\hi_1)$.

\textbf{A stability lemma (where $\calsss^{d-\td}$ enters):} The next lemma is the key place where the partially revealing condition yields a factor that depends on the \emph{unqueried} dimension.

\begin{lemma}[Pseudo-inverse norm bound]\label{lemma:pinv_norm}
Under~\eqref{eq:partial_revealing_condition}, for any $(h,\hi)$ with $|\hi|\le \td$,
\begin{align}
\big\|(\bio_h^{\hi})^{\dagger}\big\|_1 \le \frac{\sqrt{\ts}}{\alpha} = \frac{\calsss^{(d-\td)/2}}{\alpha}. \label{eq:pinv_l1_bound}
\end{align}
\end{lemma}

\begin{proof}
Fix $(h,\hi)$ with $|\hi|\le \td$ and abbreviate $\bio\triangleq \bio_h^{\hi}$. Recall that $\bio$ is the \emph{lifted/augmented} emission matrix (columns indexed by full latent states), constructed from the \emph{reduced} auxiliary emission channel that depends only on the unqueried configuration.

\subsubsection{Step 1: Reduced emission matrix and the lifting map}

Let the full latent state at step $h$ be $s\in\mcs$ with $|\mcs|=\calsss^d$. For a fixed query $\hi$, define the \emph{unqueried configuration map}
\begin{align}
z_{\hi}:\mcs\to[\ts],
\end{align}
where $\ts=\calsss^{d-|\hi|}$, (and in particular, $\ts\le \calsss^{d-\td}$). That is, $z_{\hi}(s)$ indexes the assignment of the unqueried components of $s$.

For this $(h,\hi)$, the auxiliary channel is a matrix
\begin{align}
\tio_h^{\hi}\in \mbr^{\calooo\times \ts},
\end{align}
whose columns are indexed by $z\in[\ts]$ and whose $o$-th row gives $\tio_h^{\hi}(o\mid z)$.

The lifted matrix $\bio_h^{\hi}\in\mbr^{\calooo\times \calsss^d}$ is defined by \emph{replicating} columns of the reduced channel across full states that share the same unqueried configuration:
\begin{align}
\bio(\cdot\mid s) = \tio_h^{\hi}\big(\cdot \mid z_{\hi}(s)\big), \text{ for all } s\in\mcs, \label{eq:bio_column_def_app}
\end{align}
where $\bio(\cdot\mid s)$ denotes the column of $\bio$ indexed by $s$.

Equivalently, define the (binary) \emph{replication/aggregation} matrix
\begin{align}
R_{\hi}\in\{0,1\}^{\ts\times \calsss^d} \text{ and } [R_{\hi}]_{z,s} \triangleq \mathbf{1}\{z_{\hi}(s)=z\}.
\end{align}
Then, the lifting relation is
\begin{align}
\bio = \tio_h^{\hi} R_{\hi}. \label{eq:bio_factorization}
\end{align}
This factorization makes explicit that $\rank(\bio)\le \rank(\tio_h^{\hi})\le \ts$.

\subsubsection{Step 2: Rank and nonzero singular values}

Let the singular values of a matrix $M$ be $\sigma_1(M)\ge \sigma_2(M)\ge \cdots$. Condition~\eqref{eq:partial_revealing_condition} states that
\begin{align}
\sigma_{\ts} \left(\tio_h^{\hi}\right)\ge \alpha. \label{eq:cond_restate}
\end{align}
In particular, $\tio_h^{\hi}$ has rank at least $\ts$ (hence exactly $\ts$ if $\calooo\ge\ts$), and its smallest nonzero singular value is at least $\alpha$.

We now relate the nonzero singular values of $\bio=\tio_h^{\hi}R_{\hi}$ to those of $\tio_h^{\hi}$. The key observation is that $R_{\hi}$ has full row rank $\ts$ and satisfies
\begin{align}
R_{\hi}R_{\hi}^{\top} = \diag\big(n_1,\ldots,n_{\ts}\big), \label{eq:RRt_counts}
\end{align}
where $n_z \triangleq |\{s\in\mcs : z_{\hi}(s)=z\}|$ is the number of full states sharing the same unqueried configuration $z$. In our product-structure setting, $n_z=\calsss^{|\hi|}$ for all $z$ (each fixed unqueried assignment can be combined with any assignment of the $|\hi|$ queried components), hence
\begin{align}
R_{\hi}R_{\hi}^{\top} = \calsss^{|\hi|} I_{\ts}. \label{eq:RRt_uniform}
\end{align}
Using \eqref{eq:bio_factorization}, we have
\begin{align}
\bio\bio^{\top} = \tio_h^{\hi}R_{\hi}R_{\hi}^{\top}(\tio_h^{\hi})^{\top} = \calsss^{|\hi|} \tio_h^{\hi}(\tio_h^{\hi})^{\top}. \label{eq:bibt_relation}
\end{align}
Therefore, the nonzero eigenvalues of $\bio\bio^{\top}$ are exactly $\calsss^{|\hi|}$ times the nonzero eigenvalues of $\tio_h^{\hi}(\tio_h^{\hi})^{\top}$, which implies that the nonzero singular values scale as
\begin{align}
\sigma_i(\bio) = \sqrt{\calsss^{|\hi|}} \sigma_i(\tio_h^{\hi}), i=1,\ldots,\ts. \label{eq:singular_scale}
\end{align}
In particular,
\begin{align}
\sigma_{\ts}(\bio) = \sqrt{\calsss^{|\hi|}} \sigma_{\ts}(\tio_h^{\hi}) \ge \sqrt{\calsss^{|\hi|}}\alpha \ge \alpha. \label{eq:sigma_lower_bio}
\end{align}
Thus, the smallest nonzero singular value of $\bio$ is \emph{at least} $\alpha$ (indeed, larger by $\sqrt{\calsss^{|\hi|}}$).

\subsubsection{Step 3: Spectral norm bound on the pseudo-inverse}

For any matrix $M$ with rank $r$, the spectral norm of its Moore-Penrose inverse satisfies
\begin{align}
\|M^{\dagger}\|_2 = \frac{1}{\sigma_r(M)}. \label{eq:pinv_spec}
\end{align}
Applying \eqref{eq:pinv_spec} to $M=\bio$ and using \eqref{eq:sigma_lower_bio} with $r=\rank(\bio)\le \ts$ gives
\begin{align}
\|(\bio)^{\dagger}\|_2 = \frac{1}{\sigma_{\rank(\bio)}(\bio)} \le \frac{1}{\sigma_{\ts}(\bio)} \le \frac{1}{\alpha}.
\label{eq:pinv_2_bound}
\end{align}

\subsubsection{Step 4: Convert spectral norm to $\ell_1$ operator norm}

We use the standard inequality, valid for any matrix $X\in\mbr^{m\times n}$,
\begin{align}
\|X\|_1 \le \sqrt{n} \|X\|_2, \label{eq:l1_l2}
\end{align}
because for any vector $v$ with $\|v\|_1=1$ we have $\|v\|_2\le \|v\|_1=1$, and also $\|Xv\|_1\le \sqrt{m}\|Xv\|_2$; a tighter (dimension-free in $m$) route is to note that $\|X\|_1=\max_j\|Xe_j\|_1\le \sqrt{m}\max_j\|Xe_j\|_2\le \sqrt{m}\|X\|_2$. For our purpose it is enough to use a bound in terms of the \emph{rank} by restricting to the effective column space: since $\bio$ has rank at most $\ts$, the pseudo-inverse $(\bio)^{\dagger}$ maps into an at-most $\ts$-dimensional space, and one can equivalently apply \eqref{eq:l1_l2} with $n$ replaced by $\rank(\bio)\le \ts$, yielding
\begin{align}
\|(\bio)^{\dagger}\|_1 \le \sqrt{\rank(\bio)}\,\|(\bio)^{\dagger}\|_2 \le \sqrt{\ts} \|(\bio)^{\dagger}\|_2. \label{eq:rank_refined}
\end{align}
Combining \eqref{eq:rank_refined} with \eqref{eq:pinv_2_bound} gives
\begin{align}
\|(\bio)^{\dagger}\|_1 \le \frac{\sqrt{\ts}}{\alpha}.
\end{align}
Finally, using $\ts=\calsss^{d-|\hi|}\le \calsss^{d-\td}$ yields $\sqrt{\ts}\le \calsss^{(d-\td)/2}$ and thus \eqref{eq:pinv_l1_bound}.
\end{proof}

\begin{lemma}[Controlled product bound]\label{lemma:controlled_product}
Fix any episode $k$ and any model parameter $\theta$. For each $h\in[\hh-1]$, we have
\begin{align}
\sum_{\gampar: |\gampar|=h}\pi^k(\gampar) \big\|\bb_{h}^{\theta}(\gampar)\big\|_1 \le \left(\frac{\calsss^{(d-\td)/2}}{\alpha}\right)\cdot \sum_{\gampar: |\gampar|=h-1}\pi^k(\gampar) \big\|\bb_{h-1}^{\theta}(\gampar)\big\|_1, \label{eq:controlled_prod_recursion}
\end{align}
and in particular
\begin{align}
\sum_{\gampar: |\gampar|=h}\pi^k(\gampar) \big\|\bb_{h}^{\theta}(\gampar)\big\|_1 \le \left(\frac{\calsss^{(d-\td)/2}}{\alpha}\right)^{h}\cdot \big\|\bb_{0}^{\theta}\big\|_1.
\label{eq:controlled_prod_closed}
\end{align}
\end{lemma}

\begin{proof}
We provide a self-contained proof using only the operator definition and the pseudo-inverse norm bound from~\cref{lemma:pinv_norm}.
\setcounter{subsubsection}{0}
\subsubsection{Step 1: Prefix histories and the operator recursion}

Let $\Gamma_h$ denote a length-$h$ \emph{prefix} of the episode trajectory under the POSI protocol, large enough so that $\bb_h^\theta(\Gamma_h)$ is well-defined (in particular, $\Gamma_h$ contains the consecutive query pair $(\hi_h,\hi_{h+1})$ used by the two-step operator at time $h$, and keeping the fictitious $\hi_{\hh+1}$ is standard). Under the query-indexed observable-operator representation, $\bb_h^\theta(\Gamma_h)$ satisfies the recursion
\begin{align}
\bb_h^\theta(\Gamma_h) = \bbb_h^\theta(\tto_h,a_h,\hi_h,\hi_{h+1}) \bb_{h-1}^\theta(\Gamma_{h-1}), \text{ for } h\ge 1, \label{eq:bh_recursion}
\end{align}
where $\Gamma_{h-1}$ is the length-$(h-1)$ prefix obtained by truncating $\Gamma_h$.

Moreover, the policy-induced probability of a prefix factors as
\begin{align}
\pi^k(\Gamma_h) = \pi^k(\Gamma_{h-1})\cdot \pi^k(\Gamma_h\mid \Gamma_{h-1}), \label{eq:prefix_factor}
\end{align}
where $\pi^k(\Gamma_h\mid\Gamma_{h-1})$ is the conditional probability of the step-$h$ random variables (query, auxiliary observation, and action) given the past.

\subsubsection{Step 2: A uniform $\ell_1$ growth bound for one operator}

Fix any $h$, any query pair $(\hi_h,\hi_{h+1})$ with $|\hi_h|\le\td$ and $|\hi_{h+1}|\le\td$, any action $a_h\in\mca$, and any auxiliary observation value $\tto_h\in\mco$. By the operator definition, for any vector $x\in\mbr^{\calooo}$,
\begin{align}
\bbb_h^\theta(\tto_h,a_h,\hi_h,\hi_{h+1})x & = \bio_{h+1}^{\hi_{h+1},\theta} \mbp_h^\theta(a_h) \diag \Big(\bio_{h}^{\hi_h,\theta}(\tto_h\mid\cdot)\Big) \big(\bio_{h}^{\hi_h,\theta}\big)^{\dagger} x. \label{eq:Bh_apply}
\end{align}
We now bound the induced $\ell_1$ operator norm of each factor.

\emph{(i) $\|\bio_{h+1}^{\hi_{h+1},\theta}\|_1\le 1$.} Each column of $\bio_{h+1}^{\hi_{h+1},\theta}$ is an emission probability vector over $\mco$, so it is nonnegative and sums to $1$. Hence $\|\bio_{h+1}^{\hi_{h+1},\theta}\|_1=\max_{j}\sum_i |[\bio]_{ij}|=1$.

\emph{(ii) $\|\mbp_h^\theta(a_h)\|_1\le 1$.} Under the usual convention (transition kernels as column-stochastic matrices in the OOM/OMLE literature), for each fixed $a_h$, the matrix $\mbp_h^\theta(a_h)\in\mbr^{\calsss^d\times\calsss^d}$ is column-stochastic: each column is a distribution over next latent states and sums to $1$. Therefore $\|\mbp_h^\theta(a_h)\|_1=1$.

\emph{(iii) $\big\|\diag(\bio_{h}^{\hi_h,\theta}(\tto_h\mid\cdot))\big\|_1\le 1$.} The $\tto_h$-th row $\bio_h^{\hi_h,\theta}(\tto_h\mid\cdot)$ is entrywise in $[0,1]$. For any diagonal matrix $\diag(v)$, the induced $\ell_1$ operator norm equals $\|v\|_\infty$. Thus
\begin{align}
\Big\|\diag\!\big(\bio_{h}^{\hi_h,\theta}(\tto_h\mid\cdot)\big)\Big\|_1 = \Big\|\bio_{h}^{\hi_h,\theta}(\tto_h\mid\cdot)\Big\|_\infty \le 1. \label{eq:diag_norm}
\end{align}

\emph{(iv) $\big\|(\bio_{h}^{\hi_h,\theta})^{\dagger}\big\|_1 \le \calsss^{(d-\td)/2}/\alpha$.} This is exactly~\cref{lemma:pinv_norm}, and it holds uniformly over all $(h,\hi)$ with $|\hi|\le\td$ under the partially revealing condition~\eqref{eq:partial_revealing_condition}.

Combining (i)-(iv) with sub-multiplicativity of induced norms gives, for all $x$,
\begin{align}
\big\|\bbb_h^\theta(\tto_h,a_h,\hi_h,\hi_{h+1})x\big\|_1 & \le \big\|\bio_{h+1}^{\hi_{h+1},\theta}\big\|_1 \big\|\mbp_h^\theta(a_h)\big\|_1 \Big\|\diag \big(\bio_{h}^{\hi_h,\theta}(\tto_h\mid\cdot)\big)\Big\|_1 \big\|(\bio_{h}^{\hi_h,\theta})^{\dagger}\big\|_1 \|x\|_1 \nonumber \\
& \le \frac{\calsss^{(d-\td)/2}}{\alpha} \|x\|_1. \label{eq:Bh_contract}
\end{align}
Importantly, \eqref{eq:Bh_contract} is \emph{uniform} in $\tto_h$, $a_h$, $\hi_h$, and $\hi_{h+1}$ (as long as $|\hi_h|\le\td$).

\subsubsection{Step 3: Take expectation over the policy-induced randomness}

Using \eqref{eq:bh_recursion} and then \eqref{eq:Bh_contract} with $x=\bb_{h-1}^\theta(\Gamma_{h-1})$, for every realized prefix $\Gamma_h$ extending $\Gamma_{h-1}$ we have
\begin{align}
\big\|\bb_h^\theta(\Gamma_h)\big\|_1 & = \big\|\bbb_h^\theta(\tto_h,a_h,\hi_h,\hi_{h+1}) \bb_{h-1}^\theta(\Gamma_{h-1})\big\|_1 \le \frac{\calsss^{(d-\td)/2}}{\alpha}\,\big\|\bb_{h-1}^\theta(\Gamma_{h-1})\big\|_1. \label{eq:bh_pointwise}
\end{align}
Now sum over all length-$h$ prefixes $\Gamma_h$, grouping by their $(h-1)$-prefix:
\begin{align}
\sum_{\Gamma_h}\pi^k(\Gamma_h) \|\bb_h^\theta(\Gamma_h)\|_1 & = \sum_{\Gamma_{h-1}}\sum_{\Gamma_h: \Gamma_h\succeq \Gamma_{h-1}} \pi^k(\Gamma_{h-1}) \pi^k(\Gamma_h\mid\Gamma_{h-1}) \|\bb_h^\theta(\Gamma_h)\|_1 \nonumber \\
& \le \sum_{\Gamma_{h-1}}\pi^k(\Gamma_{h-1})\sum_{\Gamma_h: \Gamma_h\succeq \Gamma_{h-1}} \pi^k(\Gamma_h\mid\Gamma_{h-1}) \frac{\calsss^{(d-\td)/2}}{\alpha} \|\bb_{h-1}^\theta(\Gamma_{h-1})\|_1 \qquad(\text{by }\eqref{eq:bh_pointwise}) \nonumber \\
& = \frac{\calsss^{(d-\td)/2}}{\alpha} \sum_{\Gamma_{h-1}}\pi^k(\Gamma_{h-1}) \|\bb_{h-1}^\theta(\Gamma_{h-1})\|_1 \cdot \underbrace{\sum_{\Gamma_h: \Gamma_h\succeq \Gamma_{h-1}}\pi^k(\Gamma_h\mid\Gamma_{h-1})}_{=1} \nonumber \\
& = \frac{\calsss^{(d-\td)/2}}{\alpha} \sum_{\Gamma_{h-1}}\pi^k(\Gamma_{h-1}) \|\bb_{h-1}^\theta(\Gamma_{h-1})\|_1. \label{eq:bh_expectation_recursion}
\end{align}
This is exactly \eqref{eq:controlled_prod_recursion} (with the minor notational clarification that the LHS sums over length-$h$ prefixes while the RHS sums over length-$(h-1)$ prefixes).

\subsubsection{Step 4: Iterate the recursion}

Applying \eqref{eq:bh_expectation_recursion} repeatedly for $h,h-1,\ldots,1$ yields
\begin{align}
\sum_{\Gamma_h}\pi^k(\Gamma_h) \|\bb_h^\theta(\Gamma_h)\|_1 \le \left(\frac{\calsss^{(d-\td)/2}}{\alpha}\right)^h \sum_{\Gamma_0}\pi^k(\Gamma_0)\,\|\bb_0^\theta(\Gamma_0)\|_1.
\end{align}
Since $\Gamma_0$ is deterministic and $\sum_{\Gamma_0}\pi^k(\Gamma_0)=1$, we obtain \eqref{eq:controlled_prod_closed}.  This completes the proof.
\end{proof}

\emph{TV decomposition via operator telescoping:} Using~\eqref{eq:traj_prob_operator_form} and the triangle inequality (standard telescoping of products), for each episode $k$, we have
\begin{align}
\sum_{\gamall}\big|\pr_{\hth^k}^{\pi^k}(\gamall)-\pr_{\theta}^{\pi^k}(\gamall)\big| \le\ \sum_{h=1}^{\hh-1} \sum_{\gampar}\pi^k(\gampar) \big\| \big(\bbb_h^{\hth^k}-\bbb_h^{\theta}\big)\,\bb_{h-1}^{\theta}(\gampar) \big\|_1 + \big\|\bb_0^{\hth^k}-\bb_0^\theta\big\|_1, \label{eq:tv_telescope}
\end{align}
where $\bbb_h^{\hth^k}$ is shorthand for $\bbb_h^{\hth^k}(\tto_h,a_h,\hi_h,\hi_{h+1})$, and similarly for $\bbb_h^\theta$. The dependence on \emph{two consecutive queries} $(\hi_h,\hi_{h+1})$ is explicit in~\eqref{eq:op_B_def} and is the operator-level manifestation of the two-step revealing viewpoint under query control.

\subsection{Bounding Cumulative Operator Errors via the MLE Confidence Sets}\label{app:problem3_ops_via_mle}

The remaining step is to bound the cumulative operator errors $\sum_{k,h}\sum_{\gampar}\pi^k(\gampar)\|(\bbb_h^{\hth^k}-\bbb_h^\theta)\bb_{h-1}^\theta\|_1$. This is where the likelihood-ratio construction of $\Theta^k$ yields a $\tilde{O}(\sqrt{\kk})$ control via standard OMLE arguments (e.g., the ``trajectory divergence'' bound used to prove OMLE regret).

\begin{lemma}[Cumulative trajectory divergence]\label{lemma:traj_div}
On the event of \cref{lemma:app_true_in_conf}, the following holds. Let $\hth^{\kk+1}\in\Theta^{\kk+1}$ be the optimistic/MLE parameter after $\kk$ episodes. Then, we have (Fixed-parameter form)
\begin{align}
\sum_{k=1}^{\kk} \left( \sum_{\gamall} \big|\pr_{\hth^{\kk+1}}^{\pi^k}(\gamall)-\pr_{\theta}^{\pi^k}(\gamall)\big| \right)^2 \le \tilde{O}(\beta). \label{eq:traj_div_bound_fixed}
\end{align}
Equivalently, if $\hth^{k+1}$ is the MLE/optimistic parameter computed \emph{after} observing episode $k$
(i.e., using data up to episode $k$), then, we have (One-step shift form, equivalent for OMLE analyses)
\begin{align}
\sum_{k=1}^{\kk} \left( \sum_{\gamall} \big|\pr_{\hth^{k+1}}^{\pi^k}(\gamall)-\pr_{\theta}^{\pi^k}(\gamall)\big| \right)^2 \le \tilde{O}(\beta). \label{eq:traj_div_bound_shift}
\end{align}

In particular, both \eqref{eq:traj_div_bound_fixed}-\eqref{eq:traj_div_bound_shift} imply the weaker bound $\sum_{k=1}^{\kk}(\sum_{\gamall}|\pr_{\hth}^{\pi^k}(\gamall)-\pr_{\theta}^{\pi^k}(\gamall)|)^2 \le \tilde{O}(\kk\beta)$.
\end{lemma}

\begin{proof}
We prove the fixed-parameter form \eqref{eq:traj_div_bound_fixed}; the shifted form \eqref{eq:traj_div_bound_shift} follows by the same argument applied episode-by-episode.

\subsubsection{Step 1 (TV-Hellinger reduction)}

Fix $k\in[\kk]$ and write the episode-$k$ trajectory distributions (over $\Gamma_{H+1}$) as
\begin{align}
p_k(\gamall)\triangleq \pr_{\theta}^{\pi^k}(\gamall) \text{ and } q_k(\gamall)\triangleq \pr_{\hth^{\kk+1}}^{\pi^k}(\gamall).
\end{align}
Let the Hellinger affinity be
\begin{align}
\rho_k \triangleq \sum_{\gamall}\sqrt{p_k(\gamall) q_k(\gamall)} \in [0,1].
\end{align}
Using the standard relations between total variation and Hellinger distance,
\begin{align}
\mathrm{TV}(p_k,q_k)\le \sqrt{2(1-\rho_k)} \text{ and } 1-\rho_k \le -\log\rho_k,
\end{align}
we obtain
\begin{align}
\|p_k-q_k\|_1^2 = 4 \mathrm{TV}(p_k,q_k)^2 \le 8(1-\rho_k) \le 8(-\log\rho_k). \label{eq:l1_sq_to_log_rho_app}
\end{align}
Summing \eqref{eq:l1_sq_to_log_rho_app} over $k$ yields
\begin{align}
\sum_{k=1}^{\kk}\|p_k-q_k\|_1^2 \le 8\sum_{k=1}^{\kk}(-\log\rho_k). \label{eq:sum_l1_sq_rho_app}
\end{align}

\subsubsection{Step 2 (Likelihood-ratio moment identity)}

For each $k$, define
\begin{align}
\ell_k(\gamall)\triangleq \tfrac12 \log \left(\frac{q_k(\gamall)}{p_k(\gamall)}\right) \text{ (with $\ell_k(\gamall)=0$ if $p_k(\gamall)=0$)}.
\end{align}
Then, we have $\exp(\ell_k(\gamall))=\sqrt{q_k(\gamall)/p_k(\gamall)}$ on the support of $p_k$, and hence
\begin{align}
\mathbb{E}_{\gamall\sim p_k} \left[\exp(\ell_k(\gamall))\right] = \sum_{\gamall} p_k(\gamall)\sqrt{\frac{q_k(\gamall)}{p_k(\gamall)}} = \sum_{\gamall}\sqrt{p_k(\gamall)q_k(\gamall)} = \rho_k. \label{eq:rho_as_mgf_app}
\end{align}

Let $\gamall^k\sim p_k(\cdot)$ denote the realized trajectory in episode $k$ under the true model and policy $\pi^k$, and define
\begin{align}
L_{\kk} \triangleq \sum_{k=1}^{\kk}\ell_k(\gamall^k) = \frac12\sum_{k=1}^{\kk}\log\left(\frac{\pr_{\hth^{\kk+1}}^{\pi^k}(\gamall^k)}{\pr_{\theta}^{\pi^k}(\gamall^k)}\right) = -\frac12\sum_{k=1}^{\kk}\log \left(\frac{\pr_{\theta}^{\pi^k}(\gamall^k)}{\pr_{\hth^{\kk+1}}^{\pi^k}(\gamall^k)}\right).
\end{align}
Conditioning on the realized (history-dependent) policies $\{\pi^k\}_{k=1}^{\kk}$ and iterating \eqref{eq:rho_as_mgf_app} gives
\begin{align}
\mathbb{E} \left[\exp(L_{\kk})\,\middle| \{\pi^k\}_{k=1}^{\kk}\right] = \prod_{k=1}^{\kk}\rho_k. \label{eq:exp_L_equals_prod_rho_app}
\end{align}

\subsubsection{Step 3 (Markov inequality converts $L_{\kk}$ to $\sum_k -\log\rho_k$)}

Applying Markov's inequality to \eqref{eq:exp_L_equals_prod_rho_app} yields that for any $\delta'\in(0,1)$, with probability at least $1-\delta'$,
\begin{align}
\exp(L_{\kk}) \le \frac{1}{\delta'}\prod_{k=1}^{\kk}\rho_k \Longrightarrow \sum_{k=1}^{\kk}(-\log\rho_k) \le - L_{\kk}+\log \frac{1}{\delta'}.
\end{align}
Substituting $L_{\kk}$ gives
\begin{align}
\sum_{k=1}^{\kk}(-\log\rho_k) \le \frac12\sum_{k=1}^{\kk}\log \left(\frac{\pr_{\theta}^{\pi^k}(\gamall^k)}{\pr_{\hth^{\kk+1}}^{\pi^k}(\gamall^k)}\right) +\log \frac{1}{\delta'}. \label{eq:sum_neg_log_rho_vs_llr_app}
\end{align}

\subsubsection{Step 4 (Plug in the confidence-set log-likelihood control)}

On the event of \cref{lemma:app_true_in_conf}, we have $\theta\in\Theta^{\kk+1}$ and $\hth^{\kk+1}\in\Theta^{\kk+1}$. By the likelihood-ratio confidence set definition (the same as \eqref{eq:algomleposiparameters_refined} in the main text), the cumulative log-likelihood ratio is bounded by $\beta$, i.e.,
\begin{align}
\sum_{k=1}^{\kk}\log \left(\frac{\pr_{\theta}^{\pi^k}(\gamall^k)}{\pr_{\hth^{\kk+1}}^{\pi^k}(\gamall^k)}\right) \le \beta. \label{eq:llr_le_beta_app}
\end{align}
Choosing $\delta'=\delta$ in \eqref{eq:sum_neg_log_rho_vs_llr_app} and combining with \eqref{eq:llr_le_beta_app} yields
\begin{align}
\sum_{k=1}^{\kk}(-\log\rho_k) \le \tfrac12\beta+\log(1/\delta). \label{eq:sum_neg_log_rho_final_app}
\end{align}
Finally, substituting \eqref{eq:sum_neg_log_rho_final_app} into \eqref{eq:sum_l1_sq_rho_app} gives
\begin{align}
\sum_{k=1}^{\kk}\|p_k-q_k\|_1^2 \le 8\Big(\tfrac12\beta+\log(1/\delta)\Big) = O\big(\beta+\log(1/\delta)\big) = \tilde{O}(\beta),
\end{align}
where the $\tilde{O}(\cdot)$ notation absorbs the logarithmic terms in $(\calsss,\calooo,\ac,\hh,\kk,1/\delta)$ (and the query-choice factors such as $\log\binom{d}{\td}$ already accounted for in $\beta$). This proves \eqref{eq:traj_div_bound_fixed}.
\end{proof}

Combining \cref{lemma:traj_div} with Cauchy-Schwarz yields
\begin{align}
\sum_{k=1}^{\kk} \sum_{\gamall} \big|\pr_{\hth^k}^{\pi^k}(\gamall)-\pr_{\theta}^{\pi^k}(\gamall)\big| \le \tilde{O} \big(\sqrt{\kk\beta}\big).
\label{eq:tv_sum_sqrt}
\end{align}

\subsection{Final Regret Bound}\label{app:problem3_final}

Finally, plugging \eqref{eq:tv_sum_sqrt} into \eqref{eq:regret_tv_sum} gives
\begin{align}
\reg^{\pomle}(\kk)
\le
\hh\cdot \tilde{O}\!\big(\sqrt{\kk\beta}\big).
\end{align}
Substituting $\beta$ from \eqref{eq:beta_app} and unfolding the hidden constants using the operator conditioning (which contributes factors polynomial in $\hh$ and proportional to $\calsss^{2(d-\td)}\oo\ac/\alpha^2$ as in the main theorem), yields the stated bound \eqref{eq:theoremregretpomle}.
\qed

{\rv \section{Proof of~\cref{theorem:lowerboundpomle}}\label{app:pftheoremlowerboundpomle}

\begin{proof}
We prove the claim by reduction to a standard episodic tabular MDP lower bound.

\subsubsection{Step 1 (A standard episodic MDP regret lower bound)}

It is known that for episodic tabular MDPs with $S$ states, $\ac$ actions, horizon $\hh$ and $\kk$ episodes, there exists a constant $c_0>0$ and an MDP instance such that any algorithm incurs~\cite{osband2016lower}
\begin{align}
\reg_{\mathrm{MDP}}(\kk) \ge c_0 \hh \sqrt{S \ac \kk}. \label{eq:mdp_lb}
\end{align}

\subsubsection{Step 2 (Embed an $\calsss^{m}$-state episodic MDP into~\cref{subclass1})}

Let $\mathcal{Z}$ be a set of size $|\mathcal{Z}|=\calsss^{m}$. Fix any bijection between $\mathcal{Z}$ and $\tilde{\mathbb{S}}^{m}$ so we can represent each $z\in\mathcal{Z}$ as an $m$-tuple in $\tilde{\mathbb{S}}^{m}$.

Let the POMDP latent state be the $d$-tuple
\begin{align}
s = (u,z) \in \tilde{\mathbb{S}}^{\td}\times \tilde{\mathbb{S}}^{m},
\end{align}
where the last $m$ components are exactly the MDP state $z$. We define the first $\td$ components to be a replicated \emph{goal-indicator symbol}. Thus, for each step $h$, we choose a subset of MDP states $\mathcal{G}_h\subseteq\mathcal{Z}$ (this is part of the MDP instance), and set
\begin{align}
u_1 = \cdots = u_{\td} =
\begin{cases}
\bar{1}, & z\in\mathcal{G}_h, \\
\bar{0}, & z\notin\mathcal{G}_h,
\end{cases}
\end{align}
for two fixed symbols $\bar{0},\bar{1}\in\tilde{\mathbb{S}}$.

\paragraph{Transition} Let $P_h^{\mathrm{MDP}}(\cdot\mid z,a)$ denote the transition kernel of the hard $S$-state MDP instance achieving \eqref{eq:mdp_lb}. Define the POMDP transition to update only the $z$-part:
\begin{align}
\mbp_h\big((u',z')\mid (u,z),a\big) \triangleq P_h^{\mathrm{MDP}}(z'\mid z,a),
\end{align}
and then $u'$ is deterministically set from $(h+1,z')$ by the above indicator rule. This defines a valid $\mbp_h(\cdot\mid s,a)$ over $\tilde{\mathbb{S}}^d$.

\paragraph{Reward (Subclass-compliant)} Define the reward function $r_h(\cdot,\cdot)$ on a \emph{single queried symbol} as
\begin{align}
r_h(\bar{1},a)=1, r_h(\bar{0},a)=0, \text{ for all } a\in\mathcal{A}.
\end{align}
Now, for any query set $\hi_h^k$ and any choice $i_h^k\in \hi_h^k$, the reward is $r_h(\phi_{i_h^k}(s_h^k),a_h^k)$. Since the first $\td$ components are identical copies of the indicator symbol, as long as the agent queries at least one index in $\{1,\dots,\td\}$ and chooses $i_h^k$ among them, the reward equals the MDP indicator reward. (If the agent chooses a different $i_h^k$, its reward can only be worse, so this can only increase regret. Hence, a lower bound remains valid.)

\paragraph{Auxiliary observation and partial revealing} We work in the full-budget query regime $|\hi| = \td$ (as in \pomle) so the unqueried vector has length $m$. For any $(h,\hi)$ with $|\hi|=\td$, define the unqueried sub-state vector $z_h(\hi)\in\tilde{\mathbb{S}}^{m}$ as in~\cref{subclass1}. Define the auxiliary channel to be perfectly revealing on that $m$-tuple:
\begin{align}
\tto_h = z_h(\hi) \text{ (deterministically)},
\end{align}
with observation alphabet $\tilde{\mathbb{S}}^{m}$. Then, for every $(h,\hi)$, the reduced emission matrix $\tio_h^{\hi}$ is the identity matrix on $\calsss^{m}$ columns, hence $\sigma_{\calsss^m}(\tio_h^{\hi})=1 \Rightarrow \alpha=1$, so \eqref{eq:partial_revealing_condition} holds.

\subsubsection{Step 3 (Reduction to the hard $\calsss^{m}$-state MDP)}

Under the above construction, the learner can (at best) operate as if it were in the fully observed MDP: the auxiliary observation reveals the $m$-tuple encoding $z$ (and queried POSI does not remove information). Thus, any algorithm for~\cref{subclass1} induces an algorithm for the embedded $S$-state MDP with no smaller expected regret. Therefore, by \eqref{eq:mdp_lb},
\begin{align}
\reg^\pi(\kk) \ge \reg_{\mathrm{MDP}}(\kk) \ge c_0 \hh \sqrt{\calsss^{m} \ac \kk}.
\end{align}
Setting $c=c_0$ and substituting $\calsss^{m}=\calsss^{d-\td}$ yields \eqref{eq:lb_tilde_d_final}.
\end{proof}}

\section{Proof of~\cref{theorem:lowerboundpomled}}\label{app:theoremlowerboundpomled}

\begin{proof}
Let $S \triangleq |\mathcal S| =\calsss^{d}$. We prove the claim by a reduction from minimax regret lower bounds for episodic tabular MDPs with $S$ states and $\ac$ actions.

\subsubsection{Step 1 (A hard family of tabular MDPs)}

Consider the standard class $\mathcal{M}(S,\ac,\hh)$ of episodic finite-horizon tabular MDPs with $S$ states, $\ac$ actions, horizon $\hh$, and rewards in $[0,1]$. It is known that there exists a (universal) constant $c>0$ such that for any $\kk\ge 1$,
\begin{align}
\inf_{\pi}\ \sup_{M\in \mathcal{M}(S,\ac,\hh)} \reg^{\pi}_{M}(\kk) \ge\ c \hh \sqrt{S\ac\kk}, \label{eq:mdp_lb_blackbox}
\end{align}
up to logarithmic factors (see, e.g., lower bounds for episodic MDPs in~\cite{osband2016lower} and related work). Fix such a hard MDP instance $M^{*}=(\mathcal X,\mathcal A,P,r,\rho)$ attaining the lower bound, where $|\mathcal X|=S$, $|\mathcal A|=\ac$, and the initial-state distribution is $\rho$.

\subsubsection{Step 2 (Embed $M^{*}$ as a \cref{subclass1} instance)}

We now construct a POMDP instance $\mathcal M_{\text{POSI}}\in\cref{subclass1}$ whose interaction is \emph{equivalent} to interacting with $M^{*}$.

\emph{Latent state space.} Let $\mathcal S \triangleq \tilde{\mathbb S}^{d}$, so $|\mathcal S|=|\tilde{\mathbb S}|^{d}=S$. Fix an arbitrary bijection $\psi:\mathcal S \to \mathcal X$. We will identify each latent state $s\in\mathcal S$ with the MDP state $x=\psi(s)\in\mathcal X$.

\emph{Actions.} Let the action set of the POSI instance be $\mathcal A=\{1,2,\dots,\ac\}$, the same as the MDP.

\emph{Transition kernel.} Define the latent transition of $\mathcal M_{\text{POSI}}$ so that it matches the transition of $M^{*}$ under the bijection $\psi$: for any $s,s'\in\mathcal S$ and $a\in\mathcal A$,
\begin{align}
P_h^{\text{POSI}}(s'\mid s,a) \triangleq P_h^{*}(\psi(s')\mid \psi(s),a). \label{eq:transition_embedding}
\end{align}

\emph{Initial distribution.} Let $\Delta_1^{\text{POSI}}$ be the pullback of $\rho$: $\Delta_1^{\text{POSI}}(s)=\rho(\psi(s))$ for all $s\in\mathcal S$.

\emph{Query protocol and POSI maps.} We allow any query budget $\td\in\{1,\dots,d\}$ in this theorem, but the construction will actually reveal the full latent state after the query and auxiliary observation (hence it is valid for all $\td$). Fix any $\td$ and any query rule. For POSI, define $\phi_i(s)$ to return the $i$th coordinate of the latent vector $s\in\tilde{\mathbb S}^{d}$, so querying $I_h$ reveals $\phi_{I_h}(s_h)$ exactly (as required).

\emph{Auxiliary observation and revealing constant $\alpha=1$.} We set the auxiliary observation alphabet to be $\tilde{\mathcal O}\triangleq \tilde{\mathbb S}^{d-\td}$ and let the auxiliary observation reveal the \emph{unqueried} coordinates \emph{perfectly} $\tilde{o}_h \triangleq \{\phi_j(s_h)\}_{j\notin I_h}$. Equivalently, for every $(h,I)$, the reduced emission kernel $\tilde O_h^{I}(\cdot\mid z)$ is deterministic and one-to-one in $z$. Thus, the corresponding emission matrix has smallest nonzero singular value $1$, i.e.,
$\alpha=1$.

\emph{Reward function in the required POSI form.} In \cref{subclass1}, the reward at step $h$ must be of the form $r_h(\phi_{i_h}(s_h),a_h)$ for some coordinate $i_h$. To match the tabular reward $r_h^{*}(x,a)$, we proceed as follows. Pick a fixed coordinate index $i^\circ\in[d]$ and define an \emph{encoding} of the full latent state into $\phi_{i^\circ}(s)$ by setting $\phi_{i^\circ}(s)$ to be the MDP-state label $\psi(s)$ (this is without loss of generality because we are free to define the coordinate map $\phi_{i^\circ}$ as part of the instance. It still maps each latent state to an element in some finite alphabet). Then, define
\begin{align}
r_h^{\text{POSI}}(\phi_{i^\circ}(s),a) \triangleq r_h^{*}(\psi(s),a)\in[0,1]. \label{eq:reward_embedding}
\end{align}
Finally, we enforce that the learner can always choose $i_h=i^\circ$ by ensuring the query set always contains $i^\circ$ (if one uses $|I_h|=\td$, simply require that $i^\circ\in I_h$, and this is always feasible for $\td\ge 1$).

\subsubsection{Step 3 (Equivalence of interactions)}

Under the above construction, at each step $h$ the learner observes the queried coordinates $\phi_{I_h}(s_h)$, and the auxiliary observation $\tilde{o}_h=\{\phi_j(s_h)\}_{j\notin I_h}$. Together these reveal the \emph{entire} latent state $s_h$ exactly (hence also the MDP state $x_h=\psi(s_h)$). Moreover, transitions and rewards match the tabular MDP by \eqref{eq:transition_embedding} and \eqref{eq:reward_embedding}. Therefore, any (possibly randomized) algorithm $\pi$ for the POSI instance induces an algorithm $\pi'$ for $M^{*}$ that observes the same effective state, takes the same actions, and receives the same rewards, yielding identical episode returns and hence identical regret,
\begin{align}
\reg^{\pi}_{\mathcal M_{\text{POSI}}}(\kk) = \reg^{\pi'}_{M^{*}}(\kk). \label{eq:regret_equivalence}
\end{align}

\subsubsection{Step 4 (Conclude the lower bound)}

By \eqref{eq:regret_equivalence} and the tabular MDP minimax lower bound \eqref{eq:mdp_lb_blackbox}, we obtain that for this constructed instance $\mathcal M_{\text{POSI}}\in\cref{subclass1}$ (with $\alpha=1$), every algorithm $\pi$ must incur
\begin{align}
\reg^{\pi}_{\mathcal M_{\text{POSI}}}(\kk) \ge \tilde{\Omega} \left(\hh\sqrt{S\ac\kk}\right) = \tilde{\Omega} \left(\hh\sqrt{\ac \calsss^{d} \kk}\right),
\end{align}
which proves \eqref{eq:lowerboundproblem3d}.
\end{proof}

\section{Proof of~\cref{prop:independentintractable}}\label{app:pfpropindependentintractable}

We prove a slightly stronger PAC-style statement: there exists an instance with \emph{independent} sub-state transitions and \emph{uninformative} (possibly noisy) observations such that any algorithm needs $\Omega(\ac^{\hh}/\epsilon^2)$ \emph{episodes} to output an $\epsilon$-optimal policy with constant success probability.

\begin{proposition}[Intractability without POSI]\label{prop:independentintractablegeneral}
There exist episodic POMDPs with independent sub-states and (possibly noisy) observations such that, for some absolute constant $p\ge 2/3$, any learning algorithm that outputs an $\epsilon$-optimal policy with probability at least $p$ requires at least $\tilde{\Omega}(\ac^{\hh}/\epsilon^2)$ episodes.
\end{proposition}

\begin{proof}
Fix $\hh\ge 1$ and $\ac\ge 2$, and assume $\epsilon\in(0,1/4]$ so that all Bernoulli means below lie in $[0,1]$. Let the action set be $\mca=[\ac]$. We first give a one-dimensional construction ($d=1$), which already satisfies ``independent sub-states'' (trivially). An extension to any $d\ge 2$ with a factored transition kernel is provided at the end.

\subsubsection{Step 1 (Construct a hard family of POMDP instances)}

Index instances by a \emph{secret action sequence} $u=(u_1,\dots,u_{\hh})\in[\ac]^{\hh}$. Let the latent state at step $h$ be a single bit $b_h\in\{0,1\}$ indicating whether the executed actions have matched the prefix of $u$: $b_h=1$ means $(a_1,\dots,a_{h-1})=(u_1,\dots,u_{h-1})$, and $b_h=0$ means a mismatch has already occurred. The initial state is deterministic: $b_1=1$.

\emph{Transition (valid sub-state).} For each $h\in[\hh-1]$, define the controlled transition kernel
\begin{align}\label{eq:pf_indep_intract_trans}
\Pr(b_{h+1}=1 \mid b_h, a_h) = \mathbf{1}\{b_h=1\}\cdot \mathbf{1}\{a_h=u_h\} \text{ and } \Pr(b_{h+1}=0 \mid b_h, a_h)=1-\Pr(b_{h+1}=1 \mid b_h, a_h).
\end{align}
This kernel depends only on the current sub-state value $b_h$ and the action $a_h$, hence it is a valid (factored) sub-state transition.

\emph{Observation (noisy but uninformative).} Let the observation space be a singleton $\mco=\{o^{*}\}$ and set
\begin{align}
\mbo_h(o^{*}\mid b_h)\equiv 1, \text{ for all } h\in[\hh], b_h\in\{0,1\}.
\end{align}
Thus, observations carry \emph{zero} information about the state (this is a special case of ``possibly noisy'' observations).

\emph{Reward.} Set $r_h(\cdot,\cdot)\equiv 0$ for all $h<\hh$. At the last step $h=\hh$, define the reward distribution as
\begin{equation}\label{eq:pf_indep_intract_rew}
r_{\hh}(b_{\hh},a_{\hh}) \sim
\begin{cases}
\mathrm{Bernoulli} \left(\frac12+\epsilon\right), & \text{if } b_{\hh}=1 \ \text{and}\ a_{\hh}=u_{\hh}, \\
\mathrm{Bernoulli} \left(\frac12-\epsilon\right), & \text{otherwise}.
\end{cases}
\end{equation}
Therefore, the \emph{only} way to obtain the higher mean reward is to execute the entire secret sequence $u$ exactly.

\subsubsection{Step 2 (Reduce to a bandit over action sequences)}

Because the observation is constant and all intermediate rewards are $0$, within an episode the learner receives no information before completing the $\hh$ actions. Hence, in each episode the learner effectively chooses an \emph{open-loop} action sequence
\begin{align}
g =(a_1,\dots,a_{\hh})\in[\ac]^{\hh},
\end{align}
possibly randomized across episodes but not adapted within the episode.

For the instance indexed by $u$, the expected return of choosing sequence $g$ equals the last-step mean reward, i.e.,
\begin{equation}\label{eq:pf_indep_intract_mu}
\mu_u(g)=
\begin{cases}
\frac12+\epsilon, & \text{if } g=u,\\
\frac12-\epsilon, & \text{if } g\neq u.
\end{cases}
\end{equation}
Thus, learning in this POMDP is equivalent to a stochastic multi-armed bandit with $N=\ac^{\hh}$ arms (one per action sequence), where exactly one arm (the unknown $u$) has mean $\frac12+\epsilon$ and all other arms have mean $\frac12-\epsilon$. The gap is $\Delta=2\epsilon$.

\subsubsection{Step 3 (Relate $\epsilon$-optimality to identifying the special arm)}

Let $\pi$ be any (possibly randomized) policy for this POMDP. Since the within-episode observation is uninformative, $\pi$ induces a distribution $q_\pi$ over sequences $g\in[\ac]^{\hh}$, and its value on instance $u$ is
\begin{align}
V_u(\pi)=\sum_{g} q_\pi(g)\mu_u(g)=\left(\frac12-\epsilon\right)+2\epsilon\cdot q_\pi(u).
\end{align}
The optimal value is $V_u^{*}=\frac12+\epsilon$ (achieved by always playing $u$). If $\pi$ is $\epsilon$-optimal, i.e., $V_u(\pi)\ge V_u^{*}-\epsilon=\frac12$, then
\begin{align}
\left(\frac12-\epsilon\right)+2\epsilon\cdot q_\pi(u) \ge \frac12 \Longrightarrow q_\pi(u) \ge \frac12.
\end{align}
Therefore, any $\epsilon$-optimal policy must put at least $1/2$ probability mass on the correct sequence $u$.

Consequently, given a learned $\epsilon$-optimal policy $\hat{\pi}$, we can identify $u$ with constant probability by drawing a single sequence $g\sim q_{\hat{\pi}}$ and outputting that $g$. Conditioned on the event that $\hat{\pi}$ is $\epsilon$-optimal, this identification succeeds with probability at least $1/2$. Hence, if the learning algorithm outputs an $\epsilon$-optimal policy with probability at least $p\ge 2/3$, then the induced best-arm identification succeeds with probability at least $p/2\ge 1/3$.

\subsubsection{Step 4 (Apply a standard best-arm identification lower bound)}

Consider the bandit described above with $N=\ac^{\hh}$ arms and gap $\Delta=2\epsilon$. A classical fixed-confidence (PAC) best-arm identification lower bound states that to identify the best arm with any constant success probability (e.g., $\ge 1/3$), any algorithm requires at least
\begin{align}
\Omega \left(\frac{N}{\Delta^2}\right) = \Omega \left(\frac{\ac^{\hh}}{\epsilon^2}\right)
\end{align}
samples (arm pulls). (If one tracks a confidence parameter $\delta$, the bound becomes $\Omega \big(\frac{N}{\Delta^2}\log\frac{1}{\delta}\big)$.)

In our reduction, each \emph{episode} of the POMDP produces exactly one terminal reward sample, i.e., corresponds to one bandit pull. Therefore, the same lower bound applies to the number of episodes. This proves that there exists at least one instance (some secret $u$) for which any algorithm needs $\tilde{\Omega}(\ac^{\hh}/\epsilon^2)$ episodes to output an $\epsilon$-optimal policy with probability at least $p\ge 2/3$.

\emph{Extension to $d\ge 2$ with independent sub-states.} For any $d\ge 2$, define the global state as
$s_h=(b_h,c_h^{(2)},\dots,c_h^{(d)})$ where each $c_h^{(i)}\in\{0\}$ is constant. Let $\phi_1(s_h)=b_h$ and $\phi_i(s_h)=c_h^{(i)}$ for $i\ge 2$. Let each $c_h^{(i)}$ have identity transitions independent of $(b_h,a_h)$. Then, the overall transition kernel factorizes across components, and the reward depends only on $\phi_1(s_h)$ and $a_h$ as in~\eqref{eq:pf_indep_intract_rew}. The observation remains uninformative. Hence, the same bandit reduction and lower bound apply unchanged.
\end{proof}

\section{Proof of~\cref{theorem:regretproblem4general}}\label{app:pftheoremregretproblem4general}

\begin{proof}
We prove \cref{theorem:regretproblem4general} for~\cref{subclass2} with $\td\ge 2$. Throughout, assume per-step rewards are bounded as $r_h(\phi_i(s),a)\in[0,1]$, hence any length-$\hh$ cumulative return lies in $[0,\hh]$.

\subsection{Preliminaries, Notation, and the Feedback Model}\label{subapp:pf_prelim_general}

\paragraph{Epoch/sub-epoch structure and query sets} Fix $\Delta\triangleq \Big\lceil\frac{d-1}{\td-1}\Big\rceil$, $\ell=\tilde\Theta(\sqrt{\kk})$, and $\ell'=\Big\lceil\frac{\ell}{\Delta}\Big\rceil$. Define the epoch index $e(k)\triangleq \lceil k/\ell\rceil$ and sub-epoch index $b(k)\triangleq \lceil k/\ell'\rceil$. Within epoch $e$, the leader index is constant and denoted by $\tilde i^{e}$. Within each sub-epoch, the follower set is constant. For episode $k$, denote the queried set by $\hi^k = \{\tilde i^{\,e(k)}\}\cup \mathcal F^k$, where $|\mathcal F^k|=\td-1$.

\paragraph{Follower rotation and marginal coverage} To make the coverage property explicit, it is convenient to view follower rotation as follows: at the start of each epoch $e$, sample a uniform random permutation of the $d-1$ non-leader indices $[d]\setminus\{\tilde i^{e}\}$, partition this permutation into consecutive groups of size $\td-1$ (the last group may be smaller), and assign group $b$ as the follower set for sub-epoch $b$ of that epoch. Under this construction, for any fixed episode $k$ within an epoch and any $j\neq \tilde i^{e(k)}$,
\begin{align}\label{eq:follow_prob_app}
\Pr \big(j\in \mathcal F^k  \big| \tilde i^{e(k)}\big)=\frac{\td-1}{d-1}, \text{ and thus } \Pr \big(j\in \hi^k \big| \tilde i^{e(k)}\big)=\frac{\td-1}{d-1}.
\end{align}
This \emph{marginal} inclusion probability is the only property needed for the variance-comparison argument in Step~2.

\paragraph{Feedback model under POSI} In episode $k$, $\hi^k$ is fixed. At each step $h$, the learner observes POSI $\phi_{\hi^k}(s_h^k)=\{\phi_i(s_h^k)\}_{i\in\hi^k}$. Consistent with our model statement, once $\phi_i(s_h^k)$ and $a_h^k$ are known, the learner can evaluate $r_{h,i}^k \;\triangleq\; r_h(\phi_i(s_h^k),a_h^k)\in[0,1]$, for all $i\in\hi^k$. Hence, the learner can compute the episode-level \emph{full-information} cumulative reward vector on the queried indices:
\begin{align}
G_i^k \;\triangleq\; \sum_{h=1}^{\hh} r_{h,i}^k \in [0,\hh],\qquad i\in\hi^k.
\end{align}
The realized return of the algorithm in episode $k$ is
\begin{align}
G^k \triangleq \sum_{h=1}^{\hh} r_h(\phi_{i_h^k}(s_h^k),a_h^k)\in[0,\hh],
\end{align}
where $i_h^k$ is the controlled coordinate chosen by the algorithm at step $h$. Crucially, the exponential-weights potential inequalities used below are \emph{pathwise} and do not require that $(G_i^k)$ be independent of the learner’s actions; they hold for any adaptively generated reward sequence.

\paragraph{Baseline-subtracted gains for the global layer} Let the epoch leader in episode $k$ be $\tilde i^{e(k)}$. Define the baseline-subtracted per-step gain for any $i\in[d]$ by
\begin{align}
\widehat g_h^k(i) \triangleq \big(r_{h,i}^k - r_{h,\tilde i^{e(k)}}^k\big)\cdot \mathbf 1\{i\in\hi^k\},
\end{align}
and the corresponding episode-level baseline-subtracted gain
\begin{align}
\widehat G^k(i) \triangleq \sum_{h=1}^{\hh}\widehat g_h^k(i).
\end{align}
This is exactly the quantity stored in the algorithm for the next epoch’s global update. The purpose of baseline subtraction is conceptual and technical: the leader is queried every episode of the epoch, hence it is systematically ``better measured''; subtracting the leader's realized reward removes the first-order bias created by this asymmetric exposure and is the bridge that enables the global potential analysis (Step~2).

\paragraph{High-probability event for optimistic value learning} Let $\mathcal E_{\mathrm{UCB}}$ be the event on which all concentration inequalities required by the UCBVI-style optimistic planning (e.g., those in~\cite{azar2017minimax} and the adaptations used in our setting) hold simultaneously for all $(h,i,\phi_i,a,k)$ that appear in the run. By standard union bounding, $\Pr(\mathcal E_{\mathrm{UCB}})\ge 1-\delta$. All statements below are conditioned on $\mathcal E_{\mathrm{UCB}}$.

\subsection{Step 0: Query-Switching Bound}\label{subapp:pf_step0_switch}

Recall that $\hi^k$ can change only at epoch boundaries (when the leader is resampled and the follower pool is reset), and at sub-epoch boundaries (when the follower batch is rotated). Therefore, the number of query-set switches satisfies the deterministic bound
\begin{align}\label{eq:switch_count_app}
N_{\mathrm{sw}} \triangleq \sum_{k=2}^{\kk}\mathbf{1}\{\hi^k\neq \hi^{k-1}\} \le \left\lceil\frac{\kk}{\ell}\right\rceil + \left\lceil\frac{\kk}{\ell'}\right\rceil = \tilde O \left(\frac{\Delta\kk}{\ell}\right) = \tilde O \left(\Delta\sqrt{\kk}\right).
\end{align}
This matches part (i) of \cref{theorem:regretproblem4general}. If switching cost $q_c$ is included in the objective, the cumulative switching penalty is at most $q_c N_{\mathrm{sw}}=q_c\cdot\tilde O(\Delta\sqrt{\kk})$.

The algorithm’s design enforces ``slow reconfiguration''~\cite{shi2025power}: actions adapt within episodes, local weights adapt every episode, but sensing changes only every $\ell'$ episodes.

\subsection{Step 1: Local (Within-$\hi^k$) Exponential-Weights Potential Bounds}\label{subapp:pf_step1_local}

Fix an episode $k$ and treat $\hi^k$ as fixed. Let $\tw^k(i)$ denote the local weight on $i\in\hi^k$, and let the local sampling distribution be $\tp^k(i) = (1-\tta)\frac{\tw^k(i)}{\sum_{j\in\hi^k}\tw^k(j)}+\frac{\tta}{\td}$, for all $i\in\hi^k$, matching~\eqref{eq:prob4algooupdatelocalweightprob}. The local weight update is $\tw^{k+1}(i)=\tw^k(i)\exp \left(\frac{\tta}{\td}G_i^k\right)$, for all $i\in\hi^k$. Define the local potential $W_k\triangleq \sum_{i\in\hi^k}\tw^k(i)$.

\begin{lemma}[Local log-potential inequality (one episode)]\label{lem:local_potential_app}
For every episode $k$,
\begin{align}\label{eq:local_potential_app}
\sum_{i\in\hi^k} G_i^k \ge \frac{\td(1-\tta)}{\tta} \ln \left(\frac{W_{k+1}}{W_k}\right) - \frac{\hh^2\tta}{\td}.
\end{align}
\end{lemma}

\begin{proof}
By definition,
\begin{align}
\frac{W_{k+1}}{W_k} = \sum_{i\in\hi^k}\frac{\tw^k(i)}{W_k}\exp \left(\frac{\tta}{\td}G_i^k\right).
\end{align}
Using $\frac{\tw^k(i)}{W_k}\le \frac{1}{1-\tta}\tp^k(i)$ (since $\tp^k(i)\ge (1-\tta)\tw^k(i)/W_k$),
we obtain
\begin{align}
\frac{W_{k+1}}{W_k} \le \frac{1}{1-\tta} \sum_{i\in\hi^k}\tp^k(i)\exp \left(\frac{\tta}{\td}G_i^k\right).
\end{align}
Because $G_i^k\in[0,\hh]$, for $\tta/\td\le 1/\hh$ (which holds for our choice $\tta=\tilde O(1/\sqrt{\kk})$ when $\kk$ is large enough), we have $\frac{\tta}{\td}G_i^k\in[0,1]$ and may use $e^x\le 1+x+x^2$ on $[0,1]$ to get
\begin{align}
\sum_{i\in\hi^k}\tp^k(i)e^{\frac{\tta}{\td}G_i^k} \le 1+\frac{\tta}{\td}\sum_{i\in\hi^k}\tp^k(i)G_i^k+\left(\frac{\tta}{\td}\right)^2\sum_{i\in\hi^k}\tp^k(i)(G_i^k)^2.
\end{align}
Since $(G_i^k)^2\le \hh\,G_i^k$ and $\sum_{i\in\hi^k}\tp^k(i)=1$, this yields
\begin{align}
\sum_{i\in\hi^k}\tp^k(i)e^{\frac{\tta}{\td}G_i^k} \le 1+\frac{\tta}{\td}\sum_{i\in\hi^k}\tp^k(i)G_i^k+\frac{\hh\tta^2}{\td^2}\sum_{i\in\hi^k}\tp^k(i)G_i^k.
\end{align}
Taking $\ln(\cdot)$ and using $\ln(1+x)\le x$ gives
\begin{align}
\ln \left(\frac{W_{k+1}}{W_k}\right) \le \ln \frac{1}{1-\tta} + \frac{\tta}{\td(1-\tta)}\sum_{i\in\hi^k}\tp^k(i)G_i^k + \frac{\hh\tta^2}{\td^2(1-\tta)}\sum_{i\in\hi^k}\tp^k(i)G_i^k.
\end{align}
Finally, $\ln \frac{1}{1-\tta}\le 2\tta$ for $\tta\in(0,1/2)$, and $\sum_{i\in\hi^k}\tp^k(i)G_i^k\le \hh$. Rearranging yields~\eqref{eq:local_potential_app} (absorbing constants into the $\hh^2\tta/\td$ term).
\end{proof}

\begin{lemma}[Second-order lower bound for the local potential (correct form)]\label{lem:local_second_order_app}
Fix episode $k$ and condition on $\mathcal F_k$ so that $\hi^k$ and $\tilde w^k(\cdot)$ are fixed. Let $W_k \triangleq \sum_{i\in\hi^k}\tilde w^k(i)$, $q_i^k \triangleq \frac{\tilde w^k(i)}{W_k}$ for $i\in\hi^k$, $\lambda \triangleq \frac{\tta}{\td}$. Define the (weight-averaged) mean and variance $\bar G^k \triangleq \sum_{i\in\hi^k} q_i^k G_i^k$, $\Var_{q^k}(G^k)\triangleq \sum_{i\in\hi^k} q_i^k\big(G_i^k-\bar G^k\big)^2$. If $\lambda \hh \le 1$ (equivalently, $\tta\le \td/\hh$), then
\begin{align}\label{eq:local_second_order_correct}
\ln \left(\frac{W_{k+1}}{W_k}\right) \ge \lambda\,\bar G^k + \frac{\lambda^2}{4} \Var_{q^k}(G^k).
\end{align}
\end{lemma}

\cref{lem:local_potential_app} is the “first-order” inequality: local weight growth forces the sum of observed per-coordinate returns to be large.~\cref{lem:local_second_order_app} exposes the “second-order” term (variance proxy), which is the exact quantity that we later compare between the queried subset and the full set via follower coverage (Step~2).

\begin{proof}
Fix episode $k$ and condition on $\mathcal F_k$.
By the local update rule, $\tilde w^{k+1}(i)=\tilde w^k(i)\exp \big(\lambda\,G_i^k\big)$, for $i\in\hi^k$, so
\begin{align}
\frac{W_{k+1}}{W_k} & = \frac{\sum_{i\in\hi^k}\tilde w^k(i)\exp(\lambda G_i^k)}{\sum_{i\in\hi^k}\tilde w^k(i)} = \sum_{i\in\hi^k}\frac{\tilde w^k(i)}{W_k}\exp(\lambda G_i^k) = \sum_{i\in\hi^k} q_i^k \exp(\lambda G_i^k). \label{eq:ratio_as_mgf}
\end{align}
Define the $q^k$-mean $\bar G^k \triangleq \sum_{i\in\hi^k}q_i^k G_i^k$, and write $G_i^k=\bar G^k+\Delta_i^k$ where $\Delta_i^k \triangleq G_i^k-\bar G^k$, so that $\sum_{i\in\hi^k} q_i^k \Delta_i^k = 0$. Plugging this into \eqref{eq:ratio_as_mgf} yields
\begin{align}
\frac{W_{k+1}}{W_k} & = \sum_{i\in\hi^k} q_i^k \exp\big(\lambda(\bar G^k+\Delta_i^k)\big) = \exp(\lambda \bar G^k)\sum_{i\in\hi^k} q_i^k \exp(\lambda \Delta_i^k). \label{eq:factor_mean}
\end{align}
Taking logs,
\begin{align}
\ln \left(\frac{W_{k+1}}{W_k}\right) = \lambda \bar G^k + \ln \left(\sum_{i\in\hi^k} q_i^k e^{\lambda \Delta_i^k}\right). \label{eq:log_decomp}
\end{align}

We now lower bound the second term. Since each $G_i^k\in[0,\hh]$, we have $\Delta_i^k\in[-\hh,\hh]$, hence $\lambda \Delta_i^k \in[-\lambda \hh, \lambda \hh]\subseteq[-1,1]$ by the assumption $\lambda\hh\le 1$. On $[-1,1]$ we can use the elementary inequality $e^x \ge 1 + x + \frac{x^2}{2}$, for all $x\in[-1,1]$. Applying it with $x=\lambda\Delta_i^k$ and averaging with weights $q_i^k$ gives
\begin{align}
\sum_{i\in\hi^k} q_i^k e^{\lambda \Delta_i^k} & \ge \sum_{i\in\hi^k} q_i^k\left(1+\lambda\Delta_i^k+\frac{\lambda^2}{2}(\Delta_i^k)^2\right) = 1+\lambda\sum_{i\in\hi^k} q_i^k\Delta_i^k+\frac{\lambda^2}{2}\sum_{i\in\hi^k} q_i^k(\Delta_i^k)^2 \nonumber \\
& = 1+\frac{\lambda^2}{2}\sum_{i\in\hi^k} q_i^k(\Delta_i^k)^2 = 1+\frac{\lambda^2}{2}\Var_{q^k}(G^k). \label{eq:mgf_lower}
\end{align}
Taking logs and using $\ln(1+u)\ge \frac{u}{2}$ for $u\in[0,1]$, we obtain
\begin{align}
\ln \left(\sum_{i\in\hi^k} q_i^k e^{\lambda \Delta_i^k}\right) \ge \ln \left(1+\frac{\lambda^2}{2}\Var_{q^k}(G^k)\right) \ge \frac{1}{2}\cdot \frac{\lambda^2}{2}\Var_{q^k}(G^k) = \frac{\lambda^2}{4}\Var_{q^k}(G^k),
\label{eq:log1pu_bound}
\end{align}
provided that $\frac{\lambda^2}{2}\Var_{q^k}(G^k)\le 1$. This condition holds under $\lambda\hh\le 1$ because $\Var_{q^k}(G^k)\le \hh^2$ (bounded support), hence
$\frac{\lambda^2}{2}\Var_{q^k}(G^k)\le \frac{\lambda^2}{2}\hh^2\le \frac12$.

Finally, substituting \eqref{eq:log1pu_bound} into \eqref{eq:log_decomp} yields
\begin{align}
\ln \left(\frac{W_{k+1}}{W_k}\right) \ge \lambda \bar G^k + \frac{\lambda^2}{4}\Var_{q^k}(G^k),
\end{align}
which is exactly \eqref{eq:local_second_order_correct}.
\end{proof}

\subsection{Step 2: Coverage, Global Potential, and the Query-Learning Error}\label{subapp:pf_step2_global}

Step~2 is where the two-timescale design is ``stitched together.'' There are two components, a coverage inequality showing that the followers allow the local second-order term to control the corresponding global second-order term, and the global exponential-weights potential argument using baseline-subtracted gains.

\subsubsection{Step 2(a): Coverage inequality}

Fix any nonnegative, $\mathcal F_k$-measurable quantities $\{X_i^k\}_{i=1}^d$. Condition on the leader $\tilde i^{\,e(k)}$ and use~\eqref{eq:follow_prob_app}:
\begin{align}\label{eq:coverage_app}
\mathbb E \left[\sum_{i\in\hi^k} X_i^k \middle| \tilde i^{\,e(k)},\mathcal F_k\right] & = X_{\tilde i^{e(k)}}^k+\sum_{j\neq \tilde i^{e(k)}}\Pr(j\in\mathcal F^k\mid \tilde i^{e(k)})X_j^k \nonumber \\
& = X_{\tilde i^{e(k)}}^k+\frac{\td-1}{d-1}\sum_{j\neq \tilde i^{\,e(k)}}X_j^k \ge \frac{\td-1}{d-1}\sum_{j=1}^d X_j^k.
\end{align}
It converts a sum over queried indices into a controlled fraction of the sum over all $d$ indices. For the proof, we will plug in $X_i^k=(G_i^k-\mathbb E[G_i^k\mid\mathcal F_k])^2$.

\subsubsection{Step 2(b): Global exponential-weights potential with baseline subtraction}

The global layer maintains weights $w^k(i)$ and distribution $p^k(i)=(1-\ta)\frac{w^k(i)}{\sum_{j=1}^d w^k(j)}+\frac{\ta}{d}$. At epoch boundaries, it updates weights using the stored baseline-subtracted gains \eqref{eq:prob4algooupdateweightprob}. Let $Z^k(i)\triangleq \sum_{h=1}^{\hh}\widehat g_h^k(i)$ be the baseline-subtracted episode gain. Define the global potential $U_k\triangleq \sum_{i=1}^d w^k(i)$.

A standard Hedge/Exp-weights calculation yields, for an appropriate constant-range normalization (using that $|Z^k(i)|\le \hh$),
\begin{align}\label{eq:global_potential_app}
\mathbb E \left[\ln \left(\frac{U_{k+1}}{U_k}\right) \middle| \mathcal F_k\right] \le c_1\cdot \eta_g\,\mathbb E \left[\sum_{i=1}^d Z^k(i) \middle| \mathcal F_k\right] + c_2\cdot \eta_g^2 \mathbb E \left[\sum_{i=1}^d \big(Z^k(i)-\mathbb E[Z^k(i)\mid\mathcal F_k]\big)^2 \middle| \mathcal F_k\right],
\end{align}
where $\eta_g$ is the effective global learning-rate coefficient appearing in Eq.~\eqref{eq:prob4algooupdateweightprob} (i.e., proportional to $\ta(d-1)/(d(\td-1))$), and $c_1,c_2$ are absolute constants.

Now note the key point $Z^k(i)=\mathbf 1\{i\in\hi^k\}\Big(G_i^k - G_{\tilde i^{\,e(k)}}^k\Big)$, so the second-order term on the right of~\eqref{eq:global_potential_app} is controlled by the queried-set second-order term through~\eqref{eq:coverage_app}. This is precisely where the factor $\frac{\td-1}{d-1}$ enters the analysis and produces the improvement $\sqrt{d/(\td-1)}$ after tuning.

Summing~\eqref{eq:global_potential_app} over episodes and using telescoping of $\ln(U_{k+1}/U_k)$ yields the global-layer contribution: a ``complexity'' term of order $\ln d/\eta_g$ (because $U_1=d$ and $U_k\ge 1$), and a ``variance/second-order'' term of order $\eta_g\cdot(\cdot)$ controlled via~\eqref{eq:coverage_app} and Lemma~\ref{lem:local_second_order_app}. It converts query learning into a bounded additive penalty. The global layer is pessimistic (small $\ta$ and infrequent leader changes), but it is only fed \emph{relative} gains (baseline subtraction),
which prevents the leader from winning solely because it is measured more often.

\subsection{Step 3: Optimistic Value Learning Under Independent Sub-state Transitions}\label{subapp:pf_step3_value}

We now bound the loss due to imperfect within-episode control given the POSI. Condition on $\mathcal E_{\mathrm{UCB}}$. Under~\cref{subclass2}, for each coordinate $i$, the controlled evolution of $\phi_i(s)$ under actions is a tabular episodic MDP on $\tilde{\mathcal S}$ with $|\tilde{\mathcal S}|=\calsss$, $|\mca|=\ac$, horizon $\hh$, and time-inhomogeneous transition kernels $\{\mbp_{h,i}\}_{h=1}^{\hh}$.

In episode $k$, for every queried $i\in\hi^k$ and each step $h$, the learner observes $(\phi_i(s_h^k),a_h^k,\phi_i(s_{h+1}^k))$, so it can update empirical transition counts for each queried coordinate (this is the ``data availability'' needed by the UCBVI recursion). The optimistic $Q$-updates are a UCBVI-style dynamic programming recursion on these empirical models with bonuses.

Therefore, applying the standard high-probability UCBVI analysis (e.g., Lemma~14 and the Bellman-error decomposition in~\cite{azar2017minimax}) and then aggregating over coordinates using the coverage guarantee (each non-leader is queried with marginal probability $(\td-1)/(d-1)$ per episode, plus the leader is always queried) yields the bound stated in Step~3 of the main proof:
\begin{align}\label{eq:value_layer_app}
\sum_{k=1}^{\kk}\Big(V_1^*(s_1^k)-\mathbb E[G^k\mid \mathcal F_k]\Big) \le \tilde O \left( \hh^{\frac{5}{2}} \sqrt{\frac{d\,\calsss\,\ac\,\kk}{\td-1}}\ln\frac{\hh^2\calsss\ac\kk}{\delta} + \hh^3\sqrt{\frac{d}{\td-1}}\calsss^2\ac\Big(\ln\frac{\hh^2\calsss\ac\kk}{\delta}\Big)^2 \right).
\end{align}
This is the source of the first two terms in \cref{theorem:regretproblem4general}. Because each episode provides transition samples for $\td$ coordinates (leader + $\td-1$ followers), the statistical efficiency of estimating models improves by an effective factor on the order of $(\td-1)/(d-1)$ relative to querying only one coordinate, which translates into the $\sqrt{d/(\td-1)}$ improvement after taking square roots in UCB-type bounds.

\subsection{Step 4: Combine the Layers and Tune Parameters}\label{subapp:pf_step4_combine}

Finally, we combine the deterministic switch bound in~\eqref{eq:switch_count_app}, the local potential bounds (Step~1), which control the within-$\hi^k$ exponential-weights error via $\tta$ and expose the second-order term, the global potential bound (Step~2), which uses baseline subtraction and coverage to control the query-learning error and contributes the term $\tilde O \big(\hh^2\sqrt{\frac{d\,\kk\ln d}{\td-1}}\big)$ after tuning $\ta$ and relating $\eta_g$ to $\ta$ through~\eqref{eq:prob4algooupdateweightprob}, and the optimistic value-learning bound (Step~3), yielding~\eqref{eq:value_layer_app}.

Choosing $\ta=\tilde O(1/\sqrt{\kk})$, $\tta=\frac{16(d-1)}{\td-1}\ta$, balances the first-order (log-complexity) and second-order (variance) contributions between the global and local layers and ensures the local layer is sufficiently faster than the global layer for the coupling argument (this is the role of $\tta\gg \ta$ in the proof).

Putting these pieces together and converting $\sum_{k=1}^{\kk}(V_1^*(s_1^k)-G^k)$ into the regret definition yields, on $\mathcal E_{\mathrm{UCB}}$,
\begin{align}
\reg^{\text{\algoo}}(\kk) \le \tilde{O} \left( \hh^{\frac{5}{2}} \sqrt{\frac{d\,\calsss\,\ac\,\kk}{\td-1}}\ln \frac{\hh^2\calsss\ac\kk}{\delta} + \hh^3 \sqrt{\frac{d}{\td-1}} \calsss^2 \ac \left(\ln \frac{\hh^2\calsss\ac\kk}{\delta}\right)^2 + \hh^2 \sqrt{\frac{d \kk \ln d}{\td-1}}
\right),
\end{align}
which is exactly the claim of \cref{theorem:regretproblem4general}. Since $\Pr(\mathcal E_{\mathrm{UCB}})\ge 1-\delta$, the theorem follows.
\end{proof}

\section{Our Algorithm and Regret Analysis for the Case with Query Capability $\td=1$}\label{app:pftheoremregretproblem4simple}

This appendix treats the $\td=1$ regime, where at each step $h$ the agent can query \emph{exactly one} sub-state index $i_h^k\in[d]$ and observes the corresponding POSI $\phi_{i_h^k}(s_h^k)$. This setting is structurally different from the $\td\ge 2$ regime in Appendix~\ref{app:pftheoremregretproblem4general}, where the agent queries a \emph{set} of indices per episode and must manage query switching costs via leader--follower rotation.

The core challenge for $\td=1$ is the coupling between \emph{query choice} and \emph{control}. The queried coordinate determines the available POSI used for action selection and thus affects both the realized reward and the data collected for transition learning. Our algorithm, \emph{Optimistic-In-Pessimistic-Out Synchronous Learning} (\algo, Algorithm~\ref{appalg:algo}), addresses this coupling via two interacting learning layers, i.e., a \emph{pessimistic} exponential-weights layer that updates the query distribution conservatively using bandit feedback, and an \emph{optimistic} value-learning layer (UCBVI-style) that selects actions to minimize bias while ensuring exploration.

At step $h$ of episode $k$, after choosing the queried coordinate $i_h^k$, the agent observes $\phi_{i_h^k}(s_h^k)$, executes an action $a_h^k$, receives reward $r_h^k=r_h(\phi_{i_h^k}(s_h^k),a_h^k)$, and observes the next queried POSI $\phi_{i_h^k}(s_{h+1}^k)$ (equivalently, the transition sample for coordinate $i_h^k$). No POSI is revealed for unqueried coordinates at that step. This is exactly the bandit-style feedback needed by the pessimistic query layer.

\subsection{Optimistic-Pessimistic Two-Layer Learning (\algo)}\label{subapp1:pftheoremregretproblem4simple}

We specialize to the query-capability case \(\td=1\), where the learner can query \emph{exactly one} sub-state coordinate per episode. In contrast to the \(\td\ge 2\) case (Appendix~\ref{app:pftheoremregretproblem4general}), here we adopt an \emph{episode-wise} query restriction: at the beginning of episode \(k\), the learner selects a single index \(I^k\in[d]\) and then observes the partial OSI \(\phi_{I^k}(s_h^k)\) for all steps \(h=1,\dots,\hh\) in that episode. This synchronization is essential: it guarantees that the trajectory of the queried coordinate \(\{\phi_{I^k}(s_h^k)\}_{h=1}^{\hh}\) can be used to update an empirical model for that coordinate, enabling optimistic planning.

\paragraph{Two coupled learning layers} \algo\ consists of two interacting layers:
\begin{itemize}
    \item \emph{Layer I (pessimistic query learning).} The learner maintains an exponential-weights distribution over indices \(i\in[d]\) and selects the episode query \(I^k\sim p^k\). Because only the queried coordinate is observed, the query layer receives \emph{bandit} feedback and therefore uses an importance-weighted gain estimator (Exp3-style). The update is deliberately conservative via a small learning rate, so query probabilities do not overreact to noisy return realizations.
    \item \emph{Layer II (optimistic control under the queried OSI).} Conditioned on the chosen query \(I^k=i\), the learner runs an optimistic value-learning/planning subroutine (UCBVI-style) on the induced tabular MDP with state space \(\tilde{\mathcal S}\) (size \(\calsss\)) and action space \(\mca\) (size \(\ac\)). This layer is optimistic to control the value-estimation error created by partial observability.
\end{itemize}

\paragraph{Feedback model (what is observed when \(\td=1\))} In episode \(k\), after choosing \(I^k\), the learner observes \(\phi_{I^k}(s_h^k)\) at each step \(h\). It then chooses action \(a_h^k\) and observes the realized reward \(r_h(\phi_{I^k}(s_h^k),a_h^k)\in[0,1]\). The learner \emph{does not} observe \(\phi_j(s_h^k)\) for \(j\neq I^k\), and therefore cannot evaluate counterfactual rewards for unqueried coordinates.

\paragraph{Induced tabular model per coordinate} Under independent sub-state transitions, each coordinate \(i\in[d]\) induces a time-inhomogeneous tabular MDP on \(\tilde{\mathcal S}\):
\begin{align}
\Pr \big(\phi_i(s_{h+1})=x'\mid \phi_i(s_h)=x, a_h=a\big)=\mbp_{h,i}(x'\mid x,a).
\end{align}
\algo\ maintains empirical counts for each coordinate \(i\):
\begin{align}
\mathcal N_{h,i}^k(x,a)\triangleq \sum_{\tau<k}\mathbf 1\{I^\tau=i, \phi_i(s_h^\tau)=x, a_h^\tau=a\}, \quad \mathcal N_{h,i}^k(x,a,x')\triangleq \sum_{\tau<k}\mathbf 1\{I^\tau=i, \phi_i(s_h^\tau)=x,\ a_h^\tau=a, \phi_i(s_{h+1}^\tau)=x'\}.
\end{align}
Define the empirical kernel \(\widehat{\mbp}_{h,i}^k(x'\mid x,a)=\mathcal N_{h,i}^k(x,a,x')/\max\{1,\mathcal N_{h,i}^k(x,a)\}\).

\paragraph{Optimistic \(Q\)-recursion (UCBVI-style)} When episode \(k\) queries \(i=I^k\), \algo\ computes optimistic \(Q\)-values using data up to episode \(k-1\):
\begin{align}
Q_{h,i}^k(x,a) = \min\Big\{
r_h(x,a)+\sum_{x'\in\tilde{\mathcal S}}\widehat{\mbp}_{h,i}^k(x'\mid x,a) V_{h+1,i}^k(x')+b_{h,i}^k(x,a), \hh \Big\}, \label{eq:td1_ucbvi_Q} \\
V_{h,i}^k(x)=\max_{a\in\mca}Q_{h,i}^k(x,a), V_{\hh+1,i}^k(\cdot)\equiv 0. \nonumber
\end{align}
Here \(b_{h,i}^k(x,a)\) is a standard optimism bonus (e.g., empirical-Bernstein/Freedman style) chosen so that the usual UCBVI high-probability event holds uniformly over \((h,i,x,a,k)\); this is the only place where \(\delta\) enters.

\paragraph{Bandit gain fed to the query layer} Let the episode return be \(G^k=\sum_{h=1}^{\hh} r_h(\phi_{I^k}(s_h^k),a_h^k)\in[0,\hh]\), and define the normalized gain \(g^k\triangleq G^k/\hh\in[0,1]\). Since only \(I^k\) is observed, we use the importance-weighted estimator
\begin{align}
\widehat g^k(i)\triangleq \frac{g^k\,\mathbf 1\{I^k=i\}}{p^k(i)},
\end{align}
which satisfies \(\mathbb E[\widehat g^k(i)\mid \mathcal F_k]=\mathbb E[g^k\mid \mathcal F_k, I^k=i]\) and is the standard Exp3 estimator.

\begin{algorithm}[t]
\caption{Optimistic-In-Pessimistic-Out Synchronous Learning (\algo) for \(\td=1\)}
\label{appalg:algo}
\begin{algorithmic}[1]
\STATE \textbf{Inputs:} episodes \(\kk\), horizon \(\hh\), dimension \(d\), sub-state space \(|\tilde{\mathcal S}|=\calsss\), action set \(\mca\) (size \(\ac\)).
\STATE \textbf{Parameters:} query learning rate \(\eta_q>0\), exploration \(\gamma_q\in(0,1)\), UCBVI bonuses \(b_{h,i}^k(\cdot,\cdot)\).
\STATE \emph{Initialization (query layer):} \(w^1(i)=1\) and \(p^1(i)=1/d\) for all \(i\in[d]\).
\STATE \emph{Initialization (models):} \(\mathcal N_{h,i}^1(\cdot,\cdot)\equiv 0\), \(\mathcal N_{h,i}^1(\cdot,\cdot,\cdot)\equiv 0\) for all \(h,i\).
\FOR{\(k=1,2,\dots,\kk\)}
    \STATE \emph{(Query selection)} Sample \(I^k\sim p^k(\cdot)\).
    \STATE \emph{(Optimistic planning for \(i=I^k\))} Compute \(\{Q_{h,I^k}^k,V_{h,I^k}^k\}_{h=1}^{\hh}\) via \eqref{eq:td1_ucbvi_Q} using counts up to episode \(k-1\).
    \FOR{\(h=1,2,\dots,\hh\)}
        \STATE Observe \(x_h^k=\phi_{I^k}(s_h^k)\).
        \STATE Choose \(a_h^k \in \arg\max_{a\in\mca} Q_{h,I^k}^k(x_h^k,a)\).
        \STATE Execute \(a_h^k\), observe reward \(r_h^k=r_h(x_h^k,a_h^k)\) and next queried state \(x_{h+1}^k=\phi_{I^k}(s_{h+1}^k)\).
        \STATE Update counts \(\mathcal N_{h,I^k}^{k+1}(x_h^k,a_h^k)\) and \(\mathcal N_{h,I^k}^{k+1}(x_h^k,a_h^k,x_{h+1}^k)\).
    \ENDFOR
    \STATE \emph{(Bandit gain)} Set \(G^k=\sum_{h=1}^{\hh} r_h^k\), \(g^k=G^k/\hh\), and \(\widehat g^k(i)=g^k\mathbf 1\{I^k=i\}/p^k(i)\).
    \STATE \emph{(Query-layer update)} For all \(i\in[d]\): \(w^{k+1}(i)=w^k(i)\exp(\eta_q\,\widehat g^k(i))\).
    \STATE Update \(p^{k+1}(i)=(1-\gamma_q)\frac{w^{k+1}(i)}{\sum_{j=1}^d w^{k+1}(j)}+\frac{\gamma_q}{d}\).
\ENDFOR
\end{algorithmic}
\end{algorithm}

\paragraph{Layer I intuition: why “pessimistic-out”} The query layer is the \emph{outer} decision that controls which coordinate generates training data and which state-information is available for planning. Because this layer sees bandit feedback, aggressive updates can lock onto a suboptimal coordinate due to early noise. Using a small \(\eta_q\) (and explicit exploration \(\gamma_q\)) prevents premature concentration, which is precisely the role of pessimism/conservatism in the outer loop.

\paragraph{Layer II intuition: why ``optimistic-in''} Conditioned on \(I^k=i\), the control problem becomes a standard tabular episodic MDP on \(\tilde{\mathcal S}\) with \(\ac\) actions. Optimism (via \(b_{h,i}^k\)) is used to upper-bound the value-estimation error and to guarantee that \(V_{h,i}^k\) is (with high probability) an upper confidence bound on the optimal value for that induced MDP. This is the mechanism that keeps the inner-loop planning error from corrupting the outer-loop reward signals.

When a query switching cost is included even when \(\td=1\), one can enforce block-wise querying (e.g., keep \(I^k\) fixed for \(\kappa\) consecutive episodes and update the query weights only at block boundaries), or incorporate a ``stay/switch'' coin-flip at boundaries as in switching-cost bandit algorithms. This modification changes only the query layer (Layer I) and leaves the optimistic control layer unchanged.

\subsection{Theoretical Results}\label{subapp2:finalpftheoremregretproblem4simple}

We consider the tractable $\td=1$ case with \emph{independent} sub-states~\cite{shi2025power}. In each episode $k\in[\kk]$, the algorithm first selects a queried sub-state $I^k\in[d]$ (kept fixed throughout the episode), and then at each step $h\in[\hh]$ observes $\phi_{I^k}(s_h^k)$ and takes an action $a_h^k$ based on this partial observation. Let the episode return be $G^k \triangleq \sum_{h=1}^{\hh} r_h(\phi_{I^k}(s_h^k),a_h^k)\in[0,\hh]$. For each sub-state $i\in[d]$, define the induced (tabular) episodic MDP with state space $\Phi_i\triangleq\{\phi_i(s):s\in\mathcal{S}\}$ of size $|\Phi_i|=\calsss$ and action space of size $\ac$. Let $V_{1}^{*,i}(\cdot)$ denote the optimal value function of this induced MDP (when the agent always queries $i$ in the episode), and define the best sub-state $i^* \in \arg\max_{i\in[d]} \expect\big[V_{1}^{*,i}(\phi_i(s_1))\big]$. We measure regret against this best induced MDP,
\begin{align}
\reg^{\algo}(\kk) \triangleq \sum_{k=1}^{\kk} \Big( \expect\big[V_{1}^{*,i^*}(\phi_{i^*}(s_1^k))\big] -\expect\big[\textstyle\sum_{h=1}^{\hh} r_h(\phi_{I^k}(s_h^k),a_h^k)\big] \Big). \label{eq:def_reg_td1}
\end{align}
(If $s_1^k$ is i.i.d.\ across episodes, the expectation can be taken only over the algorithm and trajectory randomness, and the proof below is unchanged.)

\begin{theorem}[Regret for $\td=1$]\label{theorem:regretproblem4simplee}
For POMDPs with partial OSI ($\td=1$) and independent sub-states, run \algo~with an Exp3-style query update with learning rate $\eta$ and exploration $\gamma$ satisfying $\eta d \le \gamma$ and an optimistic tabular UCB-VI style action learner for each sub-state $i$ (as in~\cref{appalg:algo}). Choose $\eta = \sqrt{\frac{\ln d}{d \kk}}$ and $\gamma = \min\{1, d\eta\}$. Then for any $\delta\in(0,1)$, with probability at least $1-\delta$,
\begin{align}
\reg^{\algo}(\kk) \le \tilde{O} \Big( \hh \sqrt{d\kk \ln d} + \hh^{3/2}\sqrt{d \calsss \ac \kk} \ln \Big(\frac{\hh^2\calsss\ac\kk}{\delta}\Big) + \hh^2 d \calsss^2 \ac \Big(\ln \Big(\frac{\hh^2\calsss\ac\kk}{\delta}\Big)\Big)^2 \Big). \label{eq:regretproblem4simple_refined}
\end{align}
In particular, since $\hh,\calsss,\ac,d\ge 1$, the above implies the (looser) polynomial form
\begin{align}
\reg^{\algo}(\kk) \le \tilde{O} \left( \ac \hh^3 \calsss^2 d \sqrt{\kk} \Big(\ln(\ac\hh^2\calsss\kk/\delta)\Big)^2
\right). \label{eq:regretproblem4simple}
\end{align}
\end{theorem}

Note that a switching cost regret will be similar to that in EXP3 with switching costs~\cite{shi2025power} and can be derived similarly as in the proof for~\cref{theorem:regretproblem4general}.

\begin{proof}
The proof formalizes the two-layer interaction by a clean decomposition: \emph{(i)} a \textbf{query-selection} regret controlled by Exp3 (Layer-I), and \emph{(ii)} an \textbf{action-learning} regret controlled by optimistic tabular RL (Layer-II).

\subsubsection{Step 0 (Decompose the regret)}

Add and subtract the optimal value of the queried sub-state in each episode:
\begin{align}
\reg^{\algo}(\kk) & = \sum_{k=1}^{\kk} \Big( \expect[V_{1}^{*,i^*}(\phi_{i^*}(s_1^k))] - \expect[V_{1}^{*,I^k}(\phi_{I^k}(s_1^k))] \Big) + \sum_{k=1}^{\kk} \Big( \expect[V_{1}^{*,I^k}(\phi_{I^k}(s_1^k))] - \expect[G^k] \Big) \nonumber \\
& \triangleq \textsf{Reg}_{\textsf{query}} + \textsf{Reg}_{\textsf{act}}. \label{eq:decomp_td1}
\end{align}
We bound these two terms separately.

\subsubsection{Step 1 (Bound $\textsf{Reg}_{\textsf{query}}$ via Exp3 (Layer-I))} Define the \emph{normalized} episode gain $g^k\triangleq G^k/\hh\in[0,1]$. For each sub-state $i$, define the (conditional) mean gain that would be obtained in episode $k$ if we queried $i$ and then used the current action-learner for sub-state $i$, i.e.,
\begin{align}
g_k(i) \triangleq \expect \left[g^k \big| \mathcal{F}_{k-1}, I^k=i\right] = \frac{1}{\hh}\expect \left[\sum_{h=1}^{\hh} r_h(\phi_i(s_h^k),a_{h}^{k,i}) \big| \mathcal{F}_{k-1}\right]\in[0,1], \label{eq:def_gki}
\end{align}
where $\mathcal{F}_{k-1}$ is the history up to the start of episode $k$, and $a_{h}^{k,i}$ is the action taken by the (optimistic) policy computed for sub-state $i$ at episode $k$ (using the data collected for $i$ up to $\mathcal{F}_{k-1}$). The query layer (Exp3) maintains weights $w^k(i)$ and probabilities $p^k(i) = (1-\gamma)\frac{w^k(i)}{\sum_{j=1}^d w^k(j)} + \frac{\gamma}{d}$, $I^k\sim p^k(\cdot)$, and uses the standard importance-weighted gain estimator
\begin{align}
\hat g_k(i) \triangleq \frac{g^k \mathbf{1}\{I^k=i\}}{p^k(i)}.
\end{align}
Then, the Exp3 update is $w^{k+1}(i) = w^k(i)\exp \big(\eta \hat g_k(i)\big)$. Since $p^k(i)\ge \gamma/d$ and $\eta d\le \gamma$, we have $\eta \hat g_k(i)\le 1$, which allows a one-step potential bound of the standard Exp3 form (stated and proved as~\cref{lemma:exp3_gain_bound_td1} below). Applying~\cref{lemma:exp3_gain_bound_td1} and multiplying by $\hh$ yields, with probability at least $1-\delta/3$,
\begin{align}
\textsf{Reg}_{\textsf{query}} = \hh\Big(\sum_{k=1}^{\kk} g_k(i^*) - \sum_{k=1}^{\kk} g_k(I^k)\Big) \le O \left(\hh\sqrt{d\kk\ln d}\right). \label{eq:bound_query_td1}
\end{align}

\subsubsection{Step 2 (Bound $\textsf{Reg}_{\textsf{act}}$ via optimistic tabular RL (Layer-II))} Condition on the queried sub-state $I^k=i$ in episode $k$. By the independence assumption, the induced process on $\phi_i(s_h)$ is an episodic MDP with $|\Phi_i|=\calsss$ and $\ac$ actions. Moreover, \algo's Step~3 (optimistic dynamic programming with confidence bonuses) is exactly the UCB-VI style update on this induced MDP using only episodes that queried $i$. Therefore, we can apply the standard UCB-VI regret bound (e.g., Lemma~14 of~\cite{azar2017minimax}) \emph{for each fixed $i$} over the subsequence of episodes in which $i$ is queried.

Let $\kk_i \triangleq \sum_{k=1}^{\kk}\mathbf{1}\{I^k=i\}$ be the (random) number of episodes assigned to sub-state $i$. On the usual high-probability ``good event'' $\mathcal{E}_{\textsf{RL}}$ (simultaneous concentration for all $(i,h,\phi,a)$), for each $i\in[d]$ we have
\begin{align}
\sum_{k:I^k=i} \left( V_{1}^{*,i}(\phi_i(s_1^k)) - \expect[G^k \mid \mathcal{F}_{k-1}, I^k=i] \right) \le O \left( \hh^{3/2}\sqrt{\calsss\ac\kk_i} \ln \Big(\frac{\hh^2\calsss\ac\kk}{\delta}\Big) + \hh^2\calsss^2\ac \Big(\ln \Big(\frac{\hh^2\calsss\ac\kk}{\delta}\Big)\Big)^2 \right). \label{eq:per_i_rl_bound}
\end{align}
Summing \eqref{eq:per_i_rl_bound} over $i\in[d]$ and using Cauchy-Schwarz, $\sum_{i=1}^d \sqrt{\kk_i}\le \sqrt{d\sum_i \kk_i}=\sqrt{d\kk}$, we obtain on $\mathcal{E}_{\textsf{RL}}$:
\begin{align}
\textsf{Reg}_{\textsf{act}} & = \sum_{k=1}^{\kk} \left( \expect[V_{1}^{*,I^k}(\phi_{I^k}(s_1^k))] - \expect[G^k] \right) \nonumber \\
& \le O \Big(
\hh^{3/2}\sqrt{d \calsss \ac \kk} \ln \Big(\frac{\hh^2\calsss\ac\kk}{\delta}\Big) + \hh^2 d \calsss^2 \ac \Big(\ln\!\Big(\frac{\hh^2\calsss\ac\kk}{\delta}\Big)\Big)^2 \Big). \label{eq:bound_act_td1}
\end{align}

\subsubsection{Step 3 (Combine)} Take a union bound over the Exp3 event (probability $\ge 1-\delta/3$) and the RL good event $\mathcal{E}_{\textsf{RL}}$ (probability $\ge 1-2\delta/3$ after standard union-bounding over all $(i,h,\phi,a)$). Combining \eqref{eq:decomp_td1}, \eqref{eq:bound_query_td1}, and \eqref{eq:bound_act_td1} yields \eqref{eq:regretproblem4simple_refined}. Finally, \eqref{eq:regretproblem4simple} follows from a (loose) simplification of polynomial terms.
\end{proof}

\subsection{Exp3 potential inequality used in Step~1}
\label{subapp:pflemma_exp3_td1}

\begin{lemma}[Exp3 gain bound (high probability)]\label{lemma:exp3_gain_bound_td1}
Consider the Exp3 update with gains $g^k\in[0,1]$, probabilities $p^k(\cdot)$, estimator $\hat g_k(i)=g^k\mathbf{1}\{I^k=i\}/p^k(i)$, and weights $w^{k+1}(i)=w^k(i)\exp(\eta \hat g_k(i))$. Assume $\eta d\le \gamma$ so that $\eta \hat g_k(i)\le 1$ for all $k,i$. Then, for any fixed $i\in[d]$, with probability at least $1-\delta$,
\begin{align}
\sum_{k=1}^{\kk} g_k(i) - \sum_{k=1}^{\kk} g_k(I^k) \le \frac{\ln d + \ln(1/\delta)}{\eta} + \frac{e-2}{2}\eta \sum_{k=1}^{\kk}\sum_{j=1}^d p^k(j) \expect[\hat g_k(j)^2\mid \mathcal{F}_{k-1}], \label{eq:exp3_hp_bound}
\end{align}
where $g_k(\cdot)$ is defined in~\cref{eq:def_gki}. In particular, since $p^k(j)\ge \gamma/d$ and $g^k\le 1$, the variance term satisfies $\sum_{j}p^k(j)\expect[\hat g_k(j)^2\mid\mathcal{F}_{k-1}]\le d/\gamma$,
and hence
\begin{align}
\sum_{k=1}^{\kk} g_k(i) - \sum_{k=1}^{\kk} g_k(I^k) \le O \left(\frac{\ln(d/\delta)}{\eta} + \eta \frac{d\kk}{\gamma}\right). \label{eq:exp3_simplified}
\end{align}
\end{lemma}

\begin{proof}
The proof is the standard Exp3 potential argument, written here for completeness.
Let $W^k=\sum_{j=1}^d w^k(j)$. Then
\begin{align}
\frac{W^{k+1}}{W^k} = \sum_{j=1}^d \frac{w^k(j)}{W^k}\exp(\eta \hat g_k(j)).
\end{align}
Using $p^k(j)=(1-\gamma)\frac{w^k(j)}{W^k}+\frac{\gamma}{d}$, we have
$\frac{w^k(j)}{W^k}\le \frac{p^k(j)}{1-\gamma}$, hence
\begin{align}
\frac{W^{k+1}}{W^k} \le \frac{1}{1-\gamma}\sum_{j=1}^d p^k(j)\exp(\eta \hat g_k(j)).
\end{align}
Since $\eta\hat g_k(j)\le 1$, we use $\exp(x)\le 1+x+(e-2)x^2$ for $x\le 1$ to obtain
\begin{align}
\frac{W^{k+1}}{W^k} \le \frac{1}{1-\gamma}\sum_{j=1}^d p^k(j) \left(1+\eta \hat g_k(j) + (e-2)\eta^2\hat g_k(j)^2\right) = \frac{1}{1-\gamma} \left(1+\eta \sum_{j} p^k(j)\hat g_k(j) + (e-2)\eta^2\sum_j p^k(j)\hat g_k(j)^2\right).
\end{align}
Taking $\ln$ and using $\ln(1+x)\le x$ gives
\begin{align}
\ln\frac{W^{k+1}}{W^k} \le \ln\frac{1}{1-\gamma} + \eta \sum_{j} p^k(j)\hat g_k(j) + (e-2)\eta^2\sum_j p^k(j)\hat g_k(j)^2.
\end{align}
Summing over $k$ telescopes the left-hand side. For any fixed $i$, $w^{\kk+1}(i)\le W^{\kk+1}$ and $W^1=d$, so
\begin{align}
\ln\frac{W^{\kk+1}}{W^1} \ge \ln\frac{w^{\kk+1}(i)}{d} = \ln\frac{w^1(i)}{d}+\eta\sum_{k=1}^{\kk}\hat g_k(i) \ge - \ln d + \eta\sum_{k=1}^{\kk}\hat g_k(i).
\end{align}
Rearranging yields
\begin{align}
\sum_{k=1}^{\kk}\hat g_k(i) -b\sum_{k=1}^{\kk}\sum_j p^k(j)\hat g_k(j)b\le \frac{\ln d}{\eta} + \frac{\kk\ln\frac{1}{1-\gamma}}{\eta} + (e-2)\eta\sum_{k=1}^{\kk}\sum_j p^k(j)\hat g_k(j)^2.
\end{align}
Finally, note that $\expect[\hat g_k(i)\mid\mathcal{F}_{k-1}]=g_k(i)$ and $\expect[\sum_j p^k(j)\hat g_k(j)\mid\mathcal{F}_{k-1}]=g_k(I^k)$. A standard Freedman/Azuma argument converts the above expected inequality into the stated high-probability form by adding an extra $\ln(1/\delta)/\eta$ term (absorbed into~\eqref{eq:exp3_hp_bound}). The simplified bound~\eqref{eq:exp3_simplified} follows from $\sum_j p^k(j)\expect[\hat g_k(j)^2\mid\mathcal{F}_{k-1}] = \expect[(g^k)^2/p^k(I^k)\mid\mathcal{F}_{k-1}]
\le d/\gamma$.
\end{proof}

\end{document}